\documentclass{article}

\PassOptionsToPackage{numbers,sort&compress}{natbib}
\usepackage[eandd, preprint]{neurips_2026}
\usepackage{microtype}
\usepackage{graphicx}
\usepackage{booktabs} %

\usepackage{algorithm}
\usepackage{algorithmic}
\usepackage{wrapfig,lipsum}
\usepackage{bbding}
\usepackage{array}
\usepackage{makecell} %

\usepackage{hyperref}
\usepackage{multirow}
\usepackage{multicol}

\usepackage{longtable}
\usepackage{rotating}

\usepackage{amsmath}
\usepackage{amssymb}
\usepackage{mathtools}
\usepackage{dsfont}
\usepackage{amsthm}
\usepackage{bbm}
\usepackage{placeins}
\usepackage{subcaption}
\usepackage[capitalize,noabbrev]{cleveref}
\usepackage[table]{xcolor}
\theoremstyle{plain}

\theoremstyle{definition}

\theoremstyle{remark}

\usepackage{amsmath,amsfonts,bm}

\def\eqref#1{equation~\ref{#1}}

\def\1{\bm{1}}

\DeclareMathAlphabet{\mathsfit}{\encodingdefault}{\sfdefault}{m}{sl}
\SetMathAlphabet{\mathsfit}{bold}{\encodingdefault}{\sfdefault}{bx}{n}

\DeclareMathOperator*{\argmax}{arg\,max}

\def\compl{o}
\def\extrcompl{\hat{o}}
\def\prompt{q}
\def\nprops{n_{\text{props}}}
\def\reward{r}

\def\prop{{\gamma_i}}
\def\propnoi{{\gamma_1}}

\def\obj{{\nu_i}}
\def\objnoi{{\nu_1}}

\def\marg{{\sigma_i}}
\def\margnoi{{\sigma_1}}

\def\reference{{x_i}}
\def\referencenoi{{x_1}}

\def\fullreward{\reward(\prompt, \extrcompl)}
\def\model{{\pi_\theta}}
\def\benchname{\textsc{MolRGen}}
\def\rollouts{{n_r}}

\def\molstralnsteps{200}

\def\pp#1{\left(#1\right)}

\newcommand{\topk}[2][n_r]{\text{top}_{#2,#1}}

\newcommand{\cs}{CNRS, CentraleSupélec}
\newcommand{\ups}{Université Paris-Saclay}
\newcommand{\mila}{Mila -- Quebec AI Institute}
\newcommand{\ets}{ÉTS Montréal}
\newcommand{\ills}{ILLS -- International Laboratory on Learning Systems}
\newcommand{\livia}{LIVIA}
\newcommand{\mistral}{Mistral AI}

\title{
    \begin{minipage}{0.17\textwidth}
    \includegraphics[width=\textwidth]{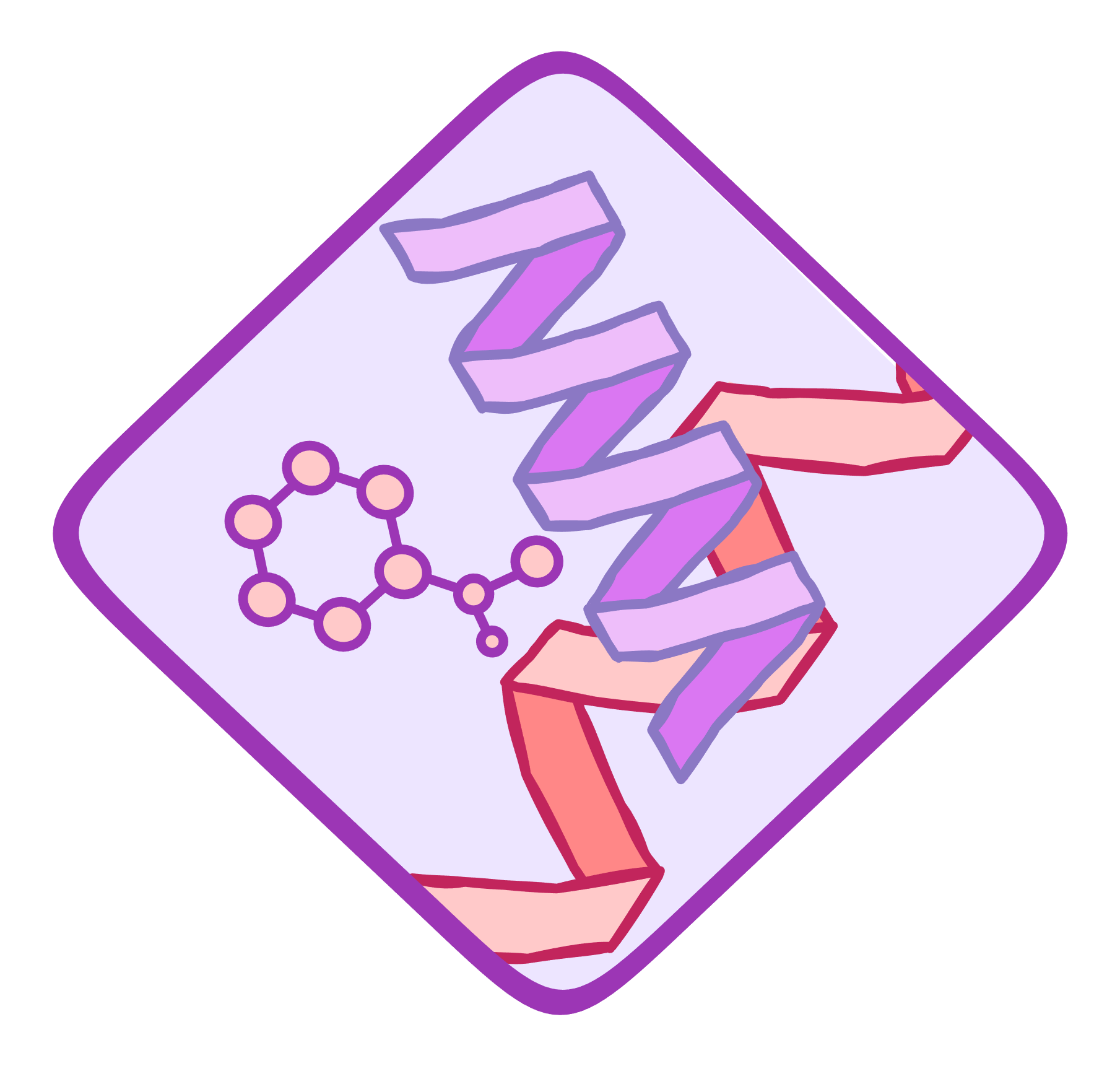}
  \end{minipage}
  \begin{minipage}{0.8\textwidth}
      \benchname: A Benchmark and Verifier for Reasoning LLMs on De Novo Molecular Generation
  \end{minipage}
}
\author{
Philippe Formont\textsuperscript{1,2,3,4}\quad
Maxime Darrin\textsuperscript{5}\quad
Ismail Ben Ayed\textsuperscript{2,3,6}\quad
Pablo Piantanida\textsuperscript{1,3,4,7}
\\[0.5ex]
\textsuperscript{1}\ups \\
\textsuperscript{2}\ets \\
\textsuperscript{3}\ills \\
\textsuperscript{4}\mila \\
\textsuperscript{5}\mistral \\
\textsuperscript{6}\livia \\
\textsuperscript{7}\cs
}

\begin{document}
\maketitle

\begin{abstract}
    Recent reasoning-based large language models have shown strong performance on tasks with verifiable outcomes, but their use in de novo molecular generation remains limited by the lack of training environments where rewards can be computed without reference molecules.
We introduce \benchname, a benchmark and molecular verifier for training and evaluating reasoning LLMs on de novo molecular generation.\footnote{\href{https://github.com/Fransou/MolRGen}{Code}, \href{https://fransou.github.io/MolRGen/}{Documentation}, \href{https://huggingface.co/datasets/Franso/MolRGen}{Dataset}}
\benchname~contains approximately 4,500 protein-pocket targets, resulting in 50k multi-objective optimization prompts combining docking scores with molecular properties such as QED, synthetic accessibility, logP, and physicochemical descriptors.
Unlike caption-based generation or molecule-editing benchmarks, \benchname~evaluates molecules proposed from scratch by computing rewards at generation time.
We benchmark general-purpose and chemistry-specialized open-source LLMs and introduce a diversity-aware top-k metric to measure whether models can generate a diverse set of high-scoring molecules.
Finally, we use the verifier to fine-tune a 128B LLM with GRPO, showing improved performance, at the cost of a diversity-exploitation trade-off.
\benchname~provides a scalable testbed for studying verifier-based reasoning and reinforcement learning in molecular design.

\end{abstract}

\section{Introduction}
The difficulty of identifying molecules with the properties required to become viable drugs has motivated the development of a wide range of molecular generative models, enabling faster exploration of chemical space~\citep{reinvent4,genetalgo}.
This task is particularly challenging because of its inherently multi-objective nature: generated candidates must simultaneously satisfy multiple criteria, such as modulating the expression of a specific protein, reducing toxicity, and having adequate pharmacokinetic properties (e.g., oral bioavailability or the ability to cross the intestinal barrier).
These requirements are typically aggregated into a single reward or scoring function to facilitate optimization and decision-making.

In this work, we focus on \textit{de novo} molecular generation, a setting in which no high-scoring molecule is known \textit{a priori}, and the generative algorithm must identify promising molecules without prior knowledge of which structures may be viable.
Several approaches have been explored to address this challenge, with reinforcement learning (RL) emerging as a natural paradigm: an agent generates molecular compounds, which are then evaluated using a scoring function, often based on computational tools such as docking software~\citep{hassan_protein-ligand_2017,verdonk_improved_2003}, and the resulting scores guide future generations.

Recent work has successfully applied LLMs to various chemical tasks, including caption-based molecular generation and molecular optimization (editing existing molecules to alter properties), and reasoning-based LLMs to drug discovery~\citep{li2025molr1explicitlongcotreasoning, ether0}, motivated by advances achieved in other domains such as mathematical problem-solving and coding.
Training a reasoning LLM relies on reinforcement learning (RL), training the model to generate completions that maximize a reward function.
This reward function is typically computed at runtime through a "verifier", which evaluates the quality of generated completions and provides feedback to the model.

These studies have demonstrated that LLMs can effectively learn molecular representations and reasoning patterns when reference data is available (e.g., a molecule corresponding to the input caption). Still, such reference data is unavailable in practice for \textit{de novo} generation.
\benchname~encapsulates \textit{de novo} molecular generation in an RL environment, leveraging an open-source verifier to compute the reward associated with the generated completion of language to a prompt, at generation time, which can be used both for evaluation and training.
To the best of our knowledge, there is currently no large-scale benchmark including \textit{de novo} molecular generation tasks to train and evaluate reasoning-based LLMs with rewards computed at generation time.

In this work, we aim to build a benchmark and an open-source verifier that would enable the training and evaluation of reasoning-based LLMs for \textit{de novo} molecular generation, by evaluating the quality of generated molecules with rewards computed at generation time, without any prior knowledge of high-scoring candidates.
We summarize our contribution as follows:
\begin{itemize}
    \item We introduce \benchname, a large-scale dataset for \textit{de novo} molecular generation, leveraging ~4.5k protein pocket targets, which we completed with property prediction auxiliary tasks.
    \item We release an open-source molecular verifier that computes rewards at generation time, enabling RL training without reference high-scoring molecules.
    \item We evaluate multiple open-source LLMs on our benchmark, and show that our verifier can be leveraged to train LLMs with RL, improving their performance on the generation task.
\end{itemize}

\section{Related Work}

\paragraph{Molecular Generation.}
Molecular generation has been a long-standing challenge in computational chemistry and drug discovery.
Traditional approaches include generative models, such as Variational Autoencoders (VAEs)~\citep{vae20218} and genetic algorithms~\citep{genetalgo}, which have demonstrated strong empirical results in generating novel molecules.
Reinforcement learning (RL) based methods have emerged as a natural paradigm for generating drug candidates while optimizing for a desirability score.
Notably, Reinvent~\citep{reinvent4} generates SMILES representations of molecules (textual representation) by adjusting the log-likelihood of a prior model using this desirability score.
Other approaches propose generating molecules in different modalities, such as graph representations~\citep{rgfn, gao2024generativeartificialintelligencenavigating, haddad_targeted_2025} or 3D point clouds~\citep{zhou2025guidingdiffusionmodelsreinforcement}.
BindGPT~\citep{zholus_bindgpt_2025} recently proposed to train a 100M parameter language model architecture, using GRPO to optimize its generation on 100 docking targets.
While using a language model architecture, this model is not trained to generate or process natural language.

\paragraph{Large Language Models in Chemistry.}
The application of large language models (LLMs) to chemistry has rapidly expanded, with early work assessing LLM knowledge in chemistry~\citep{mirza2024largelanguagemodelssuperhuman}, showing that the majority of these models outperform human scores.
To further advance the comprehension of chemistry by these models, various methods have been proposed to train and evaluate language models on fundamental tasks, such as SMILES-to-IUPAC translation, molecular captioning, and molecular generation from structural descriptions~\cite{yu2024llasmoladvancinglargelanguage, kim2025molllamageneralunderstandingmolecules}.
Recent work has leveraged 3D molecular representations~\citep{li20243dmoleculetextinterpretationlanguage} or diffusion models~\citep{textguidedmoleculegenerationdiffusion} to perform generation tasks.
A critical distinction exists between caption-based generation and our approach.
Caption-based generation aims to find compounds matching a given natural language description (e.g., \textit{``The molecule is a primary arylamine [...] and a member of anilines''} from Mol-Instructions~\cite{fang2024molinstructionslargescalebiomolecularinstruction}), while our \textit{de novo} molecular generation objective seeks to create novel compounds that optimize a desirability score, rather than match a specific description (e.g., \textit{``Generate a molecule that inhibits the expression of ...''}).

\paragraph{Reasoning in Drug Discovery.}
With the emergence of LLM reasoning models and their demonstrated capabilities on tasks such as mathematical problem-solving and coding, several works have investigated chain-of-thought reasoning for chemical and biological tasks.
BioReason~\citep{bioreason} incorporated a multimodal architecture for DNA sequence tasks, combining fine-tuning on chain-of-thought reasoning from standard LLMs with GRPO optimization on ground-truth KEGG labels.
For molecular applications, reasoning-enhanced models~\citep{zhuang2025reasoningenhancedlargelanguagemodels} leverage structured reasoning to improve molecular property and toxicity predictions.
Mol-R1~\citep{li2025molr1explicitlongcotreasoning}, Ether0~\citep{ether0}, and Chem-DFM~\citep{Zhao_2025} combine methods such as rejection sampling, distillation, supervised fine-tuning, or reinforcement learning to perform multiple chemical tasks, including chemical reaction understanding or caption-based molecular generation.
GeLLMO~\citep{dey2025gellmogeneralizinglargelanguage} tackles a different problem: modifying existing molecules to change their properties, by using at training time a set of molecular pairs labeled with their property modifications.

\section{A Benchmark for De Novo Molecular Generation}
\label{sec:dataset}

\subsection{Notations}

We denote $\mathcal{X}$ the set of all possible molecules.
For each task, we denote $\nprops$ the number of tasks to be performed simultaneously (in practice, $\nprops > 1$ only for generation tasks, where we optimize multiple properties simultaneously).
For $1\leq i\leq \nprops$, we will use the abusive notation $\prop : \mathcal{X} \rightarrow \mathbb{R}$ a molecular property, such as the docking score to a target pocket for molecular generation, or the mapping to the label of a molecule for property prediction (such as the solubility of the molecule) disregarding possible metadata required to compute the property (e.g., the target pocket for docking scores) for clarity of notations.
Additionally, we will assume that all properties are rescaled in $[0,1]$ to simplify the notations.
For property prediction tasks, we also denote $\reference \in \mathcal{X}$ the molecule the model is tasked to predict the property of.

A prompt $\prompt$ is then defined as $\prompt = \{\obj, \prop, \reference, \marg\}_{1\leq i \leq \nprops}$, where: $\obj$ is the task objective (e.g., property maximization, class prediction), and $\marg$ is a numerical metadata used to compute the reward (e.g., property threshold or label standard deviation).

Given a model $\model$ and prompt $\prompt$, a completion $\compl \sim \model(\prompt)$ is generated, and we extract the final answer $\extrcompl$ from tags \texttt{<answer>} and \texttt{</answer>}.
The notation $\prop(\extrcompl)$ denotes the property value resulting from computing a property $\prop$ to the extracted completion (rescaled in $[0,1]$).
\autoref{tab:all_rewards} illustrates these notations and provides the corresponding reward functions for each task.

\begin{figure}
    \begin{subfigure}{0.68\linewidth}
        \centering
        \includegraphics[width=\linewidth]{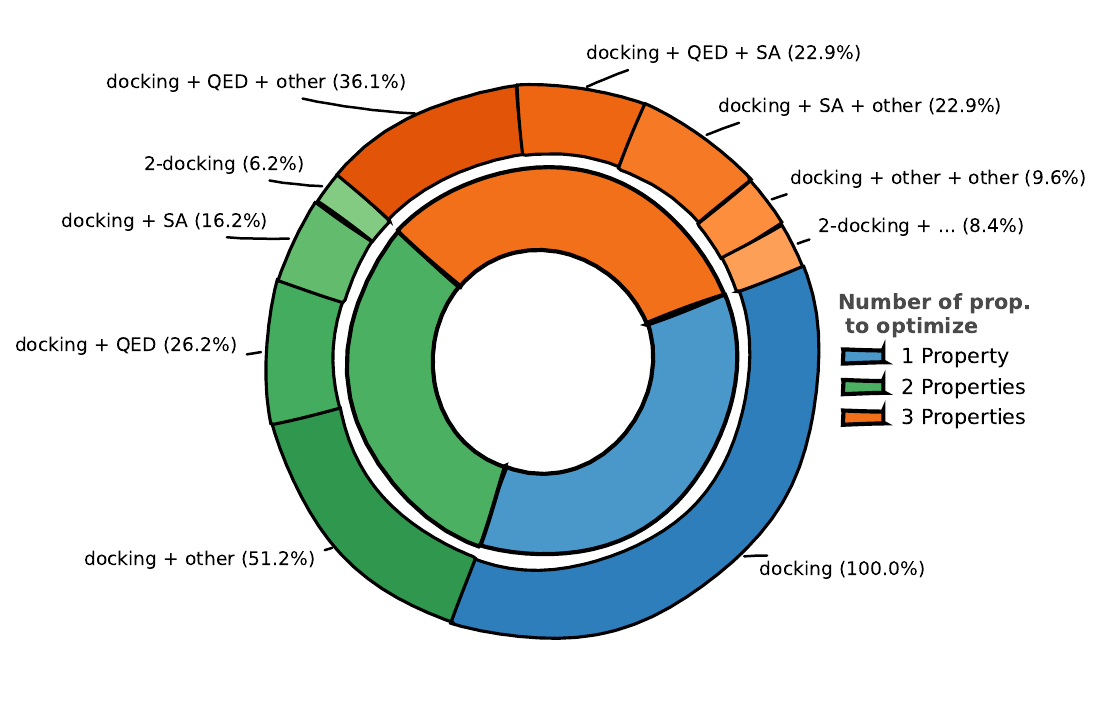}
        \caption{
            \textbf{Optimization tasks for de novo generation.}
            Representation of the optimization objectives in the dataset for de novo generation.
            The dataset is well-balanced between single-property (29\%), 2-properties (36\%), and 3-properties (35\%) optimization.
        }
        \label{fig:donut_dataset}
    \end{subfigure}
    \hspace{0.01\linewidth}
    \begin{subfigure}{0.3\linewidth}
        \centering
        \includegraphics[width=\linewidth]{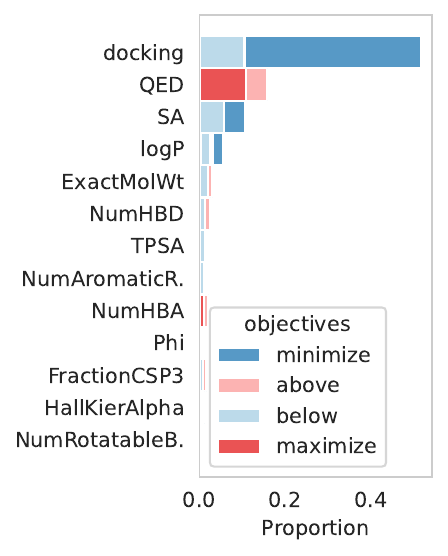}
        \caption{
            \textbf{Distribution of the properties.}
            Proportion of each property's occurrence for the generation tasks.
        }
        \label{fig:reward_type}
    \end{subfigure}
\end{figure}

\subsection{De Novo Molecular Generation.}

\subsubsection{Data Collection}
\label{subsec:gen_data_collection}

The primary task in our dataset is the \textit{de novo} molecular generation task: generating a compound while optimizing its ``desirability score``.
We focus on optimizing the predicted binding affinity to a target protein, which we measure with docking scores.
To enable the optimization of these docking scores, we need to build an extensive dataset of protein structures with defined pockets (i.e., regions of the protein where generated molecules should bind).

\paragraph{Structure extraction.}
We extracted our receptors (protein pockets) from the SAIR dataset~\citep{sair}, which consists of over 1M protein-ligand pairs from ~5k unique protein structures.
This dataset was built upon the structural prediction of Boltz1~\citep{boltz1} (a cofolding model) and notably contains predicted protein structures for which no known experimental structure exists.
To identify pockets for each structure, we performed a clustering of the protein's residues based on the predicted receptor-ligand structure (see details on the dataset creation in~\autoref{app:gen_data_details})
We completed this set with structures from the SIU dataset~\citep{SIU} (with labeled pockets).
After cleaning all structures with Meeko~\citep{martins_meeko_2025}, we obtained ~4,500 structures with defined pockets, each associated with a unique UniProt ID.
We use AutoDock-GPU~\citep{autododckgpu} to perform the docking simulations.

\paragraph{Objective Design.}
Once all protein structures have been extracted, we build our dataset by creating property-optimization combinations that will define the desirability score, or reward, to optimize.
In addition to docking scores, we included 12 classical molecular properties computed using RDKit~\citep{greg_landrum_2020_3732262}, whose occurrence in our dataset is illustrated in~\autoref{fig:reward_type} and ~\autoref{fig:donut_dataset}.
All docking targets were sampled with equal probability, whereas classical molecular descriptors were sampled with varying probabilities.
Specifically, we over-sampled the synthetic accessibility score (SA)~\citep{ertl_estimation_2009} and the quantitative estimate of drug-likeness (QED)~\citep{bickerton_quantifying_2012} to encourage the model to generate feasible, drug-like molecules.
Finally, we created three splits: a training split of 49k prompts, and test and evaluation splits of 1k prompts, where the set of protein structures in the test, evaluation, and training splits is disjoint.

\subsubsection{Evaluation of the Completions}

\begin{table}
    \centering
    \caption{
        \textbf{Reward and prompt definition.}
        We propose different reward functions $\fullreward$ depending on the objective $\obj$ of the task, that are designed to reflect how close the extracted answer $\extrcompl$ is to the desired outcome.
        All properties (and rewards) are normalized in $[0,1]$.
    }
    \label{tab:all_rewards}
    \begin{tabular}{|c||c|c|c|c|c|c|}
        \toprule
         Objective $\obj$ & Property $\prop$ & Reference $\reference$ & Margin $\marg$ & Answer $\extrcompl$ &  Reward $\fullreward$  \\
        \midrule\midrule
        \multicolumn{5}{|c}{\textbf{\textit{De Novo} Generation} (per-property reward)} & \textbf{49k} \\
        \midrule
           \midrule maximize & property & n/a & threshold & gen. SMILES & $\prop(\extrcompl)$ \\
           \midrule minimize & property & n/a & threshold & gen. SMILES & $1 - \prop(\extrcompl)$ \\
           \midrule below    & property & n/a & threshold & gen. SMILES & $\mathds{1}_{\prop(\extrcompl)\leq \marg}$ \\
           \midrule above    & property & n/a & threshold & gen. SMILES & $\mathds{1}_{\prop(\extrcompl)\geq \marg}$ \\
        \midrule\midrule
        \multicolumn{5}{|c}{\textbf{Property Prediction}} & \textbf{55k} \\
        \midrule
           \midrule regression     & property & molecule & prop. std. & pred. value & $1-\frac{(\extrcompl - \prop(\reference))^2}{\marg^2}$ \\
           \midrule classification & property & molecule & n/a & pred. value & $\mathds{1}_{\extrcompl = \prop(\reference)}$ \\
        \bottomrule
    \end{tabular}
\end{table}

\paragraph{Reward Design.}
Our next step is to define a reward function to evaluate how well a generated completion $\compl$ answers a prompt $q = \{\obj, \prop, \reference, \marg\}_{1\leq i \leq\nprops}$.
We provide in~\autoref{tab:all_rewards} a description of how this reward is computed depending on $\prompt$ for a single property optimization task: $\{\obj, \prop, \reference, \marg\} \quad 1\leq i \leq\nprops$.
To aggregate multiple scores associated with multiple properties into a single reward $\fullreward$, several strategies exist~\citep{LUUKKONEN2023102537}, and we choose to rely on the geometric mean of the rewards associated with each property.
Using the geometric mean compared to an arithmetic mean ensures that only completions effectively optimizing all properties are rewarded, compared to completions that would optimize only a subset of the properties. \footnote{Having a single property reward of $0$ would lead the aggregated reward to be $0$.}

\paragraph{\textit{De Novo} Generation Evaluation.}
Our main objective is to evaluate the ability of a model to explore the chemical space (set of all possible molecules) to find high-reward molecules, which may be particularly challenging when high-reward molecules are rare and difficult to find.
To evaluate this ability, we rely on the $\topk{k}$ score, which measures the average reward of the top-$k$ best unique generated molecules when sampling $n_r \geq k$ completions (or rollouts):
\begin{equation}
    \topk{k}(\model, \prompt) = \mathbb{E}_{
        \{\compl_1,\ldots,\compl_{n_r}\}\sim\model(\prompt)
    }\left[
         \max_{
        \substack{
            \{i_1, \ldots, i_k\}\subset\{1, \ldots, n_r\}
        }
        }{
           \frac{1}{k} \sum_{i\in\{i_1,\ldots i_k\}}{
                \reward(\prompt,\extrcompl_i)\times\mathds{1}_{\extrcompl_i \notin \{\extrcompl_j\}_{1\leq j < i}}
            }
        }
    \right],
\end{equation}
where $\mathds{1}_{\extrcompl_i \notin \{\extrcompl_j\}_{1\leq j < i}}$ ensures that only unique molecules are considered in the top-k selection.
One can notice that $\text{top}_{1, 1}$ score corresponds to the average reward of the model, and $\topk{1}$ corresponds to the pass@$n_r$ score in the LLM reasoning literature.

The $\topk{k}$ score accounts for duplicates and invalid answers, being assigned a reward of $0$, effectively merging the evaluation of the model's ability to generate valid and unique molecules with its ability to generate high-reward molecules.
However, it does not account for the diversity of the proposed molecules, which might be very similar to each other.
This can prove problematic, indeed, in practice, it is often desirable to generate a set of high-reward molecules that are also diverse, to obtain a range of candidates from different chemical series, and to mitigate the risk of failure of a specific chemical series in later stages of drug discovery.

\paragraph{Diversity-aware Top-k Score.}
To address this limitation, we propose to measure the top-k score on molecules with a constraint on their mutual similarity controlled by $s_{\text{max}}$ ($0 < s_{\text{max}} < 1$).
We introduce the diversity-aware top-k score:
\begin{equation}
    \begin{aligned}
        \topk{k}^{\text{diversity}}(\model, \prompt, s_{\text{max}}) &= \mathbb{E}_{
            \{\compl_1,\ldots,\compl_{n_r}\}\sim\model(\prompt)
        }\left[
            \frac{1}{k} {\sum_{
               1\leq i\leq k
            }{\reward(\prompt,\extrcompl^{(d)}_i)}}
        \right],\\
        \text{where} \quad \forall i \in\{1,\ldots, k\},\quad \extrcompl^{(d)}_i &= \argmax_{
                \compl\in\{\compl_0,\ldots,\compl_{n_r}\}
        }{
            \reward(\prompt,\extrcompl) \times \mathds{1}_{
                \max_{
                    j\in\{1,\ldots i-1\}
                }{\left[
                    \text{sim}(\extrcompl_a,\extrcompl^{(d)}_j) \right]< s_\text{max}
                }
            }
        },
    \end{aligned}
\end{equation}
where if not enough molecules satisfy the similarity constraint, we complete the sum with null values.

This metric can be better understood by the following procedure: we want to build a set of $k$ molecules $\{\extrcompl^{(d)}_1, \ldots, \extrcompl^{(d)}_k\}$ such that: (1) the Tanimoto similarity between any two molecules in the set is below $s_{\text{max}}$, and (2) at each step the highest reward molecules available and meeting the similarity constraint is selected.
Note that this does not exactly correspond to solving the combinatorial optimization problem of selecting $k$ molecules such that their pairwise similarities are below $s_{\text{max}}$, and the sum of their rewards is maximized, but rather a greedy approximation of this problem, with the advantage of being less costly and more interpretable (we show examples of selected molecules in~\autoref{appendix:div_aware_viz}.).

Our diversity-aware top-k score measures the trade-off between the ability of the model to generate high-reward molecules and its ability to explore the chemical space.
We will show in the next section that this metric provides additional insights into the performance of the models.

\subsection{Molecular Property prediction.}
We add to our dataset molecular property prediction auxiliary tasks, where the model is tasked to predict a given property for a molecule.
Notably, we will use these tasks to see if the performances of the models on property prediction tasks correlate with their performance on the generation task.

\subsubsection{Data Collection}

We extracted property prediction tasks from the Polaris platform~\citep{wognum_call_2024}, aggregating property prediction benchmarks from academic and industrial research.
We selected 27 benchmarks and datasets, resulting in 52K prompts, primarily focused on ADMET properties (Absorption, Distribution, Metabolism, Excretion, and Toxicity assays).
The vast majority of our tasks are regression tasks (79\% of prompts and 86\% of individual tasks), with the remainder being binary classification tasks (21\% of prompts and 14\% of tasks).
For regression tasks with wide label ranges (e.g., Half-life and VDss), we post-processed labels using a logarithmic transformation.

\subsubsection{Evaluation of the Completions}

\paragraph{Reward Design.}
For property prediction tasks, $\nprops = 1$, and the task is fully defined by: $\objnoi, \propnoi, \referencenoi, \text{ and } \margnoi$, where $\prop$ is the property to predict, $\objnoi$ the type of task (regression or classification), $\referencenoi$ is the molecule to predict the property of, and $\margnoi$ is the standard deviation of the training labels, used to normalize the predicted error for regression tasks.
The reward function definition is described in~\autoref{tab:all_rewards}, where the average reward corresponds to the average accuracy for classification tasks, and the average coefficient of determination $R^2$ for regression tasks (on valid completions).

\paragraph{Evaluation Metrics.}
Evaluating property prediction poses unique challenges: models may fail or refuse to predict a molecular property.
While invalid predictions in classification tasks can be counted as incorrect, regression metrics such as $R^2$ are unbounded, making it unclear how to penalize failures.
To address this issue, we report the Pearson correlation coefficient between the predicted values of the properties and the ground truth label (which lies between $-1$ and $1$), and we adjust the Pearson correlation by rescaling with the fraction of valid predictions made by the model as such:
\(
    \rho_{pearson}^\text{norm} = \pp{\rho_{pearson} + 1} \cdot \frac{n_\text{valid}}{n_\text{total}} - 1
\).

\subsection{Training with the Molecular Verifier}
\label{sec:training}

The reward functions defined in~\autoref{tab:all_rewards} can be computed at generation time through a \textit{Molecular Verifier}, which, given a prompt-completion pair, computes its associated reward by running docking simulations and computing the physico-chemical properties.
This verifier can be used to evaluate pre-trained models, but also to fine-tune language models with reinforcement learning.

We leveraged this verifier to fine-tune the Mistral-Small-4, containing 128B parameters using Group Relative Policy Optimization (GRPO)~\citep{shao2024deepseekmathpushinglimitsmathematical}.
GRPO aims at training a model to maximize the expected reward of its completions by generating a group of completions for each prompt, increasing the probability of completions that outperform the group average.

We denote $\pi_{\theta_{\text{old}}}$ the reference model  (or policy) we use to sample completions and $\pi_{\theta}$ the active model (or policy), which might be different from $\pi_{\theta_{\text{old}}}$ in asynchronous trainings.

For each prompt $\prompt$, we generate $\rollouts$ completions $\{\compl_i\}_{i=1}^\rollouts$ with the policy $\pi_{\theta_{\text{old}}}$, and compute their corresponding rewards $\{r_i\}_{i=1}^\rollouts$, and advantages \(
    \hat{A}_i:= {r_i - \text{mean}(\{r_j\}_{j=1}^\rollouts)}
\) for $1\leq i\leq \rollouts$.
The advantage is then normalized across all sequences in the mini-batch as
\(
    \hat{A}_{i}^{\text{norm}} := ({\hat{A}_i} - {\hat{A}}^{\text{mean}}) / {\hat{A}}^{\text{std}}
\).
The GRPO loss we used for our experiments can then be defined as~\citep{mistralai2025magistral}:
\begin{align*}
\mathcal{J}_{\text{GRPO}}(\theta)
&:=
\mathbb{E}_{\prompt \sim P(Q), \{\compl_i\}_{i=1}^\rollouts \sim \pi_{\theta_{\text{old}}}(\cdot|\prompt)}
\frac{1}{\sum_{i=1}^\rollouts |\compl_i|}\sum_{i=1}^{\rollouts} \sum_{t=1}^{|\compl_i|} \notag \\
& \min\left[\frac{\model(\compl_{i,t}|\prompt, \compl_{i,<t})}{\pi_{\theta_{\text{old}}}(\compl_{i,t}|q, \compl_{i,<t})} \hat{A}_{i,t}^{\text{norm}}, \; \text{clip}(\frac{\model(\compl_{i,t}|\prompt, \compl_{i,<t})}{\pi_{\theta_{\text{old}}}(\compl_{i,t}|\prompt, \compl_{i,<t})}, 1-\varepsilon_{\text{low}}, 1+\varepsilon_{\text{high}})\hat{A}_{i,t}^{\text{norm}}\right],
\end{align*}

In the following section, we display results for 3 models trained for $\molstralnsteps$ steps, all trained with 32 prompts per step.
We heavily focused the training of the models on the molecular generation task, with 66\% of the prompts sampled from the molecular generation task, and 33\% from the molecular property prediction task.
These models are denoted as RL-Molstral-$\rollouts$ where $\rollouts \in \{4,8,16\}$ is the group size used for training (i.e., the number of completions generated per prompt during training). We scale the batch size with the group size to maintain the same number of unique prompts per batch.

\section{Results}
\label{sec:results}

\subsection{De Novo Generation}
\label{sec:results_gen}

\begin{table}
    \caption{
        \textbf{\textit{De novo} generation results.}
        Evaluation of various LLMs on the molecular generation task.
        We report the average top-k scores among $n_r$ generations per prompt and standard deviations, averaged over the unique prompts.
        Models are ordered by their release date.
    }
    \resizebox{\textwidth}{!}{
        \centering
        \begin{tabular}{c|c||p{1.3cm}p{1.3cm}||p{1.3cm}p{1.3cm}||p{1.3cm}p{1.3cm}||p{1.3cm}p{1.3cm}||}
\toprule
 &  & \multicolumn{2}{c||}{$n_r=$16} & \multicolumn{2}{c||}{$n_r=$24} & \multicolumn{2}{c||}{$n_r=$32} & \multicolumn{2}{c||}{$n_r=$48} \\
Model & Size & $\text{top}_{1}$ & $\text{top}_{16}$ & $\text{top}_{1}$ & $\text{top}_{16}$ & $\text{top}_{1}$ & $\text{top}_{16}$ & $\text{top}_{1}$ & $\text{top}_{16}$ \\
\midrule
\midrule
\multicolumn{10}{c}{General LLMs} \\
\midrule
\midrule
gemma-4 & 31B & \textbf{\underline{{\cellcolor[rgb]{0.973, 0.925, 0.922}} 0.65 \tiny $\pm$0.05}} & {\cellcolor[rgb]{0.745, 0.780, 0.847}} 0.37 \tiny $\pm$0.05 & \textbf{\underline{{\cellcolor[rgb]{0.961, 0.902, 0.898}} 0.67 \tiny $\pm$0.04}} & {\cellcolor[rgb]{0.867, 0.875, 0.906}} 0.48 \tiny $\pm$0.03 & \textbf{\underline{{\cellcolor[rgb]{0.953, 0.890, 0.882}} 0.68 \tiny $\pm$0.03}} & \underline{{\cellcolor[rgb]{0.910, 0.910, 0.929}} 0.52 \tiny $\pm$0.02} & \underline{{\cellcolor[rgb]{0.945, 0.871, 0.867}} 0.7 \tiny $\pm$0.01} & \underline{{\cellcolor[rgb]{0.957, 0.949, 0.953}} 0.56 \tiny $\pm$0.01} \\
Mistral-Small-4 & 128B & {\cellcolor[rgb]{0.980, 0.961, 0.957}} 0.6 \tiny $\pm$0.06 & {\cellcolor[rgb]{0.690, 0.737, 0.824}} 0.32 \tiny $\pm$0.04 & {\cellcolor[rgb]{0.980, 0.949, 0.945}} 0.62 \tiny $\pm$0.05 & {\cellcolor[rgb]{0.816, 0.835, 0.878}} 0.44 \tiny $\pm$0.03 & {\cellcolor[rgb]{0.973, 0.929, 0.925}} 0.64 \tiny $\pm$0.04 & {\cellcolor[rgb]{0.867, 0.875, 0.906}} 0.48 \tiny $\pm$0.02 & {\cellcolor[rgb]{0.965, 0.914, 0.910}} 0.66 \tiny $\pm$0.02 & {\cellcolor[rgb]{0.906, 0.906, 0.925}} 0.52 \tiny $\pm$0.01 \\
MiniMax-M2 & 229B & {\cellcolor[rgb]{0.980, 0.957, 0.953}} 0.61 \tiny $\pm$0.06 & {\cellcolor[rgb]{0.682, 0.733, 0.820}} 0.31 \tiny $\pm$0.05 & {\cellcolor[rgb]{0.976, 0.937, 0.933}} 0.64 \tiny $\pm$0.05 & {\cellcolor[rgb]{0.808, 0.827, 0.875}} 0.43 \tiny $\pm$0.03 & {\cellcolor[rgb]{0.965, 0.914, 0.910}} 0.66 \tiny $\pm$0.04 & {\cellcolor[rgb]{0.867, 0.875, 0.906}} 0.48 \tiny $\pm$0.02 & {\cellcolor[rgb]{0.953, 0.890, 0.882}} 0.68 \tiny $\pm$0.02 & {\cellcolor[rgb]{0.918, 0.918, 0.933}} 0.52 \tiny $\pm$0.01 \\
Qwen3-Next & 80B & {\cellcolor[rgb]{0.969, 0.957, 0.957}} 0.58 \tiny $\pm$0.05 & {\cellcolor[rgb]{0.671, 0.725, 0.816}} 0.3 \tiny $\pm$0.04 & {\cellcolor[rgb]{0.976, 0.961, 0.957}} 0.6 \tiny $\pm$0.04 & {\cellcolor[rgb]{0.784, 0.808, 0.863}} 0.41 \tiny $\pm$0.03 & {\cellcolor[rgb]{0.980, 0.957, 0.953}} 0.61 \tiny $\pm$0.03 & {\cellcolor[rgb]{0.827, 0.843, 0.886}} 0.45 \tiny $\pm$0.02 & {\cellcolor[rgb]{0.980, 0.945, 0.941}} 0.63 \tiny $\pm$0.02 & {\cellcolor[rgb]{0.878, 0.886, 0.910}} 0.49 \tiny $\pm$0.01 \\
gpt-oss & 120B & {\cellcolor[rgb]{0.965, 0.953, 0.957}} 0.57 \tiny $\pm$0.05 & {\cellcolor[rgb]{0.600, 0.675, 0.792}} 0.23 \tiny $\pm$0.05 & {\cellcolor[rgb]{0.976, 0.961, 0.957}} 0.59 \tiny $\pm$0.04 & {\cellcolor[rgb]{0.706, 0.753, 0.831}} 0.34 \tiny $\pm$0.05 & {\cellcolor[rgb]{0.980, 0.961, 0.957}} 0.6 \tiny $\pm$0.03 & {\cellcolor[rgb]{0.788, 0.816, 0.867}} 0.42 \tiny $\pm$0.04 & {\cellcolor[rgb]{0.980, 0.953, 0.949}} 0.62 \tiny $\pm$0.02 & {\cellcolor[rgb]{0.878, 0.886, 0.910}} 0.49 \tiny $\pm$0.01 \\
Qwen3 & 30B & {\cellcolor[rgb]{0.976, 0.961, 0.957}} 0.6 \tiny $\pm$0.05 & {\cellcolor[rgb]{0.694, 0.745, 0.824}} 0.33 \tiny $\pm$0.04 & {\cellcolor[rgb]{0.980, 0.953, 0.949}} 0.62 \tiny $\pm$0.04 & {\cellcolor[rgb]{0.808, 0.827, 0.875}} 0.43 \tiny $\pm$0.03 & {\cellcolor[rgb]{0.980, 0.945, 0.941}} 0.63 \tiny $\pm$0.04 & {\cellcolor[rgb]{0.859, 0.871, 0.902}} 0.47 \tiny $\pm$0.02 & {\cellcolor[rgb]{0.973, 0.925, 0.922}} 0.65 \tiny $\pm$0.02 & {\cellcolor[rgb]{0.906, 0.906, 0.925}} 0.51 \tiny $\pm$0.01 \\
Llama-3.3 & 70B & {\cellcolor[rgb]{0.953, 0.945, 0.949}} 0.56 \tiny $\pm$0.03 & {\cellcolor[rgb]{0.537, 0.631, 0.773}} 0.16 \tiny $\pm$0.03 & {\cellcolor[rgb]{0.961, 0.949, 0.957}} 0.57 \tiny $\pm$0.02 & {\cellcolor[rgb]{0.569, 0.655, 0.780}} 0.2 \tiny $\pm$0.03 & {\cellcolor[rgb]{0.969, 0.957, 0.957}} 0.57 \tiny $\pm$0.02 & {\cellcolor[rgb]{0.600, 0.675, 0.792}} 0.23 \tiny $\pm$0.03 & {\cellcolor[rgb]{0.976, 0.961, 0.957}} 0.59 \tiny $\pm$0.01 & {\cellcolor[rgb]{0.631, 0.698, 0.800}} 0.27 \tiny $\pm$0.02 \\
gemma-3 & 27B & {\cellcolor[rgb]{0.953, 0.945, 0.949}} 0.56 \tiny $\pm$0.05 & {\cellcolor[rgb]{0.627, 0.694, 0.800}} 0.26 \tiny $\pm$0.04 & {\cellcolor[rgb]{0.965, 0.953, 0.957}} 0.57 \tiny $\pm$0.04 & {\cellcolor[rgb]{0.725, 0.769, 0.839}} 0.36 \tiny $\pm$0.04 & {\cellcolor[rgb]{0.976, 0.961, 0.957}} 0.59 \tiny $\pm$0.03 & {\cellcolor[rgb]{0.796, 0.820, 0.871}} 0.42 \tiny $\pm$0.03 & {\cellcolor[rgb]{0.980, 0.961, 0.957}} 0.6 \tiny $\pm$0.02 & {\cellcolor[rgb]{0.855, 0.863, 0.898}} 0.47 \tiny $\pm$0.01 \\
R1-Llama & 70B & {\cellcolor[rgb]{0.953, 0.945, 0.949}} 0.56 \tiny $\pm$0.05 & {\cellcolor[rgb]{0.627, 0.694, 0.800}} 0.26 \tiny $\pm$0.04 & {\cellcolor[rgb]{0.969, 0.957, 0.957}} 0.58 \tiny $\pm$0.04 & {\cellcolor[rgb]{0.733, 0.773, 0.839}} 0.37 \tiny $\pm$0.04 & {\cellcolor[rgb]{0.976, 0.961, 0.957}} 0.59 \tiny $\pm$0.04 & {\cellcolor[rgb]{0.796, 0.820, 0.871}} 0.42 \tiny $\pm$0.02 & {\cellcolor[rgb]{0.980, 0.957, 0.953}} 0.61 \tiny $\pm$0.02 & {\cellcolor[rgb]{0.855, 0.863, 0.898}} 0.47 \tiny $\pm$0.01 \\
R1-Qwen & 32B & {\cellcolor[rgb]{0.898, 0.902, 0.922}} 0.51 \tiny $\pm$0.06 & {\cellcolor[rgb]{0.569, 0.655, 0.780}} 0.2 \tiny $\pm$0.04 & {\cellcolor[rgb]{0.925, 0.922, 0.937}} 0.53 \tiny $\pm$0.05 & {\cellcolor[rgb]{0.659, 0.718, 0.812}} 0.29 \tiny $\pm$0.04 & {\cellcolor[rgb]{0.941, 0.933, 0.945}} 0.55 \tiny $\pm$0.04 & {\cellcolor[rgb]{0.725, 0.769, 0.839}} 0.36 \tiny $\pm$0.03 & {\cellcolor[rgb]{0.961, 0.949, 0.957}} 0.57 \tiny $\pm$0.03 & {\cellcolor[rgb]{0.788, 0.816, 0.867}} 0.41 \tiny $\pm$0.01 \\
\midrule
\midrule
\multicolumn{10}{c}{Chemically specialized LLMs} \\
\midrule
\midrule
ether0 & 24B & {\cellcolor[rgb]{0.949, 0.941, 0.949}} 0.55 \tiny $\pm$0.07 & {\cellcolor[rgb]{0.588, 0.667, 0.784}} 0.22 \tiny $\pm$0.04 & {\cellcolor[rgb]{0.969, 0.957, 0.957}} 0.58 \tiny $\pm$0.06 & {\cellcolor[rgb]{0.671, 0.725, 0.816}} 0.31 \tiny $\pm$0.04 & {\cellcolor[rgb]{0.976, 0.961, 0.957}} 0.59 \tiny $\pm$0.05 & {\cellcolor[rgb]{0.725, 0.769, 0.839}} 0.36 \tiny $\pm$0.03 & {\cellcolor[rgb]{0.980, 0.949, 0.945}} 0.62 \tiny $\pm$0.03 & {\cellcolor[rgb]{0.796, 0.820, 0.871}} 0.42 \tiny $\pm$0.02 \\
ChemDFM-R & 14B & {\cellcolor[rgb]{0.980, 0.953, 0.949}} 0.62 \tiny $\pm$0.07 & {\cellcolor[rgb]{0.671, 0.725, 0.816}} 0.31 \tiny $\pm$0.04 & {\cellcolor[rgb]{0.973, 0.925, 0.922}} 0.65 \tiny $\pm$0.05 & {\cellcolor[rgb]{0.784, 0.808, 0.863}} 0.41 \tiny $\pm$0.03 & {\cellcolor[rgb]{0.965, 0.910, 0.902}} 0.66 \tiny $\pm$0.05 & {\cellcolor[rgb]{0.835, 0.851, 0.886}} 0.46 \tiny $\pm$0.03 & {\cellcolor[rgb]{0.949, 0.878, 0.871}} 0.69 \tiny $\pm$0.02 & {\cellcolor[rgb]{0.894, 0.894, 0.918}} 0.5 \tiny $\pm$0.02 \\
ChemDFM-v2.0 & 14B & {\cellcolor[rgb]{0.980, 0.949, 0.945}} 0.63 \tiny $\pm$0.08 & {\cellcolor[rgb]{0.659, 0.718, 0.812}} 0.29 \tiny $\pm$0.05 & \underline{{\cellcolor[rgb]{0.969, 0.922, 0.914}} 0.66 \tiny $\pm$0.06} & {\cellcolor[rgb]{0.773, 0.800, 0.859}} 0.4 \tiny $\pm$0.04 & \underline{{\cellcolor[rgb]{0.957, 0.894, 0.890}} 0.68 \tiny $\pm$0.05} & {\cellcolor[rgb]{0.827, 0.843, 0.886}} 0.45 \tiny $\pm$0.03 & \textbf{\underline{{\cellcolor[rgb]{0.941, 0.863, 0.859}} 0.7 \tiny $\pm$0.03}} & {\cellcolor[rgb]{0.894, 0.894, 0.918}} 0.5 \tiny $\pm$0.02 \\
\midrule
\midrule
\multicolumn{10}{c}{Models trained on MolRGen} \\
\midrule
\midrule
RL-Molstral-g4 & 128B & {\cellcolor[rgb]{0.976, 0.961, 0.957}} 0.59 \tiny $\pm$0.05 & {\cellcolor[rgb]{0.745, 0.780, 0.847}} 0.37 \tiny $\pm$0.04 & {\cellcolor[rgb]{0.980, 0.953, 0.949}} 0.62 \tiny $\pm$0.04 & {\cellcolor[rgb]{0.847, 0.859, 0.894}} 0.47 \tiny $\pm$0.02 & {\cellcolor[rgb]{0.980, 0.945, 0.941}} 0.63 \tiny $\pm$0.03 & {\cellcolor[rgb]{0.878, 0.886, 0.910}} 0.49 \tiny $\pm$0.02 & {\cellcolor[rgb]{0.973, 0.929, 0.925}} 0.65 \tiny $\pm$0.02 & {\cellcolor[rgb]{0.918, 0.918, 0.933}} 0.53 \tiny $\pm$0.01 \\
RL-Molstral-g8 & 128B & {\cellcolor[rgb]{0.976, 0.961, 0.957}} 0.59 \tiny $\pm$0.04 & \underline{{\cellcolor[rgb]{0.784, 0.808, 0.863}} 0.41 \tiny $\pm$0.03} & {\cellcolor[rgb]{0.980, 0.957, 0.953}} 0.61 \tiny $\pm$0.04 & \underline{{\cellcolor[rgb]{0.867, 0.875, 0.906}} 0.48 \tiny $\pm$0.02} & {\cellcolor[rgb]{0.980, 0.949, 0.945}} 0.62 \tiny $\pm$0.03 & {\cellcolor[rgb]{0.898, 0.902, 0.922}} 0.51 \tiny $\pm$0.01 & {\cellcolor[rgb]{0.976, 0.937, 0.933}} 0.64 \tiny $\pm$0.02 & {\cellcolor[rgb]{0.925, 0.922, 0.937}} 0.53 \tiny $\pm$0.01 \\
RL-Molstral-g16 & 128B & \underline{{\cellcolor[rgb]{0.976, 0.937, 0.933}} 0.64 \tiny $\pm$0.05} & \textbf{\underline{{\cellcolor[rgb]{0.820, 0.839, 0.882}} 0.45 \tiny $\pm$0.04}} & {\cellcolor[rgb]{0.969, 0.922, 0.914}} 0.66 \tiny $\pm$0.04 & \textbf{\underline{{\cellcolor[rgb]{0.918, 0.918, 0.933}} 0.53 \tiny $\pm$0.02}} & {\cellcolor[rgb]{0.961, 0.902, 0.898}} 0.67 \tiny $\pm$0.03 & \textbf{\underline{{\cellcolor[rgb]{0.949, 0.941, 0.949}} 0.55 \tiny $\pm$0.01}} & {\cellcolor[rgb]{0.949, 0.878, 0.871}} 0.69 \tiny $\pm$0.02 & \textbf{\underline{{\cellcolor[rgb]{0.965, 0.953, 0.957}} 0.57 \tiny $\pm$0.01}} \\
\bottomrule
\end{tabular}

    }
    \label{tab:gen_results}
\end{table}

\paragraph{Baseline models.}
We evaluate a diverse set of models across two categories: (1) \textbf{General-purpose LLMs} including Qwen3-30B-A3B-Thinking and Qwen3-Next-80B-A3B-Thinking~\citep{qwen3technicalreport}, DeepSeek-R1 (distilled into Llama-70B and Qwen-32B backbones for reasonable experimental cost)~\citep{deepseekai2025deepseekr1incentivizingreasoningcapability}, Gemma-3-27B~\citep{gemma_2025}, Gemma-4-31B~\citep{gemma_2026}, MiniMax-M2~\citep{minimax2025minimaxm1scalingtesttimecompute}, GPT-OSS~\citep{gptoss} and Llama-3.3-70B-Instruct~\citep{grattafiori2024llama3herdmodels}, and (2) \textbf{Chemically-specialized models} including ChemDFM-v2.0, ChemDFM-R~\citep{Zhao_2025}, and Ether0~\citep{ether0}.

\paragraph{Top-k performances of pretrained LLMs.}
\autoref{tab:gen_results} presents the $\topk{k}$ performances on the de novo molecular generation task.
Among all evaluated models, Gemma-4 achieves the best overall performance in terms of $\topk{k}$ scores, outperforming both other general-purpose LLMs and chemically specialized models.
In general, the most recent models tend to perform better than older ones, and we measured a Spearman correlation of $0.8$ between the release date and the $\topk{k}$ score on average (correlation between the rankings induced by the two variables).
This suggests that improvements in other domains, such as code generation and mathematical reasoning, do transfer to some extent to molecular generation.
Conversely, we measured a lower Spearman correlation of $0.2$ between the number of parameters and the $\topk{k}$ score on average, and indeed, Gemma-4 with only 31B parameters outperforms significantly larger models exceeding 100B parameters.

\begin{figure}
    \centering
    \includegraphics[width=0.9\linewidth]{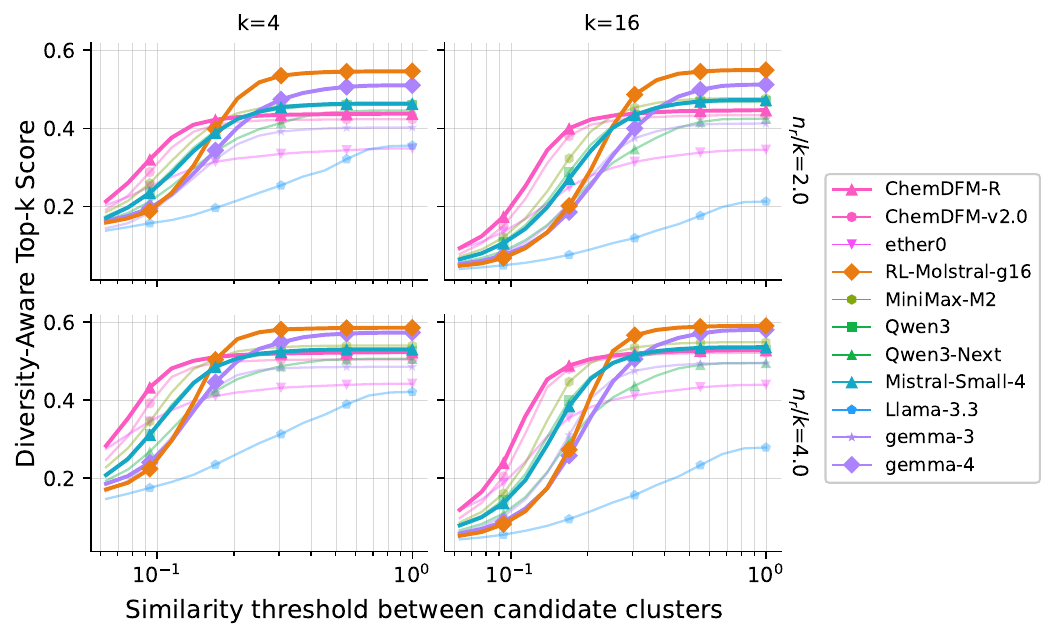}
    \caption{
        \textbf{Diversity-aware top-k score.}
        Evaluation of the diversity-aware top-k score (y-axis) against varying similarity thresholds (x-axis) between candidate clusters.
        As the similarity threshold increases, the diversity constraint relaxes, and the average score of the selected molecule pool increases.
    }
    \label{fig:diversity_reward}
\end{figure}

\paragraph{Diversity-aware top-k performances.}
\autoref{fig:diversity_reward} presents the diversity-aware top-k scores of various LLMs on the de novo molecular generation task, evaluated across a range of similarity thresholds $s_\text{max}$.
As the similarity threshold increases, the diversity-aware score increases, as expected: high thresholds (e.g., $s_\text{max} = 0.9$) impose weak diversity constraints, as two molecules are only considered to be too similar if their Tanimoto similarity is very high.
On the contrary, low thresholds (e.g., $s_\text{max}  \approx 0$) enforce strong constraints where a molecule needs to be very different from all other selected molecules to be considered for the top-k score.

The diversity-aware evaluation provides additional insights compared to the raw $\topk{k}$ score.
While Gemma-4 (the best model according to the top-k score) performs the best under most threshold values ($s_\text{max} \geq 0.3$), the chemically-specialized ChemDFM-R overtakes it under stronger constraints.
This suggests a qualitative difference in generation strategies: while Gemma-4 performs best overall, ChemDFM-R demonstrates superior capacity when generating a set of highly diverse high-reward molecules (low pairwise similarity).

\textbf{Effect of GRPO training.}
As explained in~\autoref{sec:training}, to show that our dataset and verifier can be leveraged to train LLMs for molecular generation, we fine-tuned Mistral-Small-4 with Group Relative Policy Optimization (GRPO) with 3 different group sizes $\rollouts \in \{4,8,16\}$.
As explained in~\autoref{sec:dataset}, all docking targets in the test set are excluded from the training set of our models.

\begin{wrapfigure}[26]{r}{0.41\textwidth}
    \centering
    \vspace{-0.7cm}
    \includegraphics[width=1\linewidth]{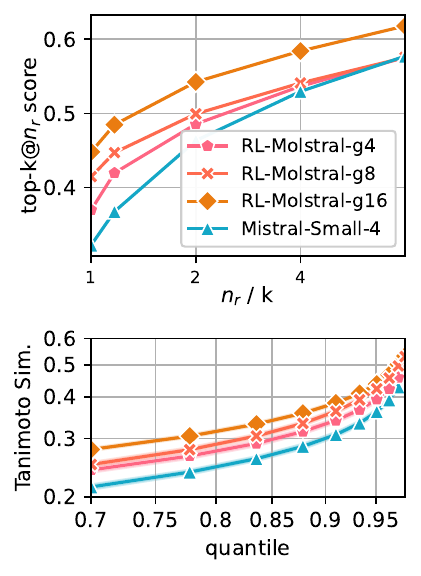}
    \caption{
        \textbf{GRPO Training Effect.}
        \textbf{(Top)}
        $\topk{k}$ scores across varying sampling scales $n_r/k$.
        \textbf{(Bottom)}
        Distribution of pairwise Tanimoto similarities for molecules generated for the same prompt.
    }
    \label{fig:grpo_effect}
\end{wrapfigure}

\autoref{fig:grpo_effect} compares the performance of RL-Molstral against the base model using the $\topk{k}$ score across varying sampling budgets $n_r$ where $n_r$ is scaled proportionally to $k\in\{2,4,8\}$.
The models trained with GRPO exhibit a significant improvement in $\topk{k}$ scores at low sampling regimes across all group sizes, and the larger the group size, the higher the improvement.

While the models trained with a group size of 4 and 8 see their improvements over the base model diminish as the sampling budget increases, the model trained with a group size of 16 maintains its performance advantage across all sampling budgets.
However, even for RL-Molstral-g16, this performance improvement comes at the cost of a decrease in the diversity of the generated molecules, as shown by the distribution of pairwise Tanimoto similarities between molecules generated from the same prompt in ~\autoref{fig:grpo_effect}, with larger group sizes leading to lower diversity scores.
This phenomenon aligns with recent findings~\citep{yue2025limit-of-rlvr}, highlighting the following tradeoff: RL improves exploitation but harms exploration.

Despite the decrease in diversity, RL-Molstral-g16 still outperforms the base model for diversity constraints up to $s_\text{max} \geq 0.3$, as shown in~\autoref{fig:diversity_reward}, outperforming all other models in this regime, notably Gemma-4, outperforming this model over all similarity thresholds.

\subsection{Molecular Property Prediction}
\label{sec:results_proppred}
\begin{wrapfigure}[31]{r}{0.4\textwidth}
        \centering
        \includegraphics[width=1\linewidth]{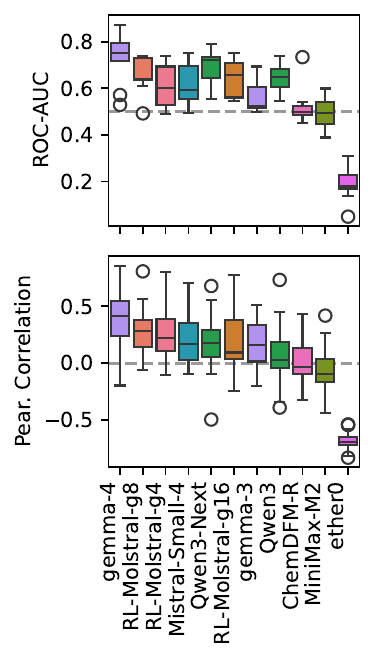}
    \caption{
            \textbf{Property prediction performances.}
            ROC-AUC scores of the LLMs on classification tasks (top), and normalized Pearson correlation on regression tasks (bottom).
    }
    \label{fig:mol_proppred_combined}
\end{wrapfigure}

Finally, we evaluate the performance of the models on the auxiliary molecular property prediction task.
We report in ~\autoref{fig:mol_proppred_combined} the performance on molecular property prediction tasks, separated between regression and classification tasks.
For readability, we only report a subset of the models evaluated, focusing on the best-performing general-purpose LLMs (MiniMax-M2, Qwen3-Next-80B-A3B-Thinking, gemma-3) and chemically-specialized models.

\paragraph{Global Results.}
Overall, on regression tasks, most models struggle to perform better than random, struggling to achieve normalized Pearson correlations above 0.
On classification tasks, most models perform better than random with ROC-AUC scores above 0.5, but these performances remain low compared to specialized models for these tasks.
Finally, RL-Molstral's training was heavily focused on the generation task, and as such, the improvement over Mistral-Small-4 on the molecular property prediction tasks is limited, improving the performance of the base model on 60\% of the tasks (see details in ~\autoref{app:molstral-propred}).

The performances of the LLMs on these property prediction tasks align with the performances on the generation tasks, with Spearman correlations of 0.57 and 0.75 between the rankings of the models on the generation task and the classification and regression tasks, respectively, highlighting the relevance of property prediction as an auxiliary task to train models for molecular generation.
Overall, we believe these results highlight well the challenge of molecular property prediction with LLMs and, in particular, for regression tasks, as all models are outperformed by specialized machine learning models~\citep{wognum_call_2024}.

\section{Summary and Concluding Remarks}

In this work, we introduced \benchname, an open-source benchmark and verifier for training and evaluating reasoning-based LLMs on \textit{de novo} molecular generation, covering approximately 4,500 protein targets and molecular property prediction tasks.
Our evaluation of 13 open-source LLMs reveals that relying solely on the raw $\topk{k}$ score to compare models does not capture the full picture of their performance, and while Gemma-4 achieves the best $\topk{k}$ scores among pretrained LLMs, it is outperformed by chemically specialized models under strict diversity constraints.

Fine-tuning Mistral-Small-4 with GRPO demonstrates that \benchname~can serve as an effective training signal: RL-Molstral consistently outperforms the base model in terms of $\topk{k}$ scores and with mild diversity constraints when evaluating the diversity-aware top-k score.
However, RL training consistently reduces chemical diversity, resulting in diminishing returns with strict diversity constraints, motivating the development and evaluation of RL training strategies that explicitly balance exploitation and exploration.

\paragraph{Limitations and future work.}
Despite our efforts to design a realistic and informative benchmark, there are still several limitations that could be explored in future work.
First, our benchmark relies on docking scores as a proxy for binding affinity, which is a common practice to evaluate molecular generation methods, which are known to be imperfect predictors of experimental binding affinities, and there can be significant discrepancies between computational predictions and real-world outcomes.
Additionally, the LLMs in our experiments receive limited information about target protein structures: only textual descriptions, and they do not have direct access to 3D structural coordinates, binding pocket geometries, or other detailed biophysical information.
Providing richer structural context (e.g., via encoded 3D representations or direct access to the PDB files) could substantially improve model performance and is an important direction for future work.

We hope \benchname~will serve as a starting point for addressing these limitations and advancing the development of reasoning-based LLMs for molecular generation, ultimately bridging the gap between computational predictions and practical drug discovery applications.

\clearpage

\bibliography{bibliography}
\bibliographystyle{plainnat}

\newpage
\appendix

\section{Molecular Generation Data Creation}
\label{app:gen_data_details}

At a high level, the dataset-generation process:
\begin{enumerate}
  \item Loads a set of property definitions, docking targets, and pocket metadata.
  \item Uses a rule-based prompt generator to sample multi-objective molecular-generation prompts.
  \item Stores the prompts and their metadata in two formats: a JSONL file and a HuggingFace dataset.
\end{enumerate}

\subsection{Per-prompt sampling loop (inner generator)}
For each prompt, the following steps are executed:

\begin{itemize}
  \item \textbf{Property Selection:} We sample the number of properties to optimize, $n_{\text{props}}$, and then select $n_{\text{props}}$ properties from the pool of standard and docking properties. The selection is weighted by the relative frequency of each property (see \autoref{tab:std_props_sampling} for standard properties, and uniform for docking properties).
  \item \textbf{Objective Assignment:} For each selected property, an objective is sampled. Docking properties use predefined objective distributions (e.g., minimize, below, above), while standard properties use distributions described in \autoref{tab:std_props_sampling}.
  \item \textbf{Rule check and yielding:} We use a RuleSet to keep checks on whether the generated prompt is allowed under current rules (notable: no duplicates, per-property occurrence counts haven't exceeded 4, etc).
\end{itemize}

\begin{table}
\centering
\caption{Allowed objectives for molecular properties, with their associated selection probability. We also include the relative frequency, defining the probability of selecting a given property (higher values, such as QED or logP, are sampled more often).}
\label{tab:std_props_sampling}
\begin{tabular}{|l||l|l|l|}
\hline
\multirow{2}{*}{Property} & \multicolumn{2}{c|}{Allowed Objectives} & \multirow{2}{*}{Relative Frequency} \\

                                   & Objectives             &  Proba.  \\
\midrule
SA                               & minimize                          & 0.5                               & \multirow{2}{*}{3.0} \\
                                 & below                             & 0.5                               & \\
\midrule
QED                              & maximize                          & 0.7                               & \multirow{2}{*}{7.0} \\
                                 & above                             & 0.3                               & \\
\midrule
ExactMolWt                       & above                             & 0.3                               & \multirow{2}{*}{0.8} \\
                                 & below                             & 0.7                               & \\
\midrule
NumAromaticRings                 & above                             & 0.2                               & \multirow{3}{*}{0.5} \\
                                 & below                             & 0.5                               & \\
                                 & minimize                          & 0.3                               & \\
\midrule
NumHBA                           & maximize                          & 0.5                               & \multirow{2}{*}{0.5} \\
                                 & above                             & 0.5                               & \\
\midrule
NumHBD                           & above                             & 0.5                               & \multirow{2}{*}{0.5} \\
                                 & below                             & 0.5                               & \\
\midrule
NumRotatableBonds                & above                             & 0.5                               & \multirow{2}{*}{0.3} \\
                                 & below                             & 0.5                               & \\
\midrule
FractionCSP3                     & above                             & 0.5                               & \multirow{2}{*}{0.5} \\
                                 & below                             & 0.5                               & \\
\midrule
TPSA                             & above                             & 0.3                               & \multirow{2}{*}{0.6} \\
                                 & below                             & 0.7                               & \\
\midrule
HallKierAlpha                    & above                             & 0.5                               & \multirow{2}{*}{0.3} \\
                                 & maximize                          & 0.5                               & \\
\midrule
Phi                              & maximize                          & 0.4                               & \multirow{4}{*}{0.5} \\
                                 & above                             & 0.4                               & \\
                                 & below                             & 0.1                               & \\
                                 & minimize                          & 0.1                               & \\
\midrule
logP                             & maximize                          & 0.1                               & \multirow{4}{*}{2} \\
                                 & above                             & 0.1                               & \\
                                 & below                             & 0.4                               & \\
                                 & minimize                          & 0.4                               & \\
\bottomrule
\end{tabular}
\end{table}

\subsection{Pocket extraction from the SAIR dataset}
\label{sec:sair_pocket_extraction}

This section describes the computational pipeline we used to extract protein binding pockets from the SAIR structural dataset.
We aim to identify consistent binding pockets for protein sequences given multiple ligand-receptor structures in the SAIR dataset.
The pipeline (1) restricts structures to high-quality ligand poses, (2) extracts the local set of residues in contact with the ligand in each structure, (3) aggregates pockets across multiple structures for the same sequence using IoU-based clustering, and (4) selects a representative conformation for each aggregated pocket by minimizing pairwise RMSD across concerned residues.

\paragraph{Filtering and quality control}
Before pocket extraction, we apply two filters to reduce noise and focus on informative examples:
\begin{enumerate}
  \item Per-sequence potency filter: for each protein sequence, keep only structures whose measured ligand potency (\texttt{pIC50}) is in the top 50\% for that sequence.
  \item Per-sequence confidence filter: among the retained structures, keep only those with a confidence score in the top 50\% of the retained set.
\end{enumerate}
This double-filter yields, for each sequence, a subset of CIF files whose ligands are both potent and associated with high-confidence measurements; these files form the input for pocket detection.

\paragraph{Pocket identification per CIF}
For each selected CIF file, the pipeline locates the ligand and defines a local pocket as the set of residues closest to ligand atoms. Concretely: for each ligand atom, compute distances to all protein atom coordinates (atoms whose residue ID flag equals the blank flag for ``standard residues''). Select the \emph{top-k} closest residues for each ligand atom ($k=3$). The union of these residues across all ligand atoms forms the pocket residue set for that CIF.

\paragraph{Aggregation across conformations (IoU clustering)}
Many sequences have multiple co-structure conformations.
To obtain robust pocket definitions, we cluster pockets computed from multiple CIF files for the same sequence using an IoU similarity metric:
\begin{itemize}
  \item We compute an IoU matrix between all pocket residue sets (intersection size divided by union size).
  \item We apply hierarchical clustering with single linkage on the condensed distance matrix and cut the dendrogram with a threshold derived from the IoU cutoff.
  \item For each resulting cluster, we compute an aggregated residue list by selecting residues that appear in at least 70\% of the cluster's member pockets. Clusters that do not aggregate to at least one residue are ignored.
\end{itemize}
The clustering step groups pockets that correspond to the same binding site across different structures while ignoring spurious or highly divergent conformations.

\paragraph{Representative conformation selection}
For each aggregated pocket cluster with more than one member, we select a ``best conformation'' as follows:
\begin{enumerate}
  \item For each member structure, extract atomic coordinates for the residues in the aggregated pocket.
  \item Compute pairwise RMSD values (using Biopython) between all structures restricted to the pocket residues.
  \item Aggregate pairwise RMSD values into a matrix and select the structure with the smallest mean RMSD relative to the others as the best conformation.
\end{enumerate}
The chosen structure is then written as a PDB file (ligand removed).

\begin{figure}
    \centering
    \begin{subfigure}{0.45\textwidth}
        \includegraphics[width=\linewidth]{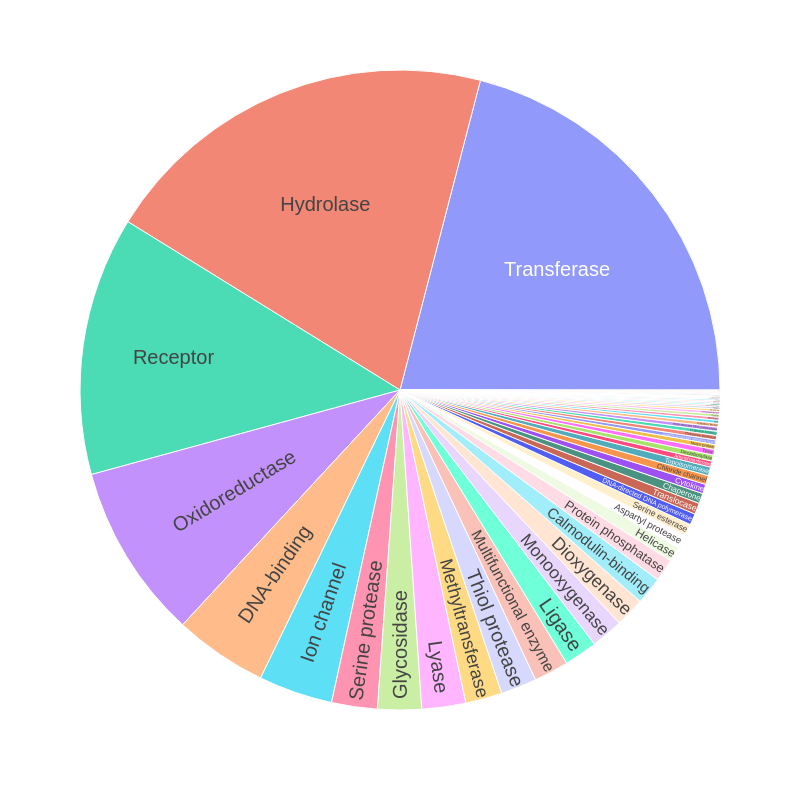}
        \label{fig:organism}
        \caption{}
    \end{subfigure}
    \begin{subfigure}{0.45\textwidth}
        \includegraphics[width=\linewidth]{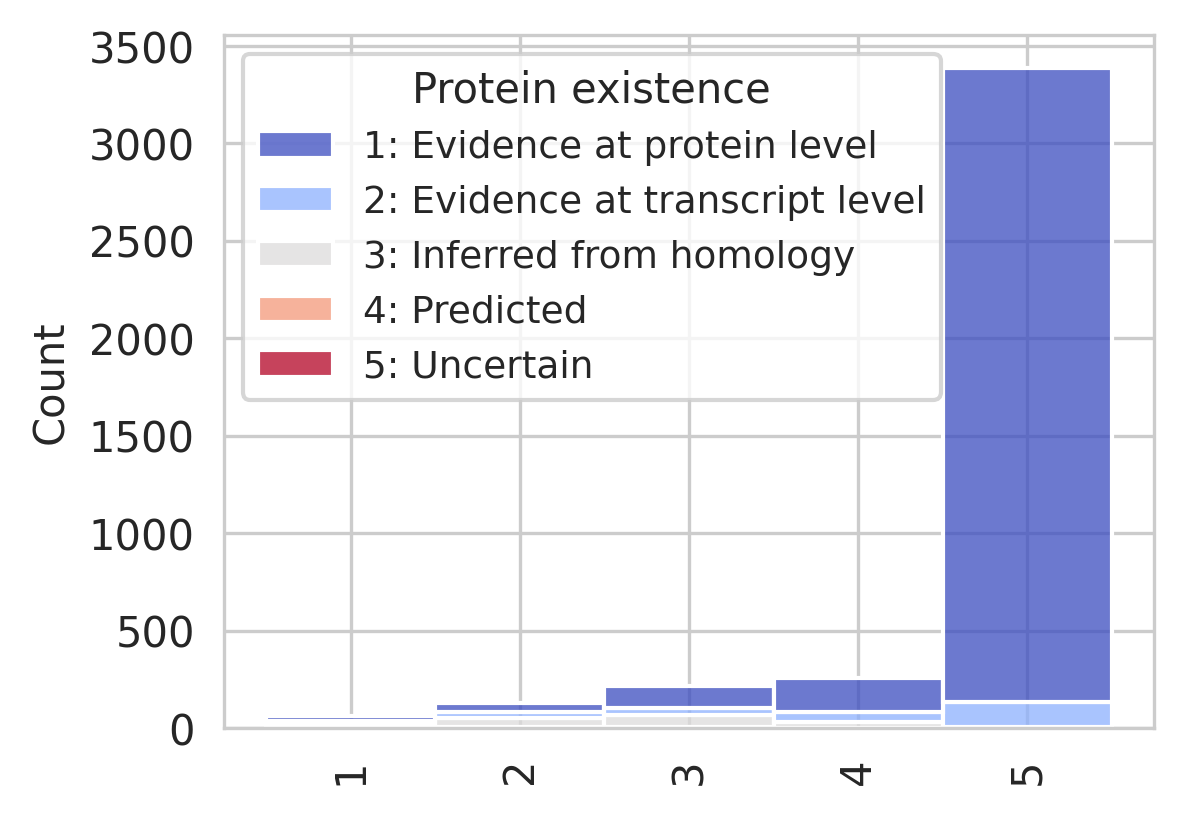}
        \label{fig:mol_function}
        \caption{}
    \end{subfigure}
    \caption{
    \textbf{Overview of the target proteins.} (a) Function of the proteins extracted from the PDB, our dataset comprises 21 molecular functions with at least 10 targets, the majority of which are kinases (30\%).
        (b) Annotation score of the proteins on UniProt (from 1 to 5). The vast majority of the target proteins are high-quality proteins with strong evidence of their existence.
    }
    \label{fig:ovw_prots}
\end{figure}

\subsection{Analysis of the extracted pockets}
\label{app:analysis_extracted_pockets}

\begin{figure}
    \centering
    \includegraphics[width=0.8\linewidth]{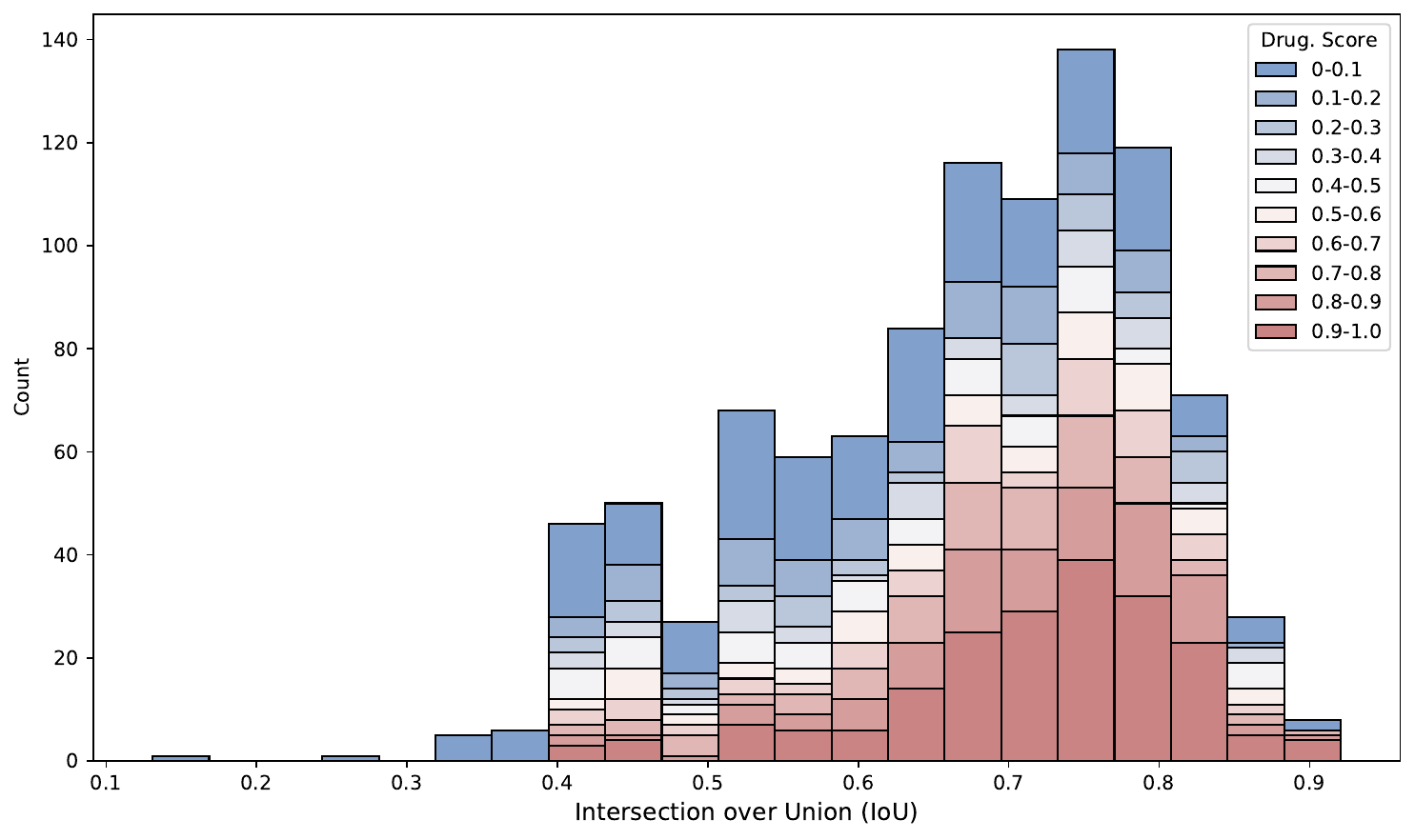}
    \caption{
    \textbf{Assessment of our pocket detection strategy compared to fpocket predictions.}
    Histogram of the Intersection over Union (IoU) between the pockets we extracted and the pockets predicted by fpocket, for 1000 targets.
    We show the predicted druggability score of the pockets with fpocket, with more than half of the pockets having a druggability score above 0.5 (considered druggable by fpocket).
    }
    \label{fig:fpocket_results}
\end{figure}

To assess our pocket detection strategy, we compared the pockets extracted from our pipeline with the predictions of \texttt{fpocket}~\citep{discngine_fpocket}.
We ran \texttt{fpocket} on our targets and computed the Intersection over Union (IoU) between all pockets found with \texttt{fpocket} and our detected pocket.
As illustrated in \autoref{fig:fpocket_results}, the vast majority of our extracted pockets overlap significantly with the pockets identified by \texttt{fpocket}.
In addition, we checked the druggability score of the matching pockets from \texttt{fpocket} and found that more than half of our pockets correspond to \texttt{fpocket} predictions with a druggability score over $0.5$.

\FloatBarrier

\section{Molecular Property Prediction Data Creation}
\label{app:mol_prop_data_details}

\subsection{Task distribution}

\begin{figure}
    \centering
    \includegraphics[width=1\linewidth]{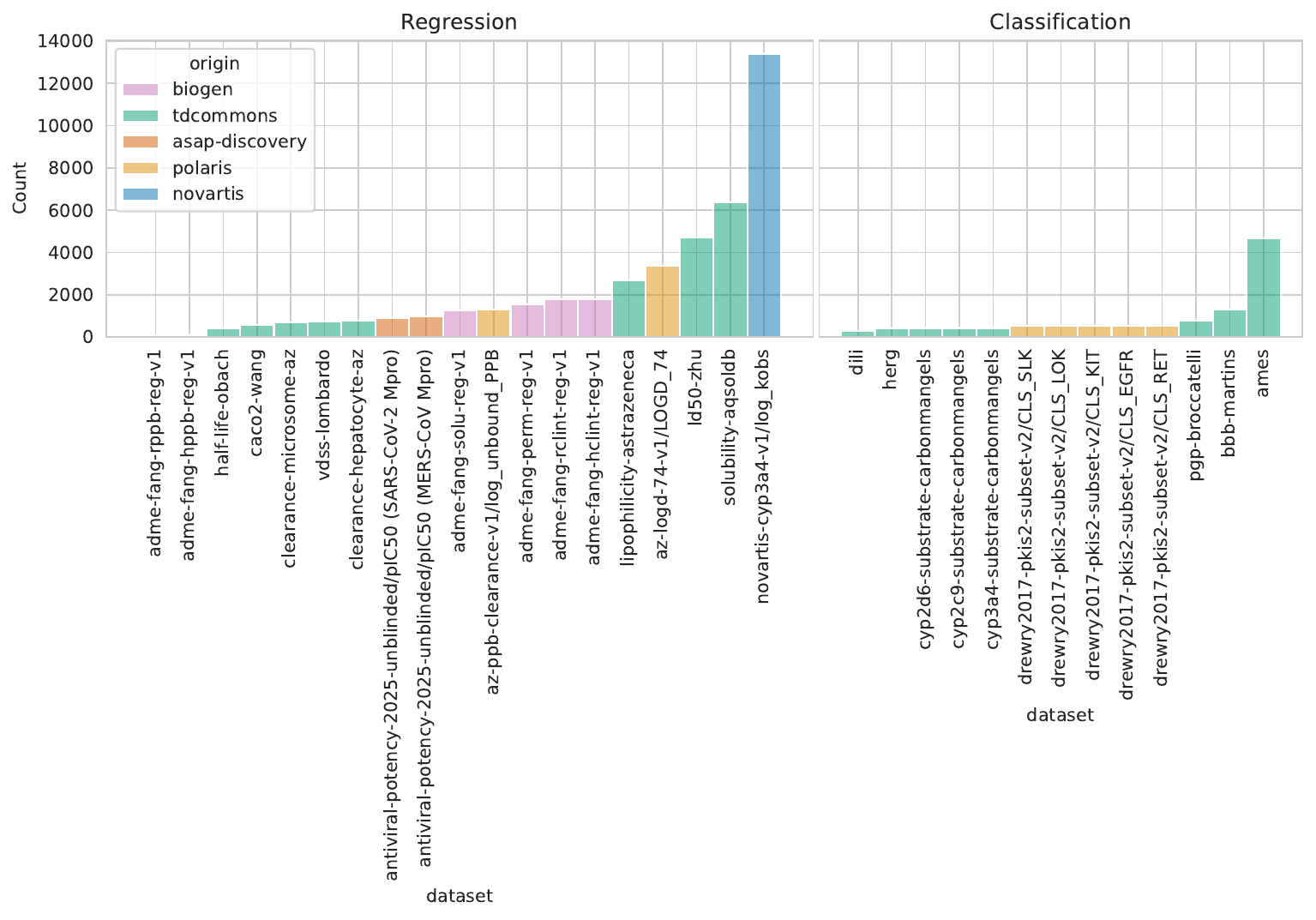}
    \caption{
    \textbf{Task sizes in the molecular property prediction objectives.}
    The vast majority of tasks consist of regression tasks, and the largest benchmark used is the TDC benchmark.
    }
    \label{fig:mol_prop_task_size}
\end{figure}

We collected property prediction tasks from the Polaris platform, a centralized hub for molecular property prediction benchmarks.
The dataset comprises 27 distinct benchmarks and datasets from academic and industrial sources, covering a wide range of ADMET (Absorption, Distribution, Metabolism, Excretion, and Toxicity) properties and biological activities.
As shown in~\autoref{fig:mol_prop_task_size}, the dataset is heavily skewed towards regression tasks, with 79\% of prompts being regression and 21\% being classification tasks.

\paragraph{Data Sources}
The property prediction benchmarks were extracted from the following sources:

\begin{itemize}
    \item \textbf{ASAP-Discovery}: Antiviral potency prediction against SARS-CoV-2 and MERS-CoV main proteases (pIC50 values).
    \item \textbf{Biogen ADME Suite}: Solubility, plasma protein binding (rat and human), MDR1 MDCK efflux ratio, and liver microsomal stability (rat and human).
    \item \textbf{Novartis Datasets}: CYP enzyme inactivation kinetics.
    \item \textbf{Therapeutic Data Commons (TDC)}: A comprehensive collection including P-glycoprotein inhibition, volume of distribution, blood-brain barrier penetration, Caco-2 cell permeability, drug-induced liver injury (DILI), hERG blocking, mutagenicity (AMES), clearance (hepatocyte and microsome), acute toxicity (LD50), CYP substrate predictions, and aqueous solubility.
    \item \textbf{Drewry Kinase Inhibitors (PKIS2)} (Polaris): Binary classification tasks for inhibition of EGFR, KIT, RET, LOK, and SLK kinases.
    \item \textbf{AstraZeneca (AZ) Datasets} (Polaris): LogD distribution coefficient, plasma protein binding clearance.
\end{itemize}

\subsection{Task Diversity}

\begin{figure}
    \centering
    \includegraphics[width=1\linewidth]{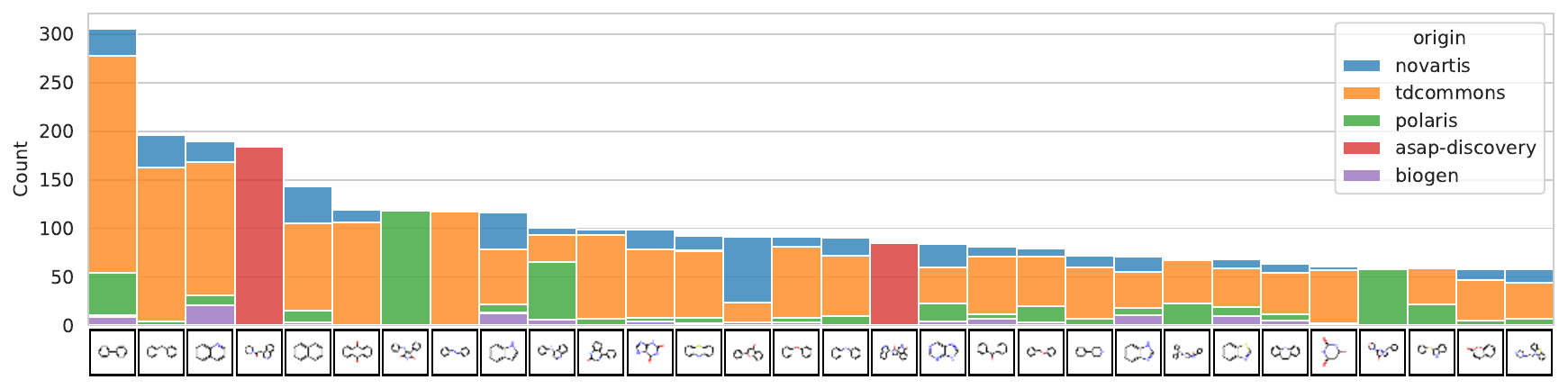}
    \caption{
    \textbf{Scaffold occurrence in the various benchmarks.}
    Occurrences of the most frequent Murcko scaffolds (of at least 6 atoms) in each benchmark, illustrating the chemical diversity across tasks.
    }
    \label{fig:mol_prop_scaff}
\end{figure}

To assess the diversity of molecules and tasks in our property prediction dataset, we analyzed Murcko scaffold distributions across benchmarks.
As shown in~\autoref{fig:mol_prop_scaff}, most benchmarks exhibit distinct scaffold patterns, indicating that the dataset covers chemically diverse molecular spaces rather than being biased towards a single scaffold class.
However, data extracted from the asa-discovery dataset are mainly centered around two main scaffolds present in no other benchmark.

\FloatBarrier

\section{Molecular Reactions Tasks}
\label{app:reaction_data}

Complementary to Molecular Generation and Molecular Property Prediction tasks, we designed tasks based on chemical reactions, focusing on the prediction of a retro-synthesis plan, i.e., a sequence of chemical reactions that can be used to synthesize a target compound from simpler building blocks.
We have not yet evaluated models on this task, and preliminary results suggest that solely relying on reinforcement learning to train LLMs for this task is complicated, notably due to the fact that the model has to learn the building blocks it can use, all of which might not fit in the context window of the LLM.
We believe that this task could benefit from the use of tool calls and information retrieval, and we leave the design of such a training pipeline and the evaluation of LLMs on this task to future work.
We therefore provide a description of our work in this appendix section, but leave the full analysis of the performance of LLMs on this task to future work.

\paragraph{Evaluation Metrics.}
Following~\citep{lee2025rethinkingmoleculesynthesizabilitychainofreaction, gao2024generativeartificialintelligencenavigating}, we evaluate on real-world synthesis prediction rather than synthetic data.
We focus on predicting synthesis routes for 1k molecules from the ChemBL dataset and 1k molecules from the Enamine dataset, with no baseline synthesis route provided.
For each molecule, we either directly prompt the model to predict the synthesis or provide the top-20 most structurally similar building blocks to assist the prediction.
The performance of the model is evaluated based on the success rate of synthesizing the target molecule using the predicted synthesis routes and available building blocks, and on the Tanimoto similarity between the target molecule and the synthesized product if the synthesis fails.

\subsection{Data Generation Pipeline}

We follow the methodology described in~\citep{lee2025rethinkingmoleculesynthesizabilitychainofreaction, gao2024generativeartificialintelligencenavigating}, which employs building blocks from the Enamine catalog and 115 chemical reaction templates described in their SMARTS notation to generate multi-step reactions.
This method iteratively creates more complex molecules by applying reactions to previous products, enabling the construction of multi-step synthesis routes in a scalable manner.
However, this approach can generate molecules that are not drug-like, particularly those with excessive molecular weight, as the number of steps in the generated synthesis plan increases.

To address this limitation, we apply property-based filtering to guide product selection toward drug-like compounds.
Importantly, we relax these filters during the early steps of multi-step syntheses, allowing non-drug-like intermediates to be generated before enforcing drug-likeness on the final product (see details below).

\subsubsection{Multi-Step Synthesis Generation (Stack-Based Approach)}

\begin{figure}
    \centering
    \includegraphics[width=1\linewidth]{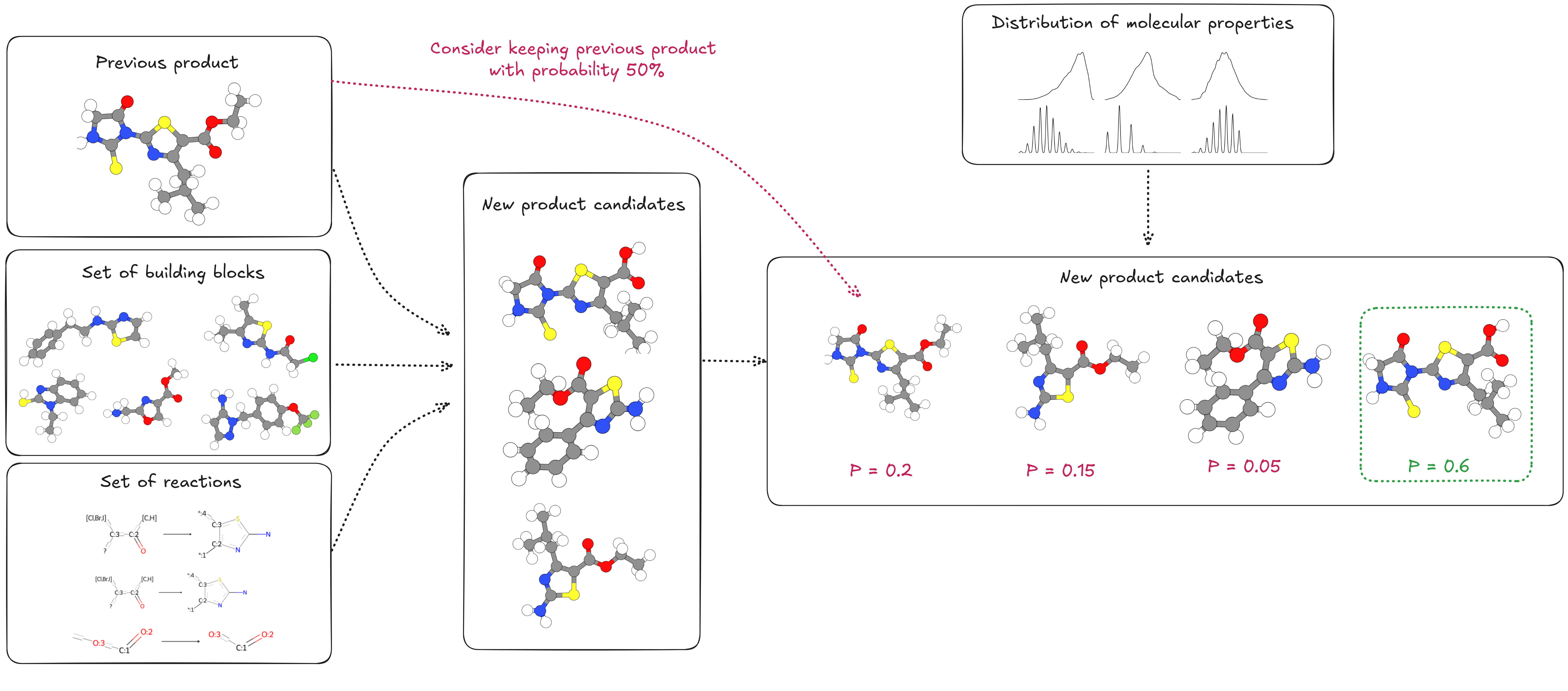}
    \caption{
    \textbf{Overview of the molecular reaction dataset generation pipeline.} Iterative stochastic process of synthesis generation: initialization with seed reactions, relaxed filtering for early steps, property filtering for later steps, probabilistic product selection, and chain extension up to 5 reaction steps.
    }
    \label{fig:MOLREAC}
\end{figure}

We generate synthetic pathways through an iterative stochastic process as follows:
\begin{enumerate}
    \item \textbf{Initialization}: Select a random seed reaction and identify available reactants via the compatibility matrix.

    \item \textbf{Relaxed Filtering for Early Steps}: For multi-step syntheses (i.e., when the total number of steps $n_{\text{steps}} > 1$), we randomly sample a number of initial steps $n_{\text{nf}} \sim \mathcal{U}\{0, \lfloor (n_{\text{steps}}+1)/2 \rfloor\}$ allowed to produce molecules with abnormal properties for drug-like compounds, and products are selected by randomly selecting one allowed reaction given the previous product, and applying this reaction to randomly selected allowed building blocks.

    \item \textbf{Probabilistic Product Selection}: After the no-filter steps (i.e., for steps $i > n_{\text{nf}}$), property-based filtering is re-enabled. For each valid product, we compute a probability score based on a target distribution over molecular properties (QED, molecular weight, TPSA, H-bond donors/acceptors, rotatable bonds, aromatic rings). Products are selected proportionally to these scores.
\end{enumerate}

In the end, with up to 5 reaction steps, we iteratively:
\begin{itemize}
    \item Select a new reaction compatible with the last product
    \item Identify available reactant partners via the matrix
    \item Apply the reaction, with or without property-based filtering, depending on the current step
    \item Add the product to the synthesis chain
\end{itemize}
The synthesis continues until the maximum number of steps is reached or no valid reactions can be applied.
Regardless of the intermediate relaxation, the \emph{final product} must pass all physicochemical property filters (see~\autoref{tab:property_filters}) and satisfy the maximum atom count constraint.
Pathways whose final product does not meet these criteria are discarded.

This process is illustrated in~\autoref{fig:MOLREAC}.

\subsubsection{Molecular Property Filtering and Modeling}

During the filtered steps of the synthesis (i.e., after the initial no-filter steps) and for the final product validation, molecules must satisfy strict physicochemical constraints to remain in the dataset, displayed in~\autoref{tab:property_filters}.

\begin{table}[h]
    \centering
    \begin{tabular}{lcc}
    \toprule
    \textbf{Property} & \textbf{Min} & \textbf{Max} \\
    \midrule
    QED (Drug-likeness) & 0.30 & 1.00 \\
    Molecular Weight (Da) & 0 & 600 \\
    TPSA (\AA$^2$) & 0 & 160 \\
    H-Bond Acceptors & 0 & 10 \\
    H-Bond Donors & 0 & 10 \\
    Rotatable Bonds & 1 & 10 \\
    Aromatic Rings & 0 & 6 \\
    \bottomrule
    \end{tabular}
    \caption{Molecular property constraints applied during synthesis generation.}
    \label{tab:property_filters}
\end{table}

On top of hard constraints, we compute log-probabilities for products via Beta distributions (parameterized by shape parameters $\alpha, \beta$) over the normalized property ranges:

$$\log p(x) = (\alpha - 1) \log(x_{\text{norm}}) + (\beta - 1) \log(1 - x_{\text{norm}})$$

This biases the stochastic selection toward drug-like molecules without rejecting valid synthetic products.
The parameters $\alpha$ and $\beta$ are tuned on the ZINC-250k dataset.

\subsection{Prompt Template Design}

We created eleven distinct objective templates to train models on complementary synthesis reasoning tasks:

\begin{enumerate}
    \item \textbf{Final Product}: Predict the final product of a multi-step synthesis given all reaction SMARTS
    \item \textbf{Reactant Prediction}: Identify a missing reactant for a single synthesis step
    \item \textbf{All Reactants}: Given a reaction SMARTS and target product, predict all required reactants
    \item \textbf{Building Block Constrained}: All reactants task with molecules restricted to a provided set
    \item \textbf{SMARTS Identification}: Predict the SMARTS representation for a reaction step
    \item \textbf{Full Synthesis Path}: Generate a multi-step synthesis pathway to a target molecule
    \item \textbf{Path with Building Block Reference}: Synthesis design constrained to a provided set of building blocks
    \item \textbf{Path with SMARTS Reference}: Synthesis design using only reactions from a curated set
    \item \textbf{Path with Both References}: Full pathway design under both building block and reaction constraints
    \item \textbf{Path with Intermediate Products}: Given a target molecule and a \emph{shuffled} list of intermediate products (i.e., all products of the synthesis route except the final one), determine the correct ordering of intermediates and provide the full synthesis route, including the reactants for each step. No building blocks or reaction templates are provided, requiring the model to identify appropriate reactants autonomously.
    \item \textbf{Path with Intermediate Products and Building Blocks}: Same as the previous task, but the model is additionally provided with a set of commercially available building blocks (containing the ground-truth reactants mixed with random distractors) to select from when constructing the synthesis route.
\end{enumerate}

Each prompt is formatted with a system message establishing chain-of-thought reasoning, followed by a user query. All numerical values (molecular weights, counts, etc.) are replaced with placeholders in the reaction string to facilitate generalization.
To prevent trivial memorization, approximately 10\% of prompts are converted to negative samples by:
\begin{itemize}
    \item Swapping the answer(s) with incorrect alternatives from the full dataset
    \item For multi-step paths, replacing the reaction SMARTS or building blocks with incompatible alternatives
    \item Maintaining the same prompt format to create balanced classification challenges
\end{itemize}

Samples marked as ``impossible'' during training allow models to learn when synthesis is infeasible.

This results in a dataset of 50k chemical reactions, including 12k single-step reactions, 17k two-step reactions, and 21k multi-step reactions (3 to 5 steps).
From this dataset, we define four main training tasks: predicting the product of multi-step reactions (12\%), identifying missing reactants (20\%), predicting the SMARTS representation of reactions (9\%), and predicting complete retro-synthesis plans (59\%), mainly focusing on our main objective task: multi-step synthesis.

\subsection{Dataset Description}

\subsubsection{Synthesis Complexity Analysis}

The complexity of synthetic pathways is characterized by two main metrics: the number of reaction steps and the structural similarity between consecutive intermediates.

\begin{figure}
    \centering
    \begin{subfigure}{0.45\linewidth}
        \includegraphics[width=1\linewidth]{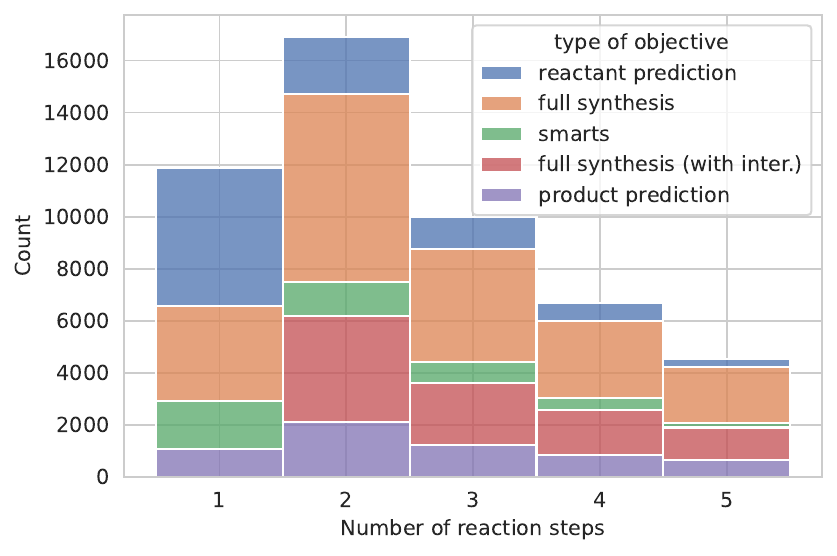}
        \caption{
        \textbf{Distribution of reaction steps per synthesis pathway.}
        }
        \label{fig:reaction_steps_histogram}
    \end{subfigure}
    \hspace{0.03\linewidth}
    \begin{subfigure}{0.45\linewidth}
        \includegraphics[width=1\linewidth]{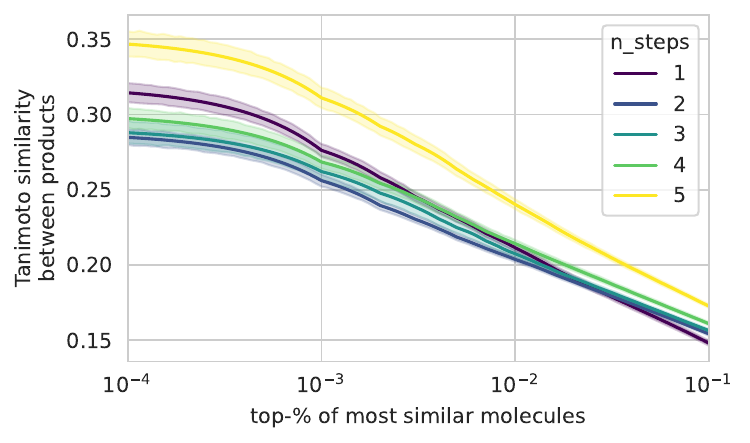}
        \caption{
        \textbf{Structural similarity between products generated with the same number of steps.}
        }
        \label{fig:tanimoto_similarity_per_reaction_step}
    \end{subfigure}
    \end{figure}

The distribution of reaction steps shows that most pathways are short: 1 to 3 steps (\autoref{fig:reaction_steps_histogram}), as increasing the number of steps tends to produce less realistic molecules.
\autoref{fig:tanimoto_similarity_per_reaction_step} illustrates how, when performing multiple reaction steps, the molecules generated are more diverse.

\subsubsection{Reaction Template Analysis}

\begin{figure}
    \centering
    \includegraphics[width=1\linewidth]{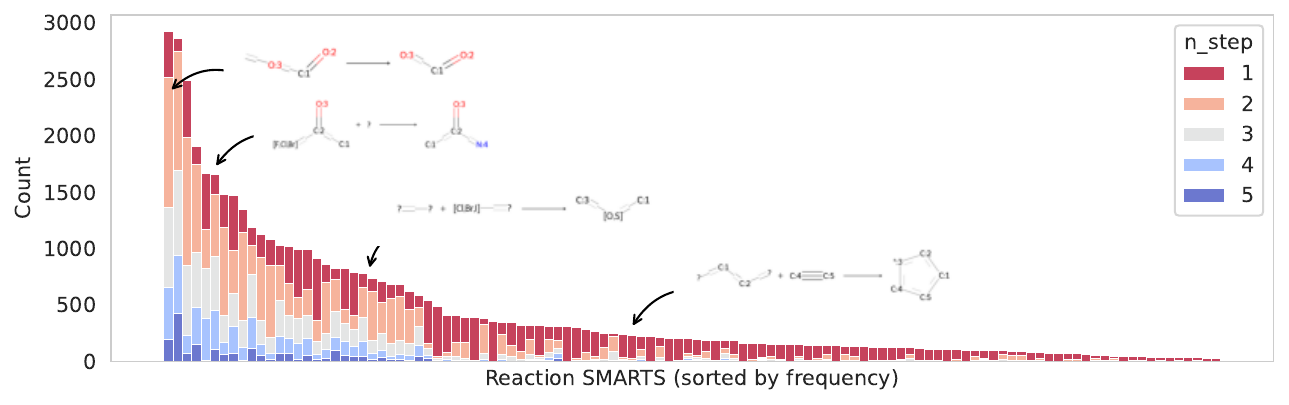}
    \caption{
    \textbf{Frequency and chemical diversity of reaction templates.}
    }
    \label{fig:smarts_frequency}
\end{figure}

The reaction templates form the core vocabulary of the synthesis dataset.
We examine both the frequency distribution and chemical diversity of the SMARTS patterns used during generation in \autoref{fig:smarts_frequency}.
The reaction template distribution follows a power-law pattern, with a small number of highly frequent transformations and a long tail of "specialized" reactions, especially when the number of steps increases, where some reactions are clearly more frequent than others.

\paragraph{Reward Design.}
For the chemical reaction tasks, we compute rewards differently depending on the task type.
\autoref{tab:all_rewards} describes how $\prompt$ is defined for each task type, and how the reward is computed.
For reactant and product prediction, the reward is $1$ if the extracted answer matches the expected label. We also provide a non-zero reward if the model generates partially correct answers (with the introduction of the intersection over union term).
Similarly, for  SMARTS prediction, the model gets a small reward if applying the predicted SMARTS to the reactant produces the right product (which could be due to producing a SMARTS too specific for the reactants/products), and $1$ if both SMARTS are equal.
Finally, for the prediction of a synthesis plan, we go through the proposed $n$-step synthesis, and store $n_{\text{valid}}$, the number of valid steps, i.e, correct chemical reactions only using Enamine building blocks or previous products of valid steps.
We then store the product of the last valid step, and provide to the model a reward increasing with the Tanimoto similarity between this product and the target product, and with the proportion of valid steps.
The exponents used in the formula are used to ensure the rewards given to incorrect synthesis routes remain low.

\section{Creating a Reasoning Molecular Modeling Agent}
\label{appendix:training_details}

\begin{figure}
    \centering
    \includegraphics[width=1\linewidth]{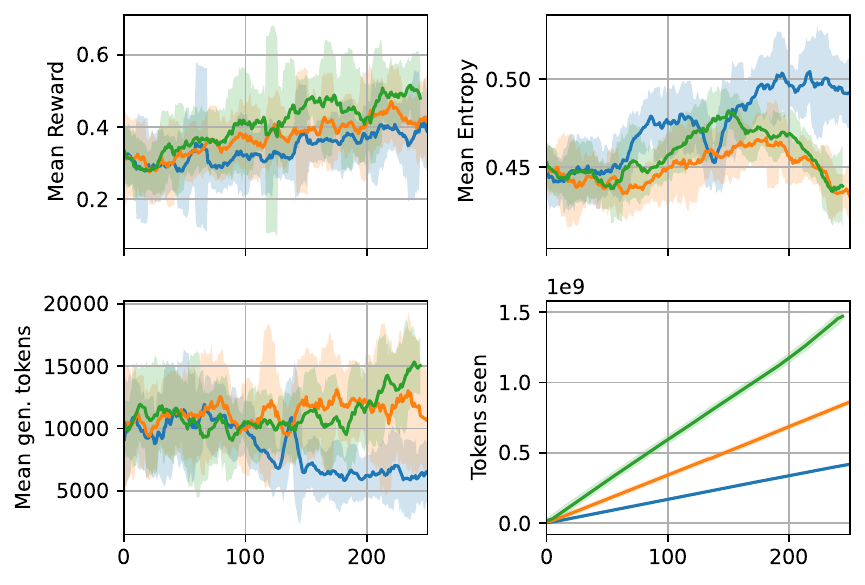}
    \caption{
        \textbf{Training curves of RL-Molstral.}
        Evolution of the average reward (left) and average completion length (right) during training.
    }
    \label{fig:training_curves}
\end{figure}

We fine-tune Mistral-Small 4, a 128B reasoning model with a mixture of experts architecture, resulting in 6B parameters being active.

We trained the model for ~250 steps, with a group size of 4,8 and 16 and a number of prompts per batch of 32.
To ensure the model primarily focuses on the molecular generation tasks, we sampled prompts from each task with a ratio of: 2 for de novo molecular generation, 1 for molecular property prediction.
Note that all three runs are not iso-compute: the run with group size 16 is more computationally intensive than the run with group size 4, as it requires more rollouts per prompt to compute the reward and update the model, which can be seen with the amount of tokens seen by each models in~\autoref{fig:training_curves}.

We show in~\autoref{fig:training_curves} the evolution of the average reward, entropy, tokens generated and the number of tokens seen by each models at each step.
Overall it can be seen that for all three runs, the average reward increases during training, with a more significant increase for the run with a group size of 16, which is expected as it relies on more rollouts to compute the reward and update the model.
Interestingly, a low groupsize seems to diminish the model's ability to generate long chains of thought, as the average completion length is around two times higher for the run with group size 16 compared to the run with group size 4, which decrease compared to the base model.

Finally, we note that the model with the largest entropy at the end of training is the model with group size of 4, which could align with our observations in~\autoref{sec:results} that larger group size tend to decrease the diversity of the generated molecules.
However, its entropy is larger than the base model while being less diverse, and the entropy of the two other models (groupsize 8 and 16) are very similar at the end of training, while we did notice a difference in the diversity of the generated molecules.
Overall this suggests that the entropy of the model is not a good indicator of whether the model can generate diverse compounds, which can be explained by the fact that the model can generate diverse chain of thoughts that might all lead to similar molecules, highlighting the need to assess these two metrics separately.

\subsection{Compute Resources}
\label{app:compute_resources}

All experiments in this work were conducted on NVIDIA H100 GPUs.
For the model evaluation experiments, compute requirements vary depending on model size and inference speed, but we recorded an average of 500 GPU-hours per model for comprehensive evaluation across all tasks and rollouts reported in the paper.
For the RL training of RL-Mistral, we utilized 72 H100 GPUs over a training period of ~1 days per model.

\FloatBarrier

\FloatBarrier
\section{De Novo Generation results details}
\label{appendix:molgen_results_details}

\subsection{Completion validity}
\begin{figure}
    \centering
    \includegraphics[width=0.5\linewidth]{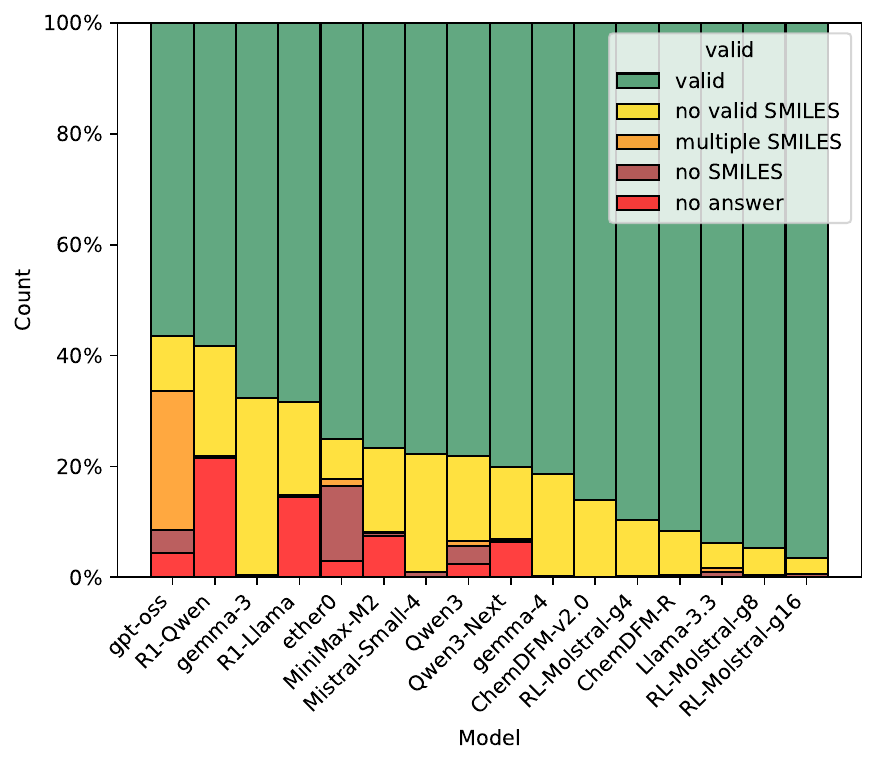}
    \caption{
        \textbf{Validity of the generated completions.}
        Description of the validity of the generated completions.
        Generations can be invalid due to no answer being generated in the expected format, no SMILES being parsed in the answer, or no valid SMILES or multiple SMILES being proposed.
    }
    \label{fig:validity}
\end{figure}

When generating molecular completions, we expect the model to generate answers in a specific format, i.e., \texttt{<answer>}SMILES\texttt{</answer>}, with a single valid SMILES being generated.
However, some models may generate invalid completions that do not conform to this format.
ChemDFM-v2.0, for instance, does not generate the \texttt{<answer>} tokens if prompted for, although this model does not generate any additional text than a molecular SMILES.
We then relaxed the expected generation template specifically for this model.

\autoref{fig:validity} presents the validity of the generated completions for the different models evaluated, with the reasons for invalidity.
This figure provides multiple insights:
\begin{itemize}
    \item Most models struggle generating valid completions, and only a few models manage to generate more than 80\% valid completions.
    \item Llama-3.3 is among the best models when it comes to generating valid completions. Still, as we will see in the next section, these molecules are most often duplicates.
    \item Gemma-3 shows a unique behavior, where almost all completions are correctly parsed, but more than 30\% of them are invalid SMILES.
    \item While ether0 was tuned on chemical data, it generates a relatively high amount of completions with no SMILES in the answer or no answer at all, a behavior further explored in~\autoref{appendix:ether0_analysis}.
\end{itemize}

\subsection{Diversity and Uniqueness of the generated molecules}
\label{app:diversity_uniqueness}

\begin{figure}
    \centering
    \includegraphics[width=1\linewidth]{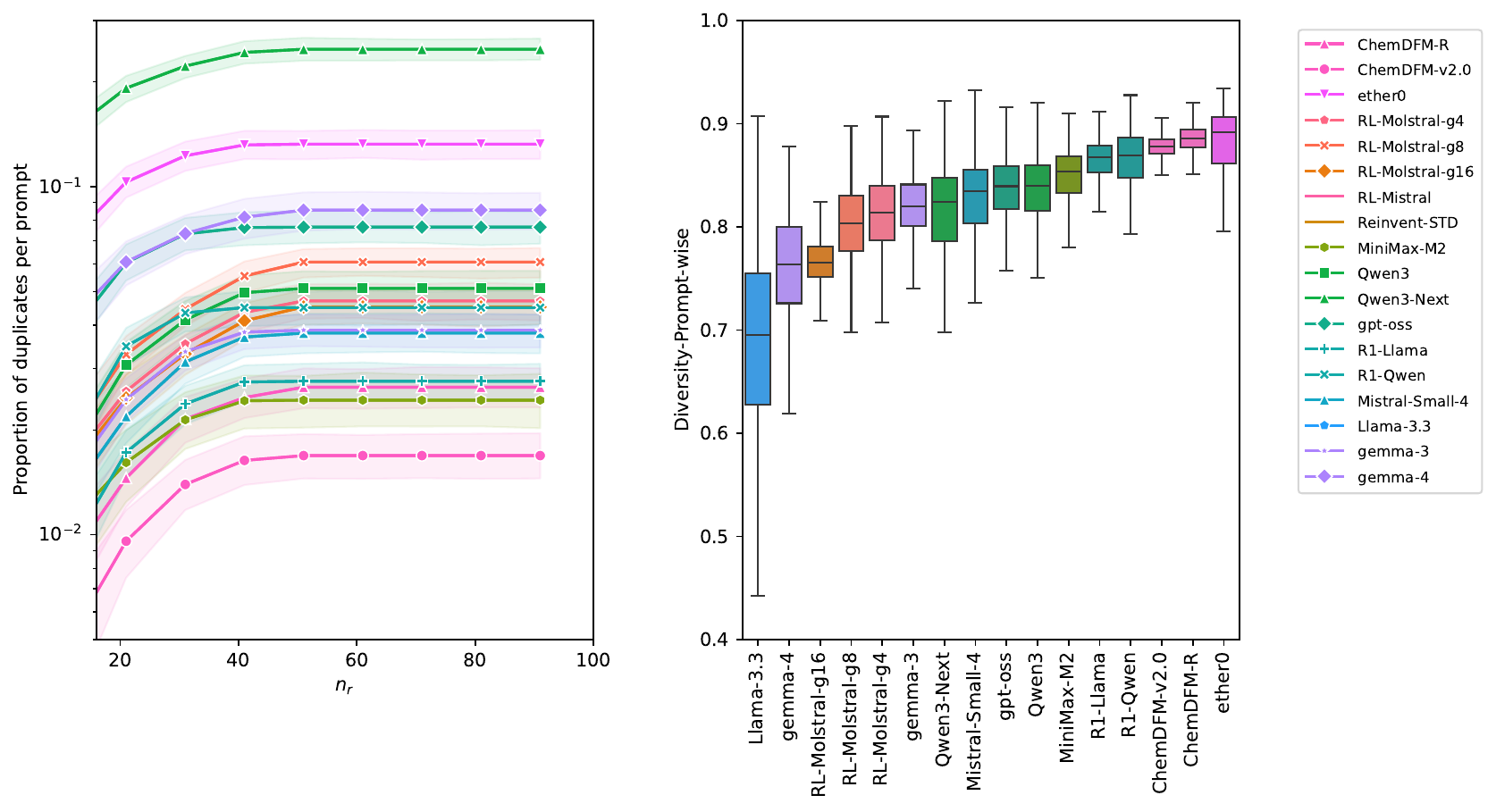}
    \caption{
        \textbf{Uniqueness and diversity evolution with the number of rollouts.}
        We display the uniqueness (left) and diversity (right) of the generated molecules with respect to the number of rollouts.
    }
    \label{fig:uniqueness_diversity}
\end{figure}

\autoref{fig:uniqueness_diversity} presents the evolution of the uniqueness and diversity of the generated molecules.

We observe that most models achieve generating unique molecules for a given prompt, with the exception of Llama-3.3.

\begin{figure}
    \vspace{-1cm}
    \centering
    \begin{subfigure}{0.48\linewidth}
        \includegraphics[width=\linewidth]{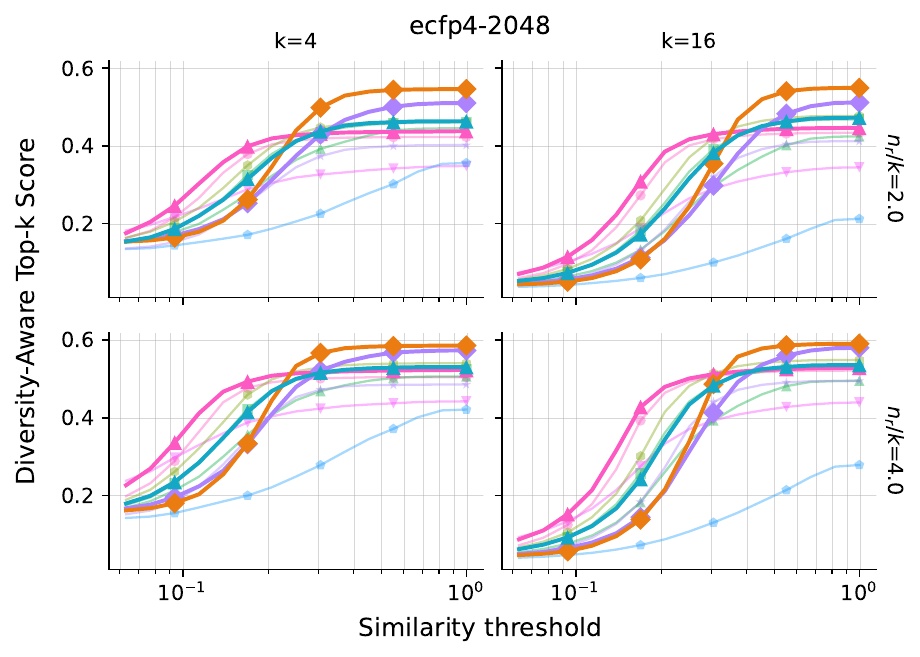}
    \end{subfigure}
    \begin{subfigure}{0.48\linewidth}
        \includegraphics[width=\linewidth]{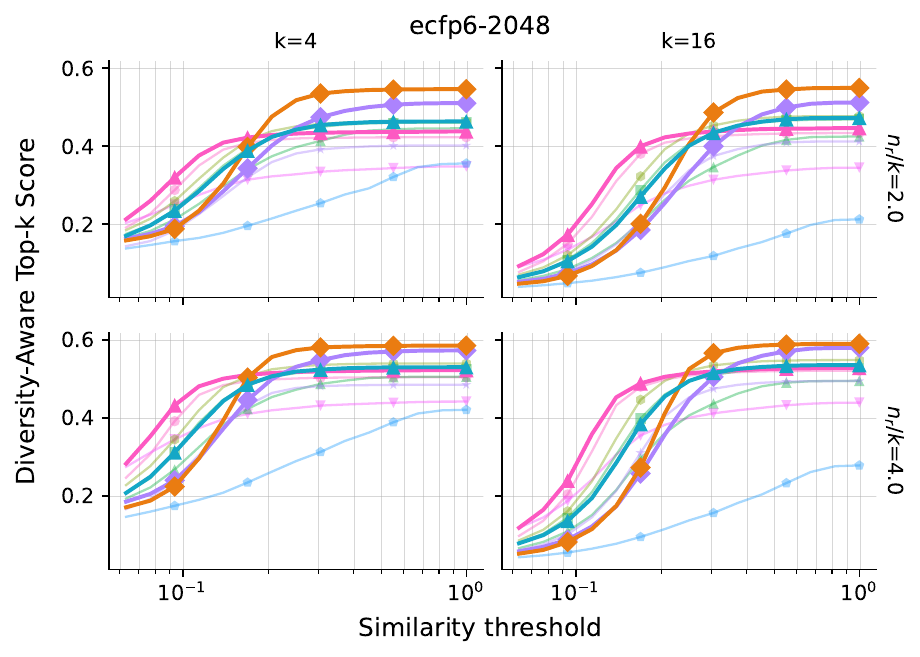}
    \end{subfigure}
    \begin{subfigure}{0.48\linewidth}
        \includegraphics[width=\linewidth]{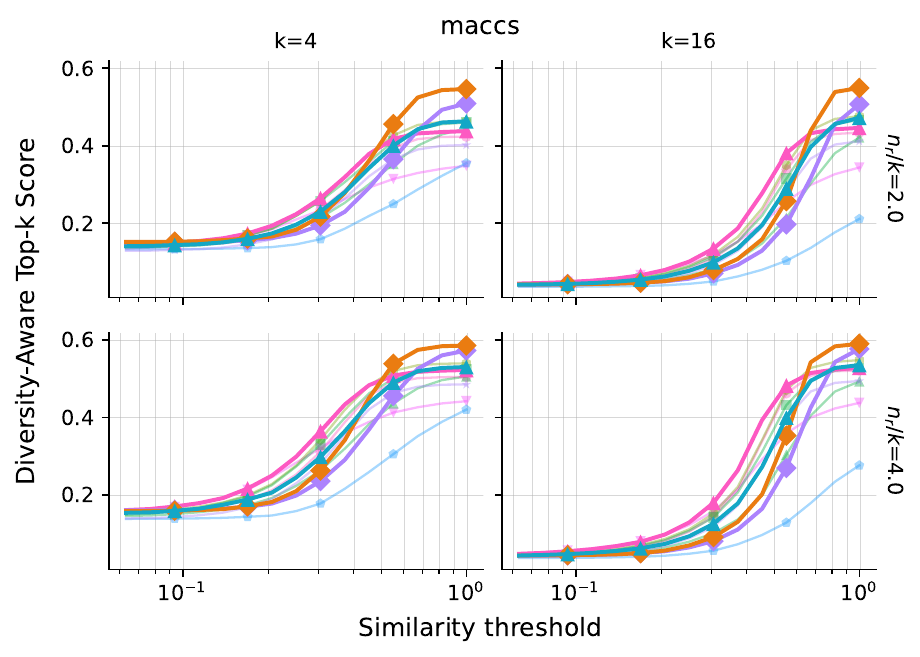}
    \end{subfigure}
    \begin{subfigure}{0.48\linewidth}
        \includegraphics[width=\linewidth]{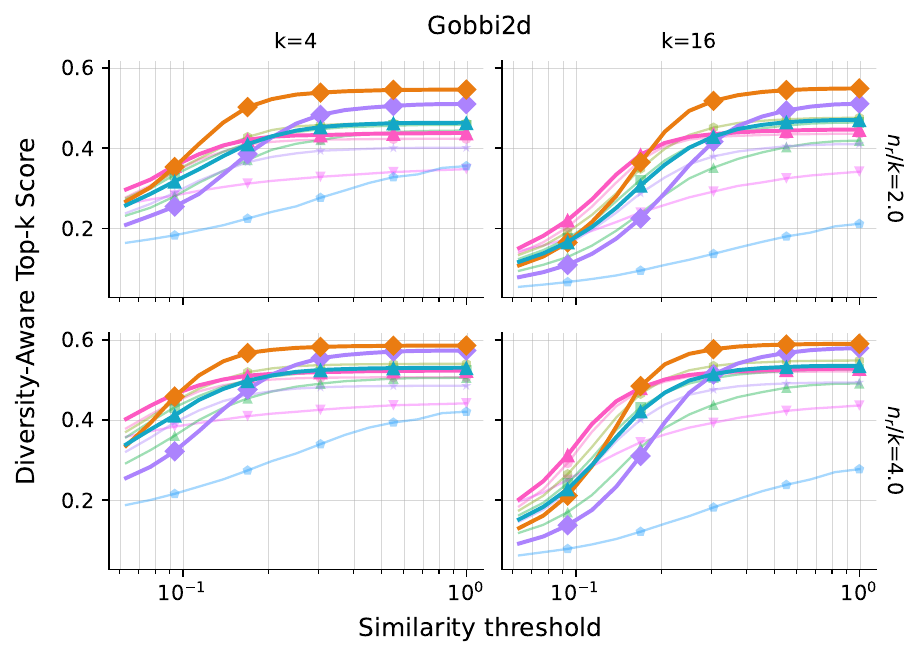}
    \end{subfigure}
    \begin{subfigure}{0.65\linewidth}
        \includegraphics[width=\linewidth]{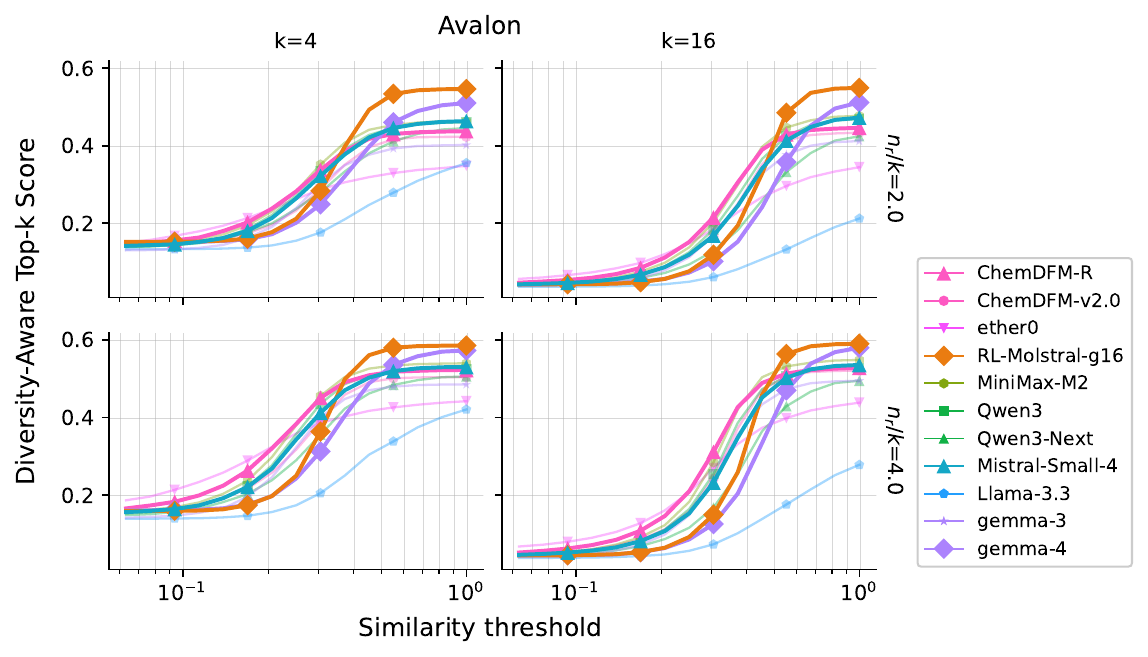}
    \end{subfigure}
        \caption{\textbf{Diversity-aware top-k score for different fingerprints.}
        We display the diversity-aware metric when the similarity between molecules is based on various families of fingerprints.
    }
    \label{fig:diversity_fps}
\end{figure}

Finally, we proposed a diversity-aware top-k score in the main text to account for the chemical diversity of the generated molecules.
We provide in~\autoref{fig:diversity_fps} the evolution of the diversity-aware top-k score for various fingerprinting methods used to compute the similarity between molecules.
Depending on the choice of the fingerprint, the diversity-aware top-k score can vary, although the overall ranking of the models remains similar.

\subsection{Comparison to Traditional Generation Algorithm}

Compared to traditional generation algorithms, LLMs offer one main potential benefit: by conditioning the generation process on a prompt and training in such a setup, LLMs can generate molecules in a zero-shot manner, i.e., they do not need specific training for each task, unlike most traditional generation algorithms.
However, this likely comes at the cost of a lower performance compared to these traditional algorithms, especially as most pretrained models have not been specifically specialized for molecular generation.
Additionally, properly comparing the quality of the generations under the same number of rollouts is complicated. When performing a task-specific training, each generated batch we compute the loss upon could be counted as a rollout.
In this scenario, these task-specific trainings would hardly reach more than 4 steps, hurting the performance of these baselines.
Conversely, in the rest of the section, the baselines we trained generated ~100k molecules, from which they got feedback to update their policy, which is hardly comparable with the setting used for the LLMs.

Despite these discrepancies, making a head-to-head comparison between LLMs and traditional generation algorithms difficult, we believe comparing the performances of LLMs to a traditional generation algorithm baseline can be insightful to understand the current gap between these two approaches, and have an idea of what performance can be expected from LLMs if they managed to reach the performance of traditional generation algorithms.
To compare the models, we assumed the molecules generated at training time do not count as rollouts, and once the training is finished, we generate $n_r$ molecules with the best checkpoint obtained.

We implemented a variation of Reinvent~\citep{reinvent4}, using a transformer-based policy network instead of a recurrent one~\citep{reinventtransf}.
We also found that instead of using the original Reinvent loss: $L(\prompt, \compl) = \pp{NLL_\text{ref}(\compl | \prompt) + r(\compl, \prompt) - NLL_\text{actor}(\compl | \prompt)}^2$, where $NLL_\text{ref}$ is the negative log-likelihood of the reference model, $NLL_\text{actor}$ is the negative log-likelihood of the policy network, and $r(\compl, \prompt)$ is the reward of the generated molecule, using the GRPO loss described in~\autoref{sec:training} led to more stable performance, and we used it for the comparison.
We ran a hyperparameter grid over the learning rate (1e-5, 5e-5, 1e-4), batch size (16, 32, 64, 128), and KL coefficient (1, 0.1, 0.01), and selected the best-performing set of hyperparameters across all tasks.

Each run was performed on a single H100 GPU for 1 hour, which is three times longer than obtaining the same number of rollouts with RL-Molstral-g16, without accounting for hyperparameter tuning.
For this reason, we performed the comparison only on the first 100 prompts of the test set.

\begin{table}
    \caption{
        \textbf{\textit{De novo} generation results.}
        Evaluation of various LLMs on the molecular generation task.
        We report the average top-k scores among $n_r$ generations per prompt and standard deviations, averaged over the unique prompts.
        Models are ordered by their release date.
    }
    \resizebox{\textwidth}{!}{
        \centering
        \begin{tabular}{c|c||p{1.3cm}p{1.3cm}||p{1.3cm}p{1.3cm}||p{1.3cm}p{1.3cm}||p{1.3cm}p{1.3cm}||}
\toprule
 &  & \multicolumn{2}{c||}{$n_r=$16} & \multicolumn{2}{c||}{$n_r=$24} & \multicolumn{2}{c||}{$n_r=$32} & \multicolumn{2}{c||}{$n_r=$48} \\
Model & Size & $\text{top}_{1}$ & $\text{top}_{16}$ & $\text{top}_{1}$ & $\text{top}_{16}$ & $\text{top}_{1}$ & $\text{top}_{16}$ & $\text{top}_{1}$ & $\text{top}_{16}$ \\
\midrule
reinvent & 0.2B & \textbf{\underline{{\cellcolor[rgb]{0.906, 0.784, 0.776}} 0.78 \tiny $\pm$0.03}} & \textbf{\underline{{\cellcolor[rgb]{0.929, 0.925, 0.937}} 0.53 \tiny $\pm$0.04}} & \textbf{\underline{{\cellcolor[rgb]{0.902, 0.776, 0.769}} 0.78 \tiny $\pm$0.01}} & \textbf{\underline{{\cellcolor[rgb]{0.980, 0.945, 0.941}} 0.63 \tiny $\pm$0.03}} & \textbf{\underline{{\cellcolor[rgb]{0.898, 0.765, 0.757}} 0.79 \tiny $\pm$0.02}} & \textbf{\underline{{\cellcolor[rgb]{0.969, 0.922, 0.914}} 0.66 \tiny $\pm$0.02}} & \textbf{\underline{{\cellcolor[rgb]{0.894, 0.757, 0.749}} 0.8 \tiny $\pm$0.01}} & \textbf{\underline{{\cellcolor[rgb]{0.941, 0.863, 0.859}} 0.7 \tiny $\pm$0.01}} \\
gemma-4 & 31B & {\cellcolor[rgb]{0.976, 0.941, 0.937}} 0.64 \tiny $\pm$0.08 & {\cellcolor[rgb]{0.702, 0.749, 0.827}} 0.33 \tiny $\pm$0.05 & \underline{{\cellcolor[rgb]{0.957, 0.894, 0.890}} 0.68 \tiny $\pm$0.05} & {\cellcolor[rgb]{0.835, 0.851, 0.886}} 0.45 \tiny $\pm$0.03 & {\cellcolor[rgb]{0.953, 0.890, 0.882}} 0.68 \tiny $\pm$0.05 & {\cellcolor[rgb]{0.886, 0.890, 0.914}} 0.5 \tiny $\pm$0.02 & \underline{{\cellcolor[rgb]{0.941, 0.859, 0.851}} 0.71 \tiny $\pm$0.02} & {\cellcolor[rgb]{0.929, 0.925, 0.937}} 0.54 \tiny $\pm$0.01 \\
Mistral-Small-4 & 128B & {\cellcolor[rgb]{0.965, 0.953, 0.957}} 0.57 \tiny $\pm$0.06 & {\cellcolor[rgb]{0.659, 0.718, 0.812}} 0.29 \tiny $\pm$0.04 & {\cellcolor[rgb]{0.976, 0.961, 0.957}} 0.59 \tiny $\pm$0.05 & {\cellcolor[rgb]{0.773, 0.800, 0.859}} 0.4 \tiny $\pm$0.03 & {\cellcolor[rgb]{0.980, 0.953, 0.949}} 0.62 \tiny $\pm$0.04 & {\cellcolor[rgb]{0.808, 0.827, 0.875}} 0.43 \tiny $\pm$0.02 & {\cellcolor[rgb]{0.980, 0.945, 0.941}} 0.63 \tiny $\pm$0.02 & {\cellcolor[rgb]{0.859, 0.871, 0.902}} 0.47 \tiny $\pm$0.01 \\
MiniMax-M2 & 229B & {\cellcolor[rgb]{0.976, 0.961, 0.957}} 0.59 \tiny $\pm$0.09 & {\cellcolor[rgb]{0.663, 0.722, 0.812}} 0.3 \tiny $\pm$0.05 & {\cellcolor[rgb]{0.980, 0.949, 0.945}} 0.62 \tiny $\pm$0.06 & {\cellcolor[rgb]{0.776, 0.804, 0.859}} 0.41 \tiny $\pm$0.04 & {\cellcolor[rgb]{0.976, 0.941, 0.937}} 0.63 \tiny $\pm$0.07 & {\cellcolor[rgb]{0.835, 0.851, 0.886}} 0.46 \tiny $\pm$0.02 & {\cellcolor[rgb]{0.957, 0.894, 0.890}} 0.68 \tiny $\pm$0.01 & {\cellcolor[rgb]{0.875, 0.878, 0.910}} 0.49 \tiny $\pm$0.01 \\
ChemDFM-R & 14B & {\cellcolor[rgb]{0.973, 0.957, 0.957}} 0.58 \tiny $\pm$0.08 & {\cellcolor[rgb]{0.671, 0.725, 0.816}} 0.3 \tiny $\pm$0.04 & {\cellcolor[rgb]{0.980, 0.953, 0.949}} 0.62 \tiny $\pm$0.05 & {\cellcolor[rgb]{0.765, 0.796, 0.855}} 0.39 \tiny $\pm$0.03 & {\cellcolor[rgb]{0.980, 0.945, 0.941}} 0.63 \tiny $\pm$0.05 & {\cellcolor[rgb]{0.808, 0.827, 0.875}} 0.43 \tiny $\pm$0.02 & {\cellcolor[rgb]{0.973, 0.925, 0.922}} 0.65 \tiny $\pm$0.02 & {\cellcolor[rgb]{0.855, 0.863, 0.898}} 0.47 \tiny $\pm$0.02 \\
RL-Molstral-g16 & 128B & \underline{{\cellcolor[rgb]{0.976, 0.937, 0.933}} 0.64 \tiny $\pm$0.08} & \underline{{\cellcolor[rgb]{0.796, 0.820, 0.871}} 0.42 \tiny $\pm$0.04} & {\cellcolor[rgb]{0.965, 0.914, 0.910}} 0.66 \tiny $\pm$0.08 & \underline{{\cellcolor[rgb]{0.875, 0.878, 0.910}} 0.49 \tiny $\pm$0.02} & \underline{{\cellcolor[rgb]{0.953, 0.882, 0.878}} 0.69 \tiny $\pm$0.07} & \underline{{\cellcolor[rgb]{0.906, 0.906, 0.925}} 0.51 \tiny $\pm$0.02} & {\cellcolor[rgb]{0.941, 0.859, 0.851}} 0.71 \tiny $\pm$0.04 & \underline{{\cellcolor[rgb]{0.937, 0.929, 0.941}} 0.54 \tiny $\pm$0.01} \\
\bottomrule
\end{tabular}

    }
    \label{tab:gen_table_reinvent}
\end{table}

\begin{figure}
    \centering
    \includegraphics[width=0.95\linewidth]{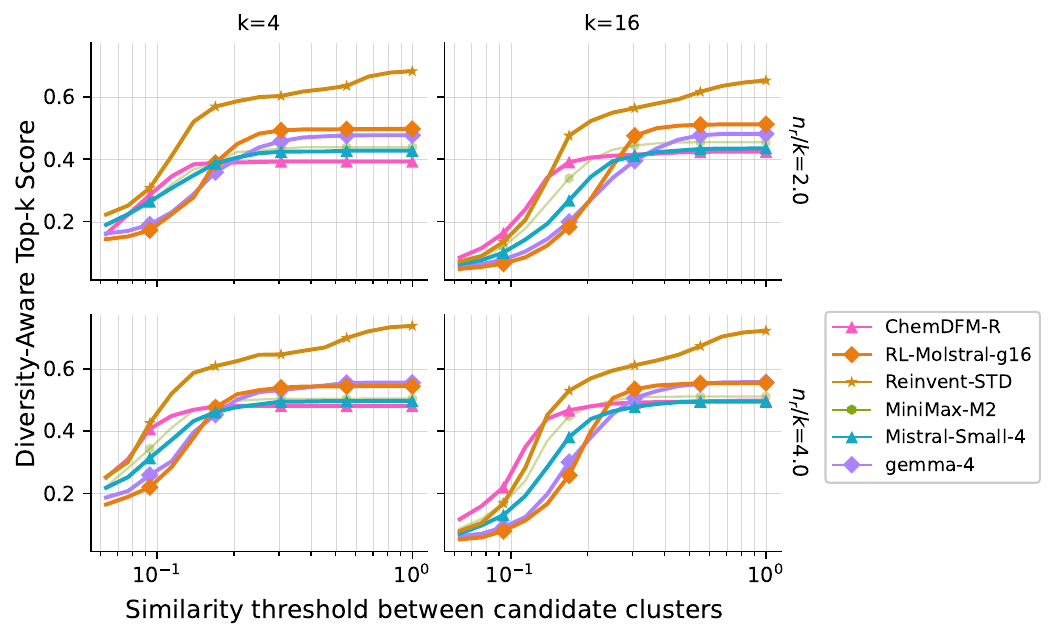}
    \caption{
        \textbf{Diversity-Aware scores with Reinvent.}
        We compare the performance of RL-Molstral-g16 to a traditional generation algorithm baseline, which is a genetic algorithm optimizing the same reward as the one used for the LLMs.
    }
    \label{fig:rl_molstral_comparison}
\end{figure}

As expected, the results presented in \autoref{tab:gen_table_reinvent} and \autoref{fig:rl_molstral_comparison} demonstrate that LLMs significantly underperform compared to the traditional Reinvent baseline.
However, if we account for the fact that generating compounds is three times faster with gemma-4 or RL-Molstral-g16, we can compare the results of Reinvent at 16 rollouts with the results of the LLMs at 48 rollouts.
In this setting, the performance gap is significantly reduced, especially when looking at the top-16 score.

\FloatBarrier
\section{Molecular Prediction results details}
\label{appendix:molprop_results_details}

\subsection{Global Additional Results}

\begin{figure}
    \centering
    \includegraphics[width=1\linewidth]{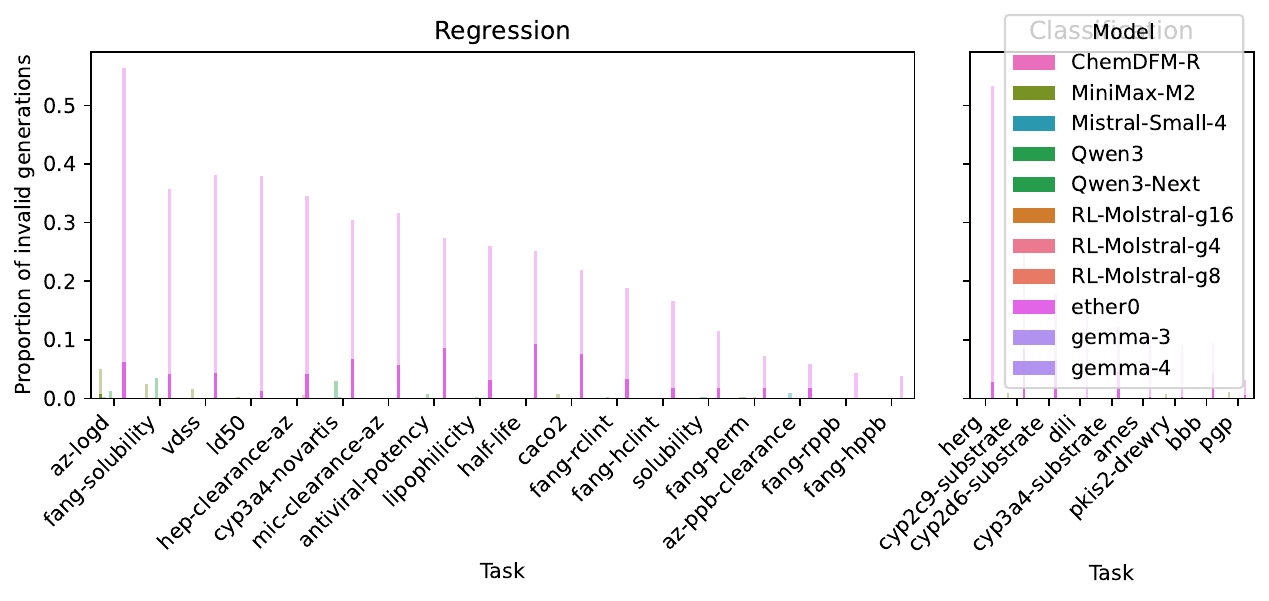}
    \caption{
        \textbf{Extraction results for the molecular property prediction.}
        Proportion of prompts for which none of the 5 generated answers could be processed into a numerical prediction.
        We highlight with the colors of the barplots the proportion of prompts where at least one of the completion had a number in the extracted answer, but was rejected (most often due to ambiguous answer).
    }
    \label{fig:valid_}
\end{figure}

To get the predictions of the model for the molecular property prediction tasks, we parse the generated text between the \texttt{<answer>} and \texttt{</answer>} tags.
The extracted text is then passed through a series of regex patterns to extract the predicted value.
This procedure has the advantage of being fast to compute, but can fail when the model generates long answers between the tags, which was for instance the case for MiniMax-M2, and we hence decided to not report its results on these tasks, as these might not reflect the model's performance.

\autoref{fig:valid_} displays the proportion of prompts for which none of the 5 generated answers could be processed into a numerical prediction.
The colors of the barplots shows the proportion of prompts where at least one of the completion had a number in the extracted answer (more saturated), but was rejected (most often due to ambiguous answer), or if no number was found in any of the completions (less saturated).
We observe first that all models succeed in generating parsable answer for classification tasks, but struggle more for regression tasks.
For instance, Qwen3-30B-A3B-Thinking only manages to provide predictions on the majority of the prompts of a task on 3 tasks out of 18.
However, we can observe that the vast majority of the rejections come from the fact that no number exists in any of the 5 completions of a prompt, which mostly occurs when the model  decides not to answer due to lack of knowledge on the topic.

Finally,~\autoref{tab:comprehensive_cls_res} and~\autoref{tab:comprehensive_reg_res} provide the comprehensive results of all models on all classification and regression tasks, respectively.

\begin{table}
    \caption{
        \textbf{Comprehensive Classification Results.}
    }
    \centering
    \begin{tabular}{c||ccccccccc}
\toprule
 & ames & bbb & cyp2c9 & cyp2d6 & cyp3a4 & dili & herg & pgp & pkis2-drewry \\
\midrule
\midrule
 & \multicolumn{9}{c}{ROC-AUC} \\
\midrule
\midrule
ChemDFM-R & 0.51 & 0.73 & 0.48 & 0.45 & 0.54 & 0.48 & 0.53 & 0.50 & 0.49 \\
MiniMax-M2 & 0.60 & 0.54 & 0.45 & 0.50 & 0.41 & 0.52 & 0.57 & 0.45 & 0.39 \\
Mistral-Small-4 & 0.70 & 0.73 & 0.55 & 0.64 & 0.54 & 0.56 & \underline{0.75} & 0.59 & 0.50 \\
Qwen3 & 0.71 & 0.62 & 0.60 & 0.65 & 0.56 & 0.68 & 0.74 & 0.66 & 0.55 \\
Qwen3-Next & \underline{0.74} & 0.73 & 0.64 & 0.65 & 0.56 & \textbf{\underline{0.79}} & 0.73 & \underline{0.72} & \underline{0.55} \\
RL-Molstral-g16 & 0.71 & \underline{0.75} & 0.54 & \underline{0.71} & 0.56 & 0.58 & 0.70 & 0.66 & \textbf{\underline{0.56}} \\
RL-Molstral-g4 & 0.71 & 0.74 & 0.49 & 0.66 & 0.53 & 0.60 & 0.69 & 0.54 & 0.50 \\
RL-Molstral-g8 & 0.73 & 0.73 & \underline{0.66} & 0.63 & \textbf{\underline{0.61}} & 0.64 & 0.74 & 0.63 & 0.49 \\
gemma-3 & 0.60 & 0.66 & 0.52 & 0.51 & 0.53 & 0.51 & 0.69 & 0.52 & 0.50 \\
gemma-4 & \textbf{\underline{0.79}} & \textbf{\underline{0.81}} & \textbf{\underline{0.72}} & \textbf{\underline{0.75}} & \underline{0.57} & \underline{0.74} & \textbf{\underline{0.87}} & \textbf{\underline{0.77}} & 0.53 \\
\midrule
\midrule
 & \multicolumn{9}{c}{Accuracy} \\
\midrule
\midrule
ChemDFM-R & 0.48 & 0.72 & 0.76 & 0.62 & 0.53 & 0.44 & 0.33 & 0.45 & \textbf{\underline{0.83}} \\
MiniMax-M2 & 0.61 & 0.47 & 0.28 & 0.50 & 0.42 & 0.55 & 0.62 & 0.47 & 0.18 \\
Mistral-Small-4 & 0.71 & 0.65 & 0.31 & 0.59 & 0.56 & 0.58 & \underline{0.85} & 0.64 & 0.16 \\
Qwen3 & 0.70 & 0.47 & 0.38 & 0.64 & 0.58 & 0.68 & 0.82 & 0.69 & 0.44 \\
Qwen3-Next & \underline{0.74} & 0.67 & 0.66 & 0.63 & 0.57 & \textbf{\underline{0.80}} & 0.82 & \underline{0.74} & 0.28 \\
RL-Molstral-g16 & 0.71 & 0.72 & \underline{0.79} & \textbf{\underline{0.79}} & 0.57 & 0.57 & 0.71 & 0.63 & \underline{0.64} \\
RL-Molstral-g4 & 0.72 & \underline{0.72} & 0.36 & 0.61 & 0.55 & 0.62 & 0.82 & 0.58 & 0.15 \\
RL-Molstral-g8 & 0.74 & 0.70 & 0.64 & 0.63 & \textbf{\underline{0.63}} & 0.66 & 0.83 & 0.67 & 0.16 \\
gemma-3 & 0.62 & 0.53 & 0.21 & 0.33 & 0.55 & 0.54 & 0.82 & 0.57 & 0.16 \\
gemma-4 & \textbf{\underline{0.80}} & \textbf{\underline{0.86}} & \textbf{\underline{0.80}} & \underline{0.78} & \underline{0.59} & \underline{0.74} & \textbf{\underline{0.91}} & \textbf{\underline{0.79}} & 0.22 \\
\midrule
\midrule
 & \multicolumn{9}{c}{Avg. Precision} \\
\midrule
\midrule
ChemDFM-R & 0.54 & 0.85 & 0.17 & 0.29 & 0.55 & 0.52 & 0.72 & 0.55 & 0.16 \\
MiniMax-M2 & 0.56 & 0.64 & 0.15 & 0.30 & 0.41 & 0.52 & 0.61 & 0.44 & 0.12 \\
Mistral-Small-4 & 0.65 & 0.84 & 0.20 & 0.37 & 0.54 & 0.56 & 0.82 & 0.60 & 0.15 \\
Qwen3 & 0.68 & 0.79 & 0.21 & 0.39 & 0.55 & 0.65 & \underline{0.82} & 0.65 & \underline{0.17} \\
Qwen3-Next & \underline{0.69} & 0.85 & 0.25 & 0.39 & 0.55 & \textbf{\underline{0.74}} & 0.82 & \underline{0.69} & \textbf{\underline{0.17}} \\
RL-Molstral-g16 & 0.69 & \underline{0.86} & 0.20 & \underline{0.51} & 0.56 & 0.58 & 0.81 & 0.67 & 0.17 \\
RL-Molstral-g4 & 0.66 & 0.85 & 0.17 & 0.39 & 0.54 & 0.58 & 0.80 & 0.57 & 0.15 \\
RL-Molstral-g8 & 0.68 & 0.85 & \underline{0.25} & 0.38 & \textbf{\underline{0.58}} & 0.61 & 0.82 & 0.63 & 0.14 \\
gemma-3 & 0.59 & 0.82 & 0.18 & 0.30 & 0.54 & 0.53 & 0.80 & 0.56 & 0.16 \\
gemma-4 & \textbf{\underline{0.75}} & \textbf{\underline{0.88}} & \textbf{\underline{0.34}} & \textbf{\underline{0.51}} & \underline{0.56} & \underline{0.71} & \textbf{\underline{0.91}} & \textbf{\underline{0.73}} & 0.17 \\
\bottomrule
\end{tabular}

    \label{tab:comprehensive_cls_res}
\end{table}

\clearpage
\begin{sidewaystable}
    \caption{
        \textbf{Comprehensive Regression Results.}
    }
    \centering
    \begin{tabular}{c||ccccccccc}
\toprule
 & antiviral-potency & az-logd & az-ppb-clearance & caco2 & cyp3a4-novartis & fang-hclint & fang-hppb & fang-perm & fang-rclint \\
\midrule
\midrule
 & \multicolumn{9}{c}{PearsonR} \\
\midrule
\midrule
ChemDFM-R & 0.19 & \underline{0.44} & -0.04 & 0.44 & 0.00 & -0.04 & -0.27 & -0.02 & 0.07 \\
MiniMax-M2 & 0.37 & 0.09 & 0.07 & 0.39 & \underline{0.22} & 0.13 & 0.57 & 0.18 & 0.08 \\
Mistral-Small-4 & 0.36 & -0.00 & 0.27 & 0.51 & \textbf{\underline{0.28}} & 0.13 & \textbf{\underline{0.70}} & 0.16 & 0.04 \\
Qwen3 & 0.33 & -0.01 & 0.08 & 0.44 & 0.04 & 0.05 & 0.46 & 0.16 & 0.00 \\
Qwen3-Next & \underline{0.40} & 0.43 & \underline{0.34} & \textbf{\underline{0.66}} & 0.13 & \underline{0.14} & 0.53 & \underline{0.19} & \underline{0.23} \\
RL-Molstral-g16 & 0.16 & 0.39 & 0.07 & 0.41 & 0.03 & 0.02 & 0.08 & 0.00 & 0.05 \\
gemma-3 & 0.38 & 0.35 & 0.18 & 0.49 & 0.15 & -0.05 & 0.31 & 0.11 & -0.12 \\
gemma-4 & \textbf{\underline{0.53}} & \textbf{\underline{0.62}} & \textbf{\underline{0.55}} & \underline{0.64} & 0.10 & \textbf{\underline{0.29}} & \underline{0.61} & \textbf{\underline{0.22}} & \textbf{\underline{0.33}} \\
\midrule
\midrule
 & \multicolumn{9}{c}{SpearmanR} \\
\midrule
\midrule
ChemDFM-R & 0.19 & 0.48 & -0.11 & 0.38 & 0.13 & 0.07 & -0.50 & 0.04 & 0.08 \\
MiniMax-M2 & \underline{0.39} & 0.35 & 0.18 & 0.53 & 0.22 & \underline{0.15} & \underline{0.61} & \textbf{\underline{0.26}} & 0.09 \\
Mistral-Small-4 & 0.37 & \underline{0.50} & 0.24 & 0.47 & 0.25 & 0.15 & \textbf{\underline{0.68}} & 0.20 & 0.04 \\
Qwen3 & 0.34 & 0.43 & 0.13 & 0.46 & \underline{0.27} & 0.06 & 0.41 & 0.15 & -0.02 \\
Qwen3-Next & 0.36 & 0.42 & \underline{0.33} & \textbf{\underline{0.67}} & \textbf{\underline{0.34}} & 0.12 & 0.55 & 0.22 & \underline{0.22} \\
RL-Molstral-g16 & 0.13 & 0.38 & 0.10 & 0.34 & 0.05 & 0.00 & 0.01 & 0.00 & 0.06 \\
gemma-3 & 0.38 & 0.36 & 0.17 & 0.45 & 0.13 & -0.05 & 0.32 & 0.09 & -0.11 \\
gemma-4 & \textbf{\underline{0.50}} & \textbf{\underline{0.61}} & \textbf{\underline{0.56}} & \underline{0.63} & 0.12 & \textbf{\underline{0.30}} & 0.52 & \underline{0.23} & \textbf{\underline{0.33}} \\
\bottomrule
\end{tabular}
\\
    \begin{tabular}{c||ccccccccc}
\toprule
 & fang-rppb & fang-solubility & half-life & hep-clearance-az & ld50 & lipophilicity & mic-clearance-az & solubility & vdss \\
\midrule
\midrule
 & \multicolumn{9}{c}{PearsonR} \\
\midrule
\midrule
ChemDFM-R & -0.03 & -0.15 & -0.04 & 0.06 & -0.03 & \textbf{\underline{0.52}} & \textbf{\underline{0.20}} & 0.08 & 0.31 \\
MiniMax-M2 & \underline{0.42} & 0.16 & \textbf{\underline{0.53}} & -0.03 & -0.20 & 0.35 & 0.10 & 0.82 & \underline{0.44} \\
Mistral-Small-4 & \textbf{\underline{0.46}} & \underline{0.31} & 0.40 & 0.04 & 0.10 & \underline{0.43} & \underline{0.16} & \underline{0.83} & 0.43 \\
Qwen3 & 0.04 & 0.24 & 0.17 & \underline{0.17} & -0.33 & 0.36 & 0.08 & 0.76 & 0.24 \\
Qwen3-Next & -0.02 & 0.29 & \underline{0.49} & -0.01 & -0.49 & 0.39 & 0.11 & 0.79 & 0.28 \\
RL-Molstral-g16 & -0.24 & 0.10 & 0.38 & -0.06 & \textbf{\underline{0.41}} & 0.38 & 0.08 & 0.78 & 0.31 \\
gemma-3 & -0.20 & -0.02 & 0.19 & 0.16 & \underline{0.12} & 0.35 & -0.05 & 0.51 & 0.11 \\
gemma-4 & 0.42 & \textbf{\underline{0.41}} & 0.48 & \textbf{\underline{0.17}} & -0.20 & 0.30 & 0.05 & \textbf{\underline{0.86}} & \textbf{\underline{0.48}} \\
\midrule
\midrule
 & \multicolumn{9}{c}{SpearmanR} \\
\midrule
\midrule
ChemDFM-R & 0.01 & -0.13 & 0.04 & 0.05 & \underline{0.14} & \textbf{\underline{0.53}} & \textbf{\underline{0.23}} & -0.07 & 0.34 \\
MiniMax-M2 & 0.24 & 0.14 & 0.55 & 0.05 & -0.23 & 0.33 & 0.13 & \underline{0.87} & \textbf{\underline{0.51}} \\
Mistral-Small-4 & \textbf{\underline{0.46}} & \underline{0.29} & 0.39 & 0.09 & 0.07 & \underline{0.40} & 0.18 & 0.86 & 0.44 \\
Qwen3 & 0.03 & 0.24 & 0.20 & \textbf{\underline{0.17}} & -0.35 & 0.33 & 0.11 & 0.78 & 0.32 \\
Qwen3-Next & -0.06 & 0.27 & \underline{0.55} & -0.03 & -0.52 & 0.38 & \underline{0.21} & 0.83 & 0.37 \\
RL-Molstral-g16 & -0.00 & 0.11 & \textbf{\underline{0.58}} & -0.07 & \textbf{\underline{0.46}} & 0.34 & 0.08 & 0.82 & 0.46 \\
gemma-3 & -0.20 & 0.01 & 0.17 & \underline{0.16} & 0.14 & 0.32 & -0.08 & 0.43 & 0.09 \\
gemma-4 & \underline{0.35} & \textbf{\underline{0.40}} & 0.52 & 0.16 & -0.19 & 0.26 & 0.08 & \textbf{\underline{0.88}} & \underline{0.51} \\
\bottomrule
\end{tabular}

    \label{tab:comprehensive_reg_res}
\end{sidewaystable}

\FloatBarrier

\subsection{Effect of Reinforcement learning}
\label{app:molstral-propred}

\begin{figure}[H]
    \centering
    \includegraphics[width=1\linewidth]{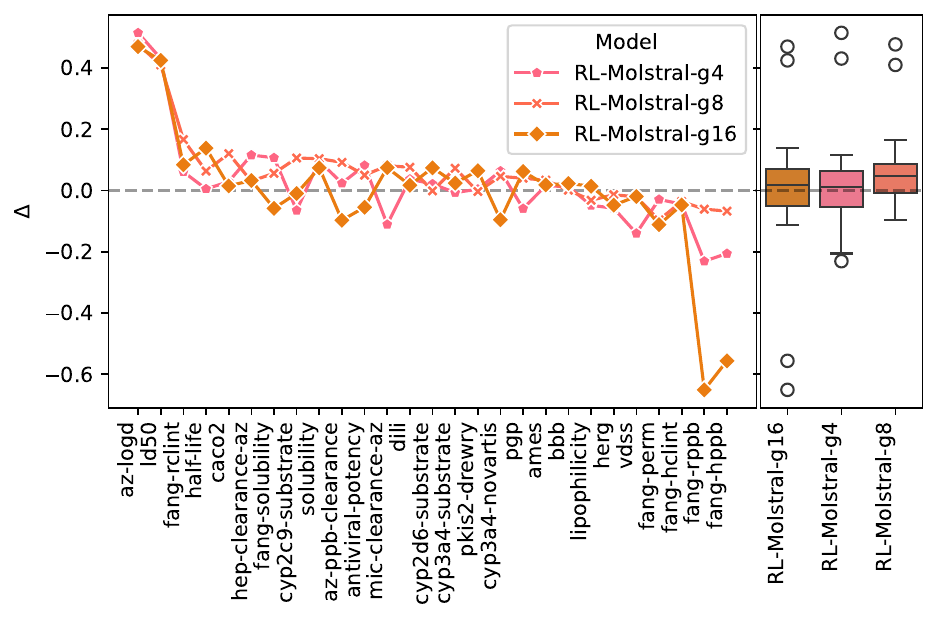}
    \caption{
        \textbf{Effect of reinforcement learning on the molecular property prediction performance.}
        We display the difference $\Delta$ in performance between RL-Mistral and Mistral-Small-4 on all classification and regression tasks.
    }
    \label{fig:molecular_proppred_delta}
\end{figure}

RL-Molstral's training was heavily focused on the generation task, with 66\% of the prompts sampled from the molecular generation task, and 33\% from the molecular property prediction task.
As such, the improvement over Mistral-Small-4 on the molecular property prediction tasks is limited, with all RL-Molstral models outperforming Mistral-Small-4 on 60\% of the tasks, and the highest improvements observed for the regression tasks, notably with the logd and ld50 regression tasks.

For most tasks, the performance of RL-Molstral is slightly better than Mistral-Small-4, however for two tasks: fang-rppb and fang-hppb, the performance decreases, which are also by far the two smallest tasks in our dataset.

\FloatBarrier

\FloatBarrier
\subsection{Docking Results Illustrations}
\label{appendix:docking_results_illustrations}

\begin{figure}[H]
    \centering
    \begin{subfigure}[b]{0.24\textwidth}
        \centering
        \includegraphics[trim=8cm 8cm 8cm 8cm, clip, width=\textwidth]{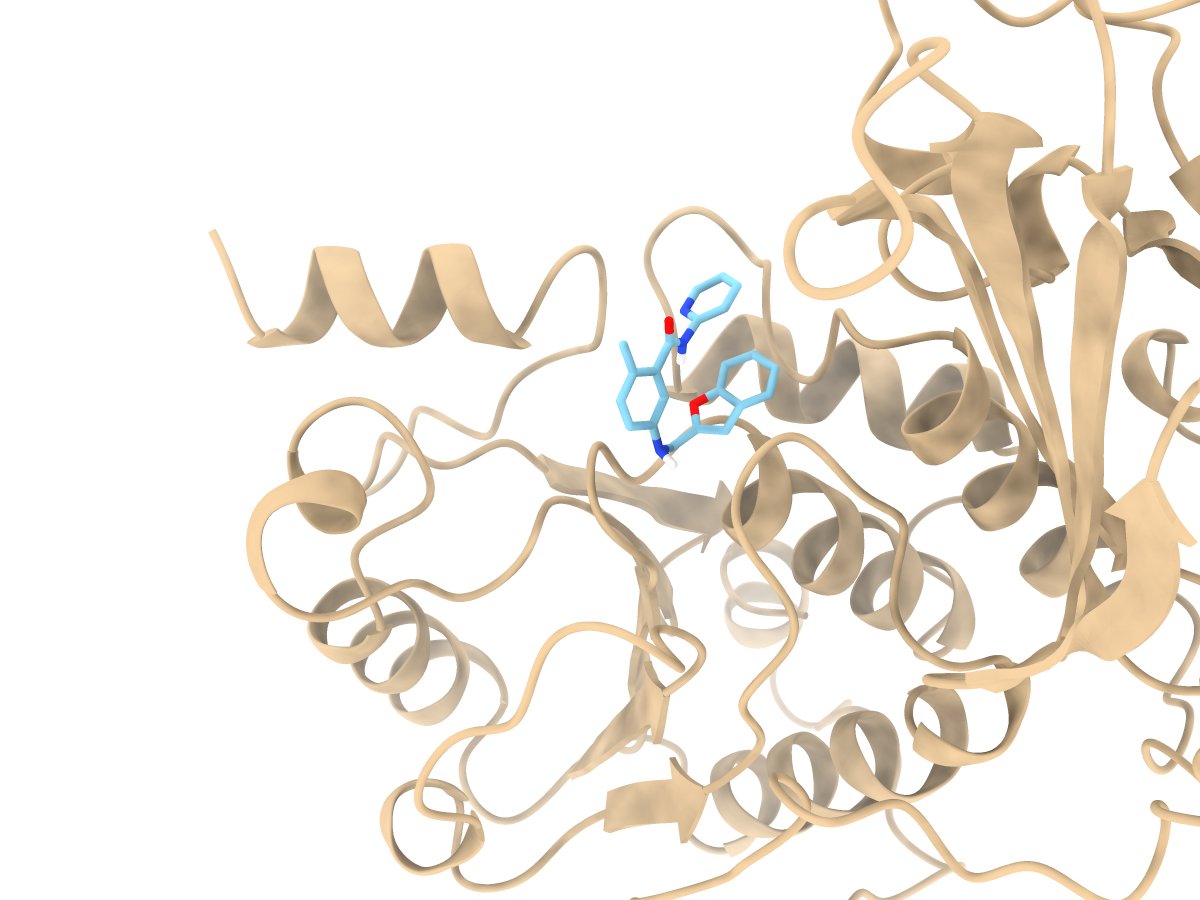}\\
        Reward: 0.44\\
        \vspace{0.1cm}
        \includegraphics[trim=8cm 8cm 8cm 8cm, clip, width=\textwidth]{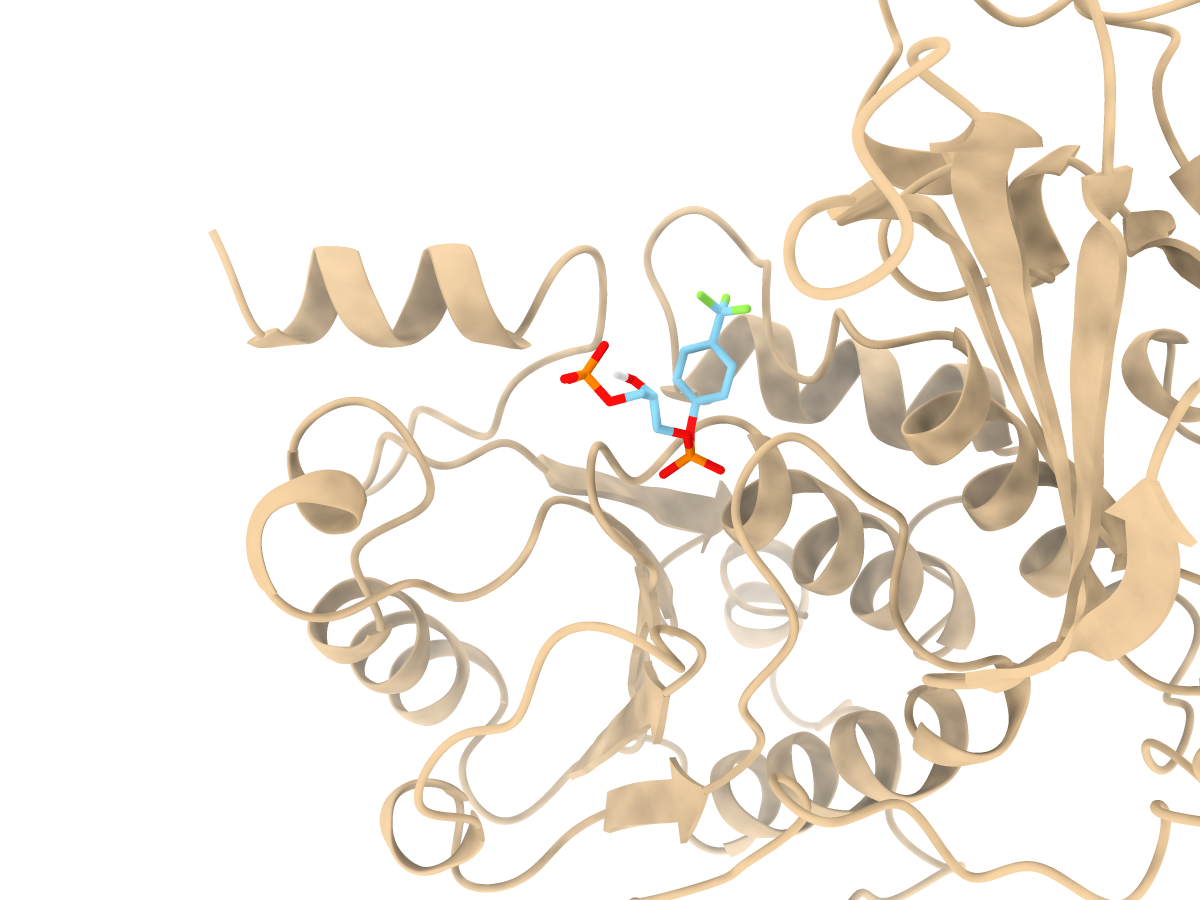}\\
        Reward: 0.89\\
        \vspace{0.1cm}
        \includegraphics[trim=8cm 8cm 8cm 8cm, clip, width=\textwidth]{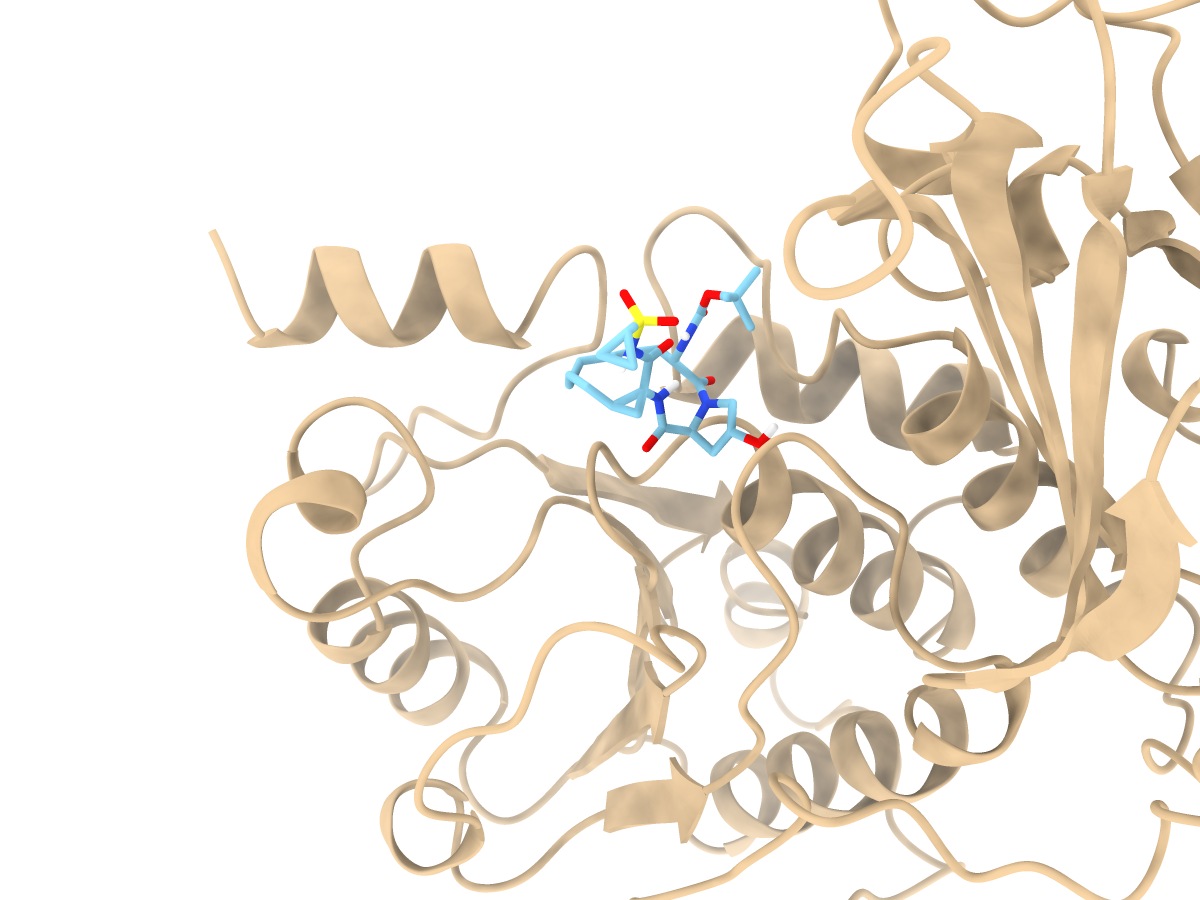}\\
        Reward: 0.0
        \caption{ChemDFM-R}
    \end{subfigure}
    \hfill
    \begin{subfigure}[b]{0.24\textwidth}
        \centering
        \includegraphics[trim=8cm 8cm 8cm 8cm, clip, width=\textwidth]{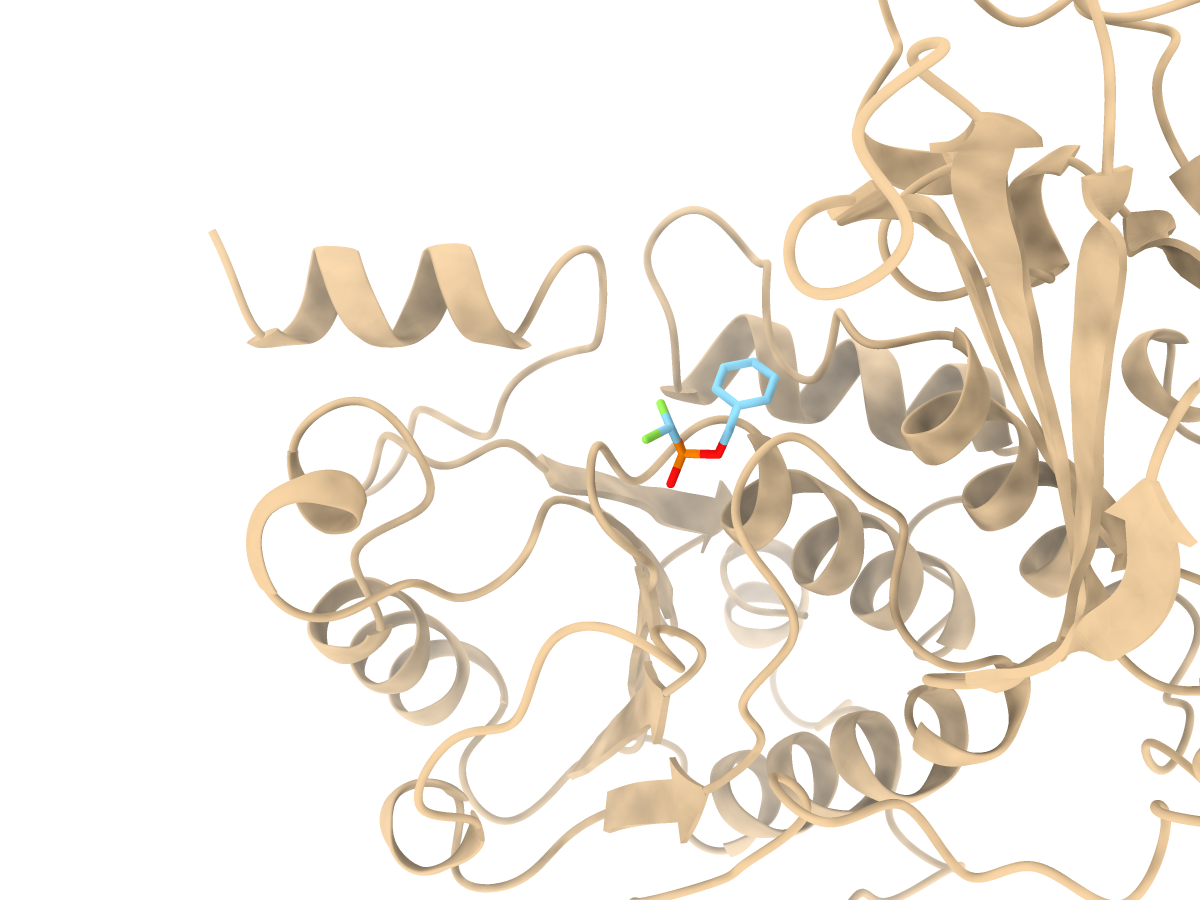}\\
        Reward: 0.57\\
        \vspace{0.1cm}
        \includegraphics[trim=8cm 8cm 8cm 8cm, clip, width=\textwidth]{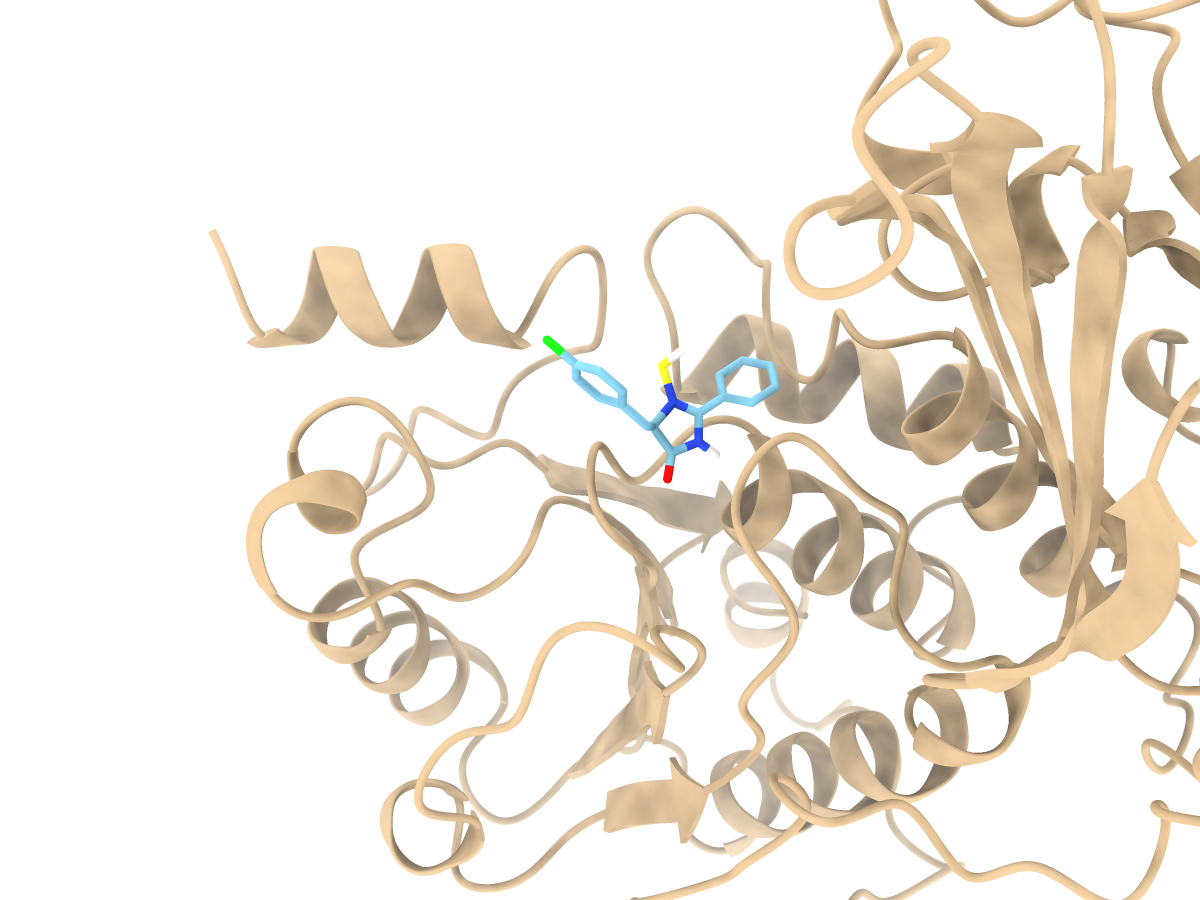}\\
        Reward: 0.36\\
        \vspace{0.1cm}
        \includegraphics[trim=8cm 8cm 8cm 8cm, clip, width=\textwidth]{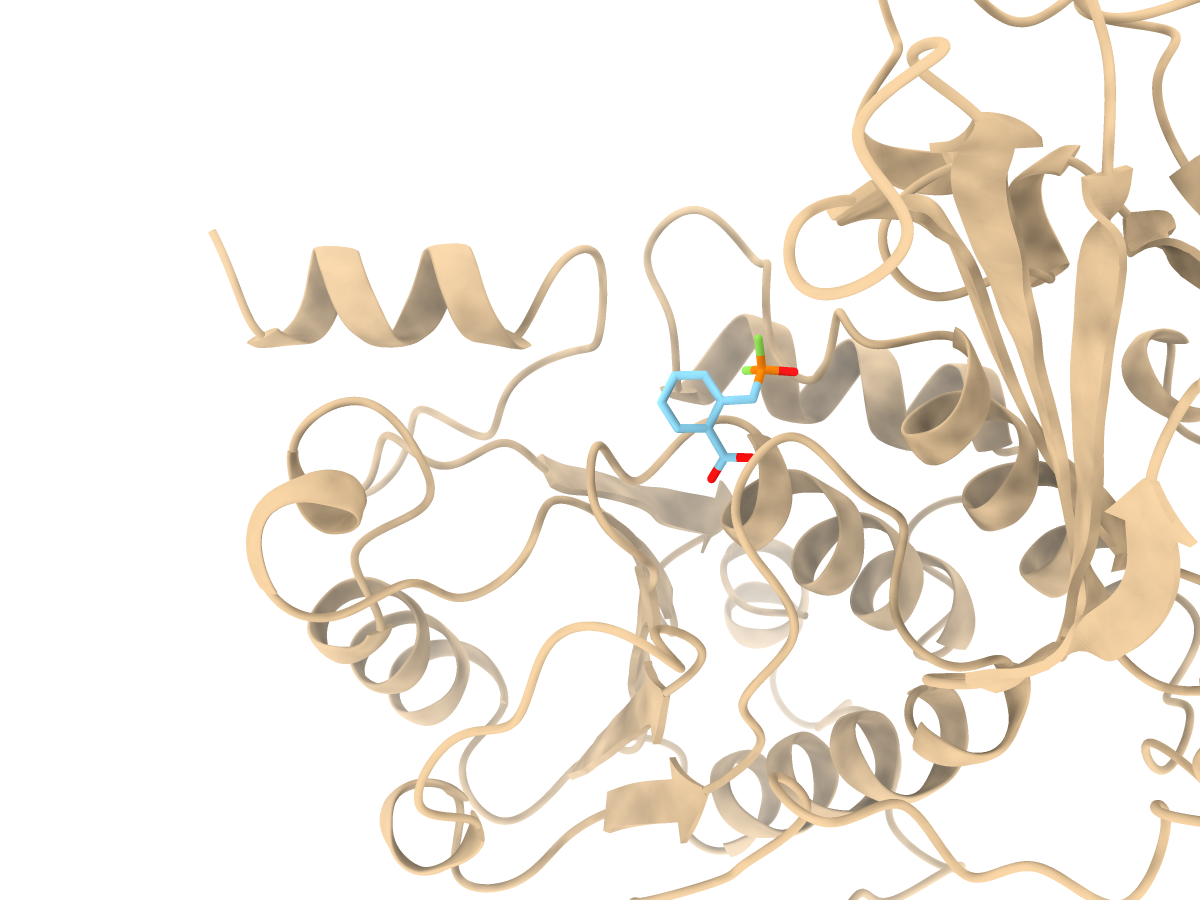}\\
        Reward: 0.64
        \caption{Mistral-Small}
    \end{subfigure}
    \hfill
    \begin{subfigure}[b]{0.24\textwidth}
        \centering
        \includegraphics[trim=8cm 8cm 8cm 8cm, clip, width=\textwidth]{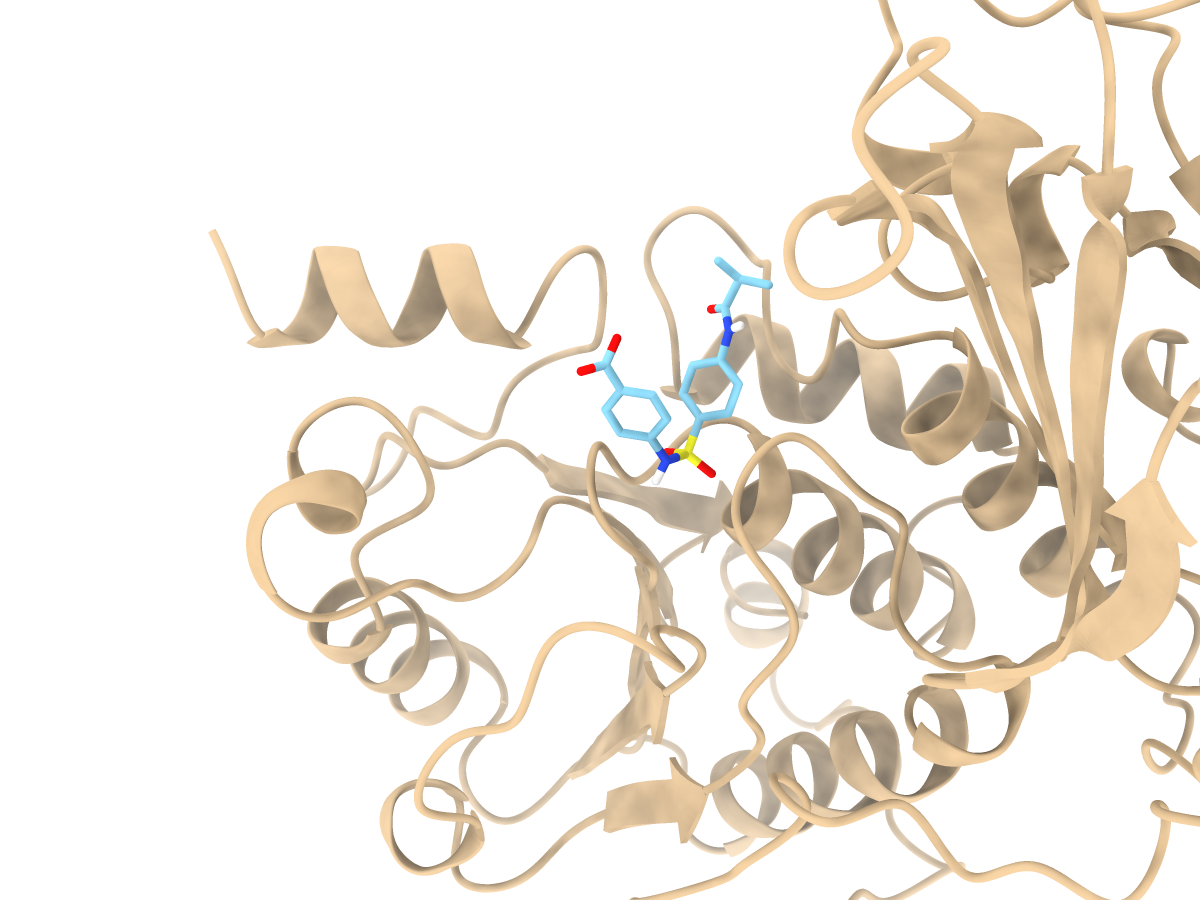}\\
        Reward: 0.77\\
        \vspace{0.1cm}
        \includegraphics[trim=8cm 8cm 8cm 8cm, clip, width=\textwidth]{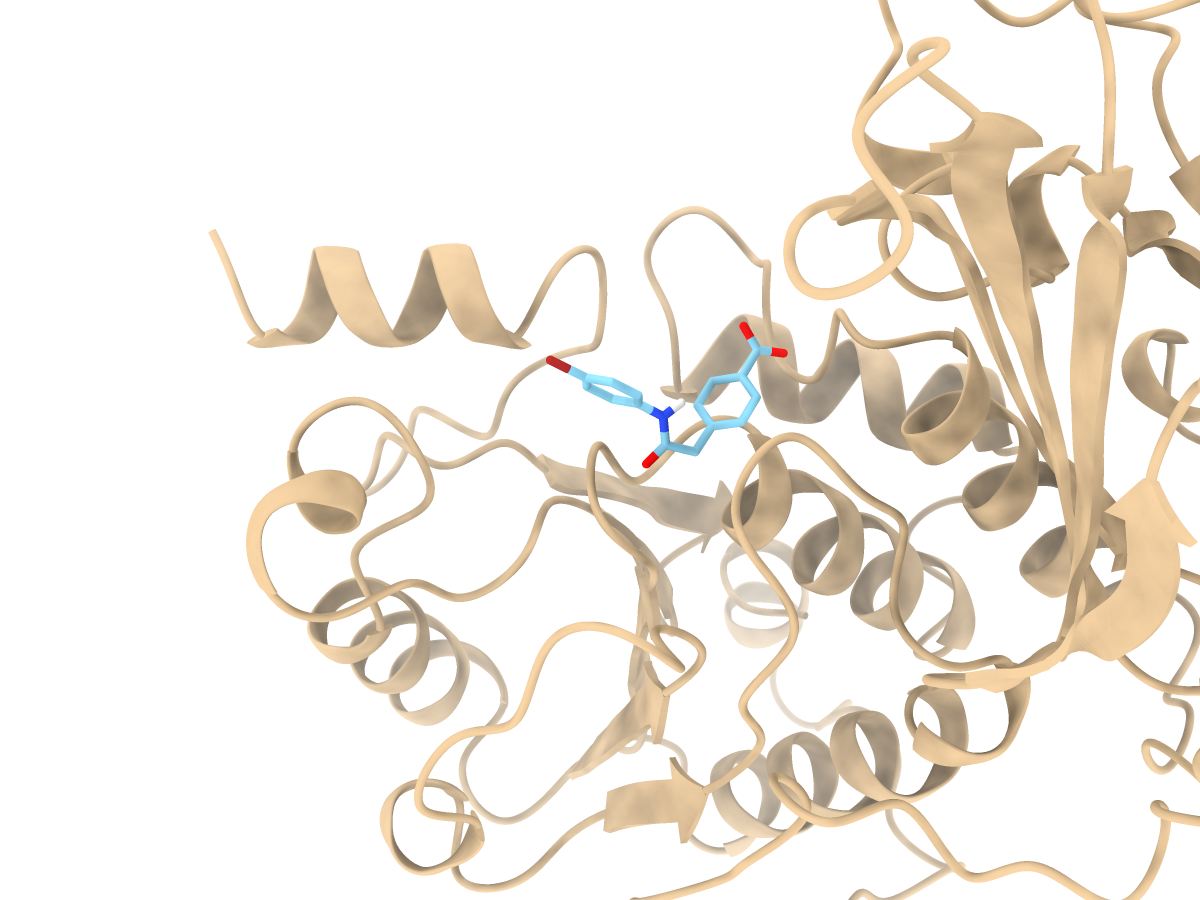}\\
        Reward: 0.62\\
        \vspace{0.1cm}
        \includegraphics[trim=8cm 8cm 8cm 8cm, clip, width=\textwidth]{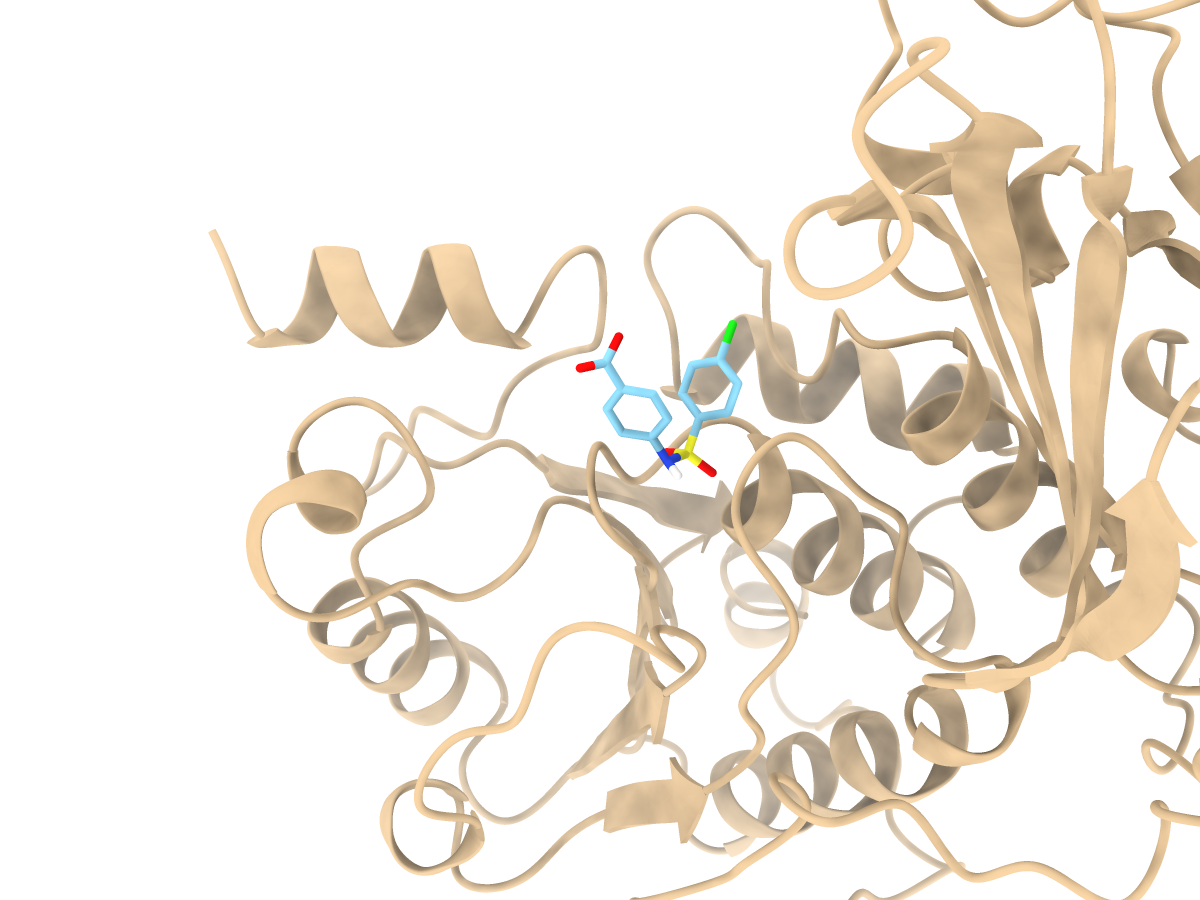}\\
        Reward: 0.75
        \caption{RL-Molstral}
    \end{subfigure}
    \hfill
    \begin{subfigure}[b]{0.24\textwidth}
        \centering
        \includegraphics[trim=8cm 8cm 8cm 8cm, clip, width=\textwidth]{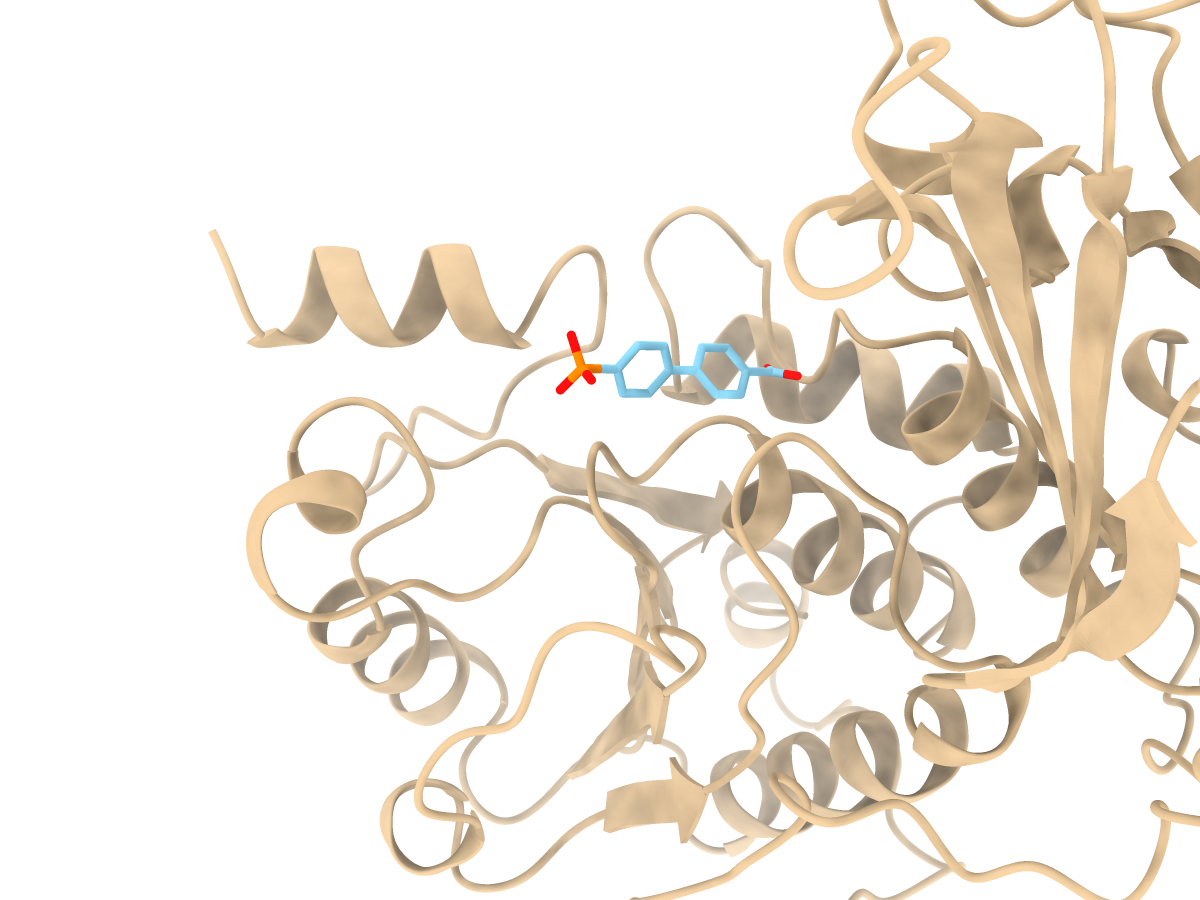}\\
        Reward: 0.94\\
        \vspace{0.1cm}
        \includegraphics[trim=8cm 8cm 8cm 8cm, clip, width=\textwidth]{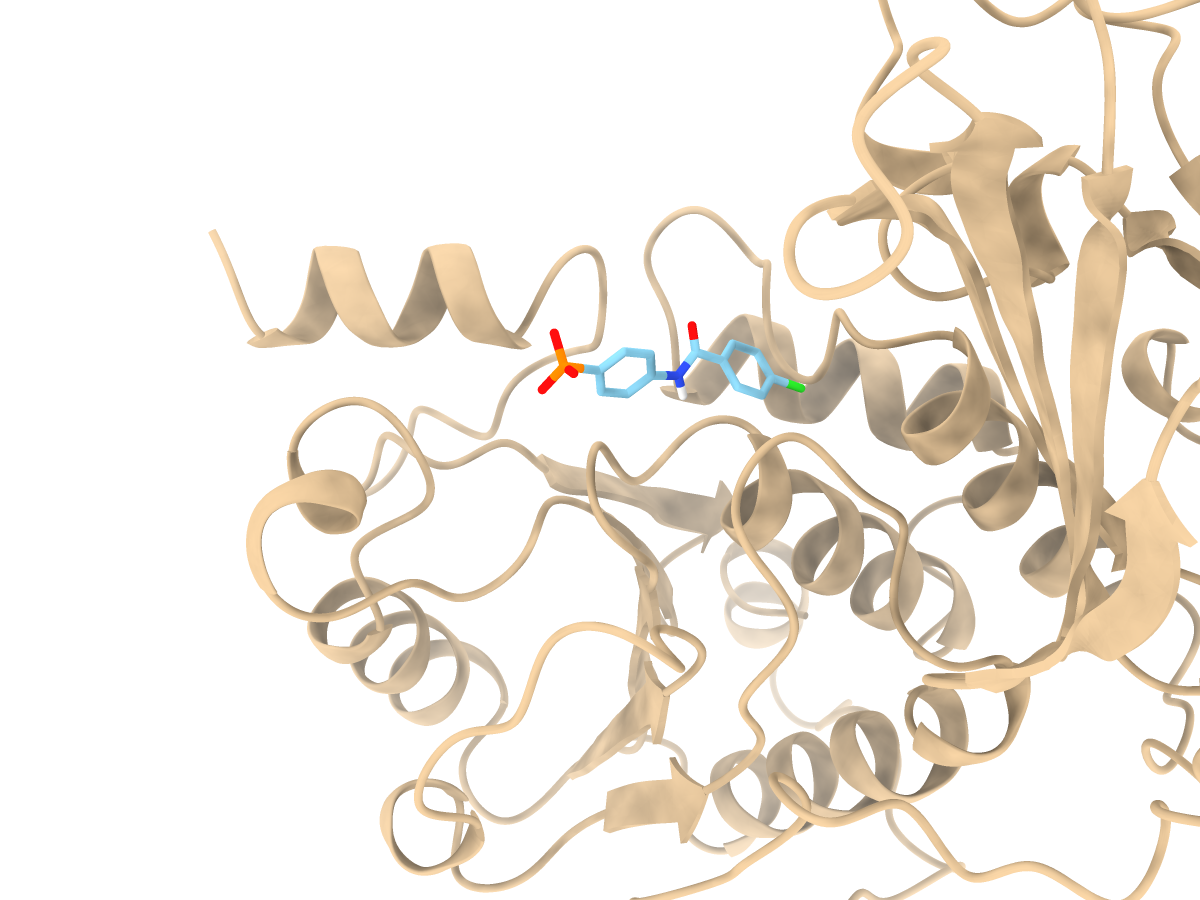}\\
        Reward: 0.50\\
        \vspace{0.1cm}
        \includegraphics[trim=8cm 8cm 8cm 8cm, clip, width=\textwidth]{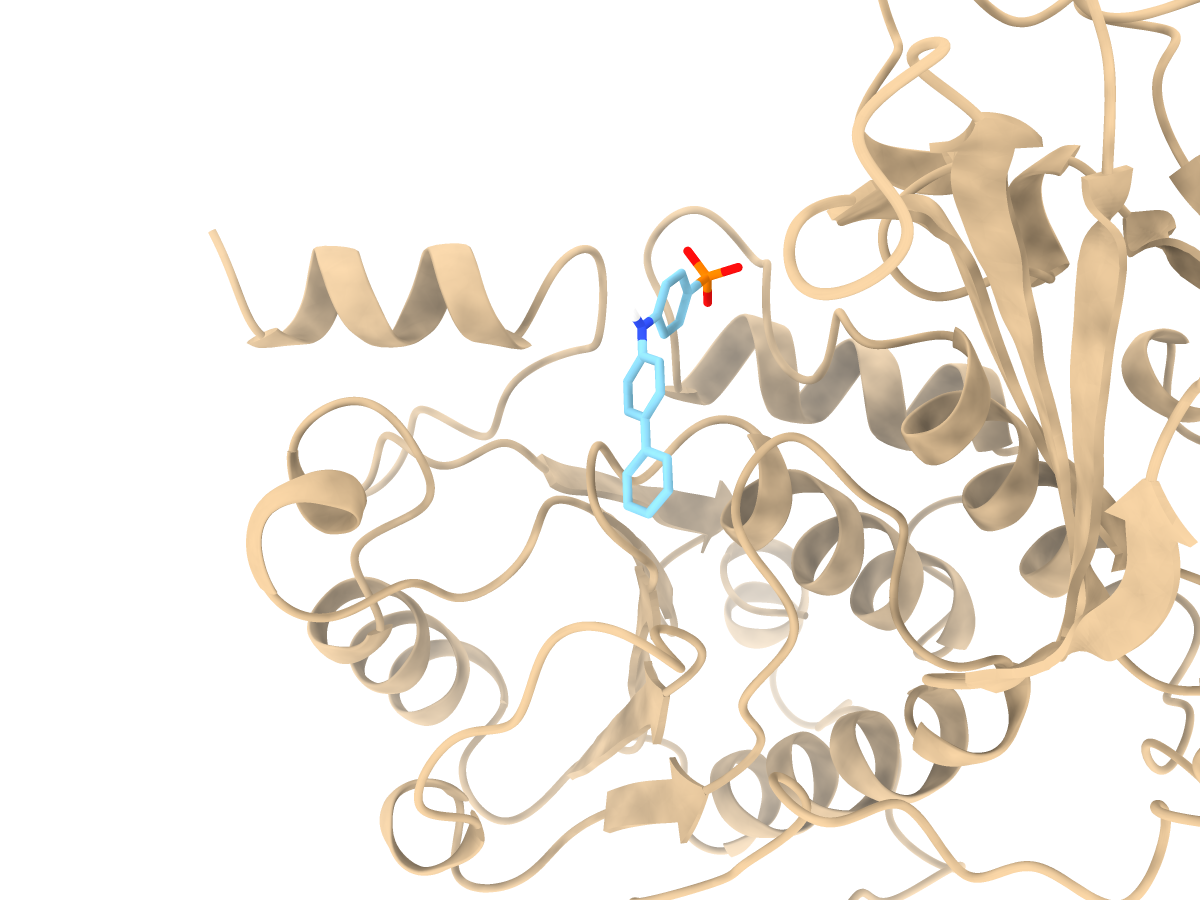}\\
        Reward: 0.38
        \caption{gemma}
    \end{subfigure}
    \caption{Docking poses for sample 772296 generated by different models trying to minimize the docking score to this target.}
    \label{fig:docking_sample_772296}
\end{figure}

\begin{figure}[H]
    \centering
    \begin{subfigure}[b]{0.24\textwidth}
        \centering
        \includegraphics[trim=8cm 8cm 12cm 8cm, clip, width=\textwidth]{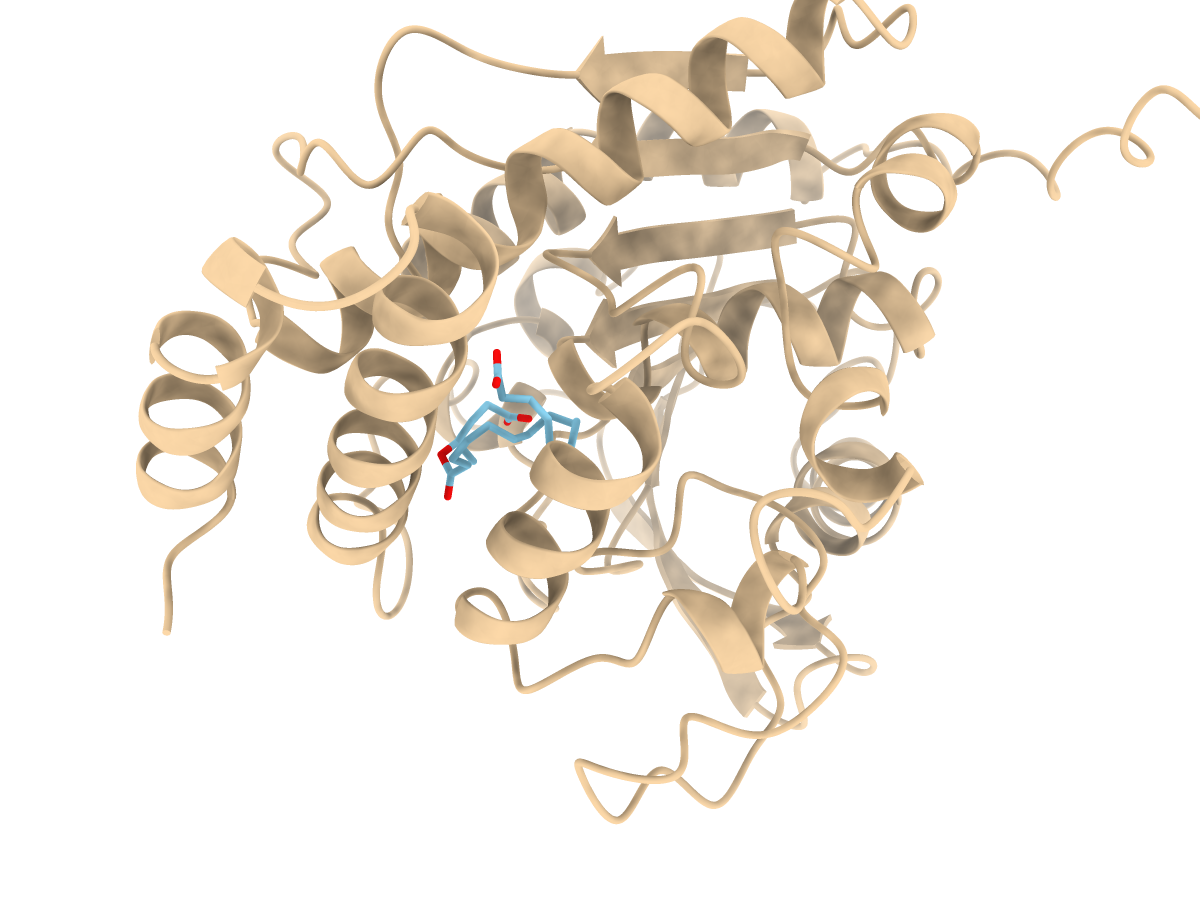}\\
        Reward: 0.72\\
        \vspace{0.1cm}
        \includegraphics[trim=8cm 8cm 12cm 8cm, clip, width=\textwidth]{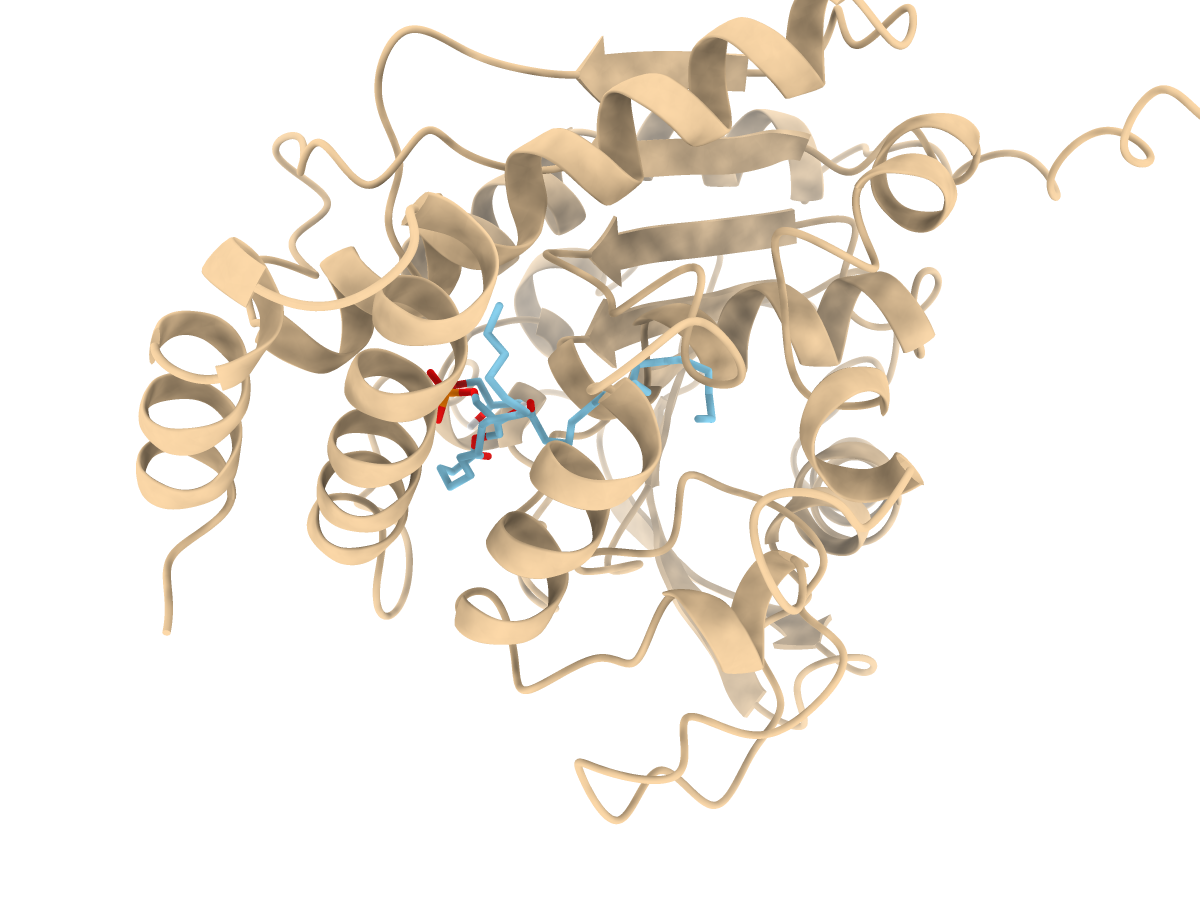}\\
        Reward: 0.0\\
        \vspace{0.1cm}
        \includegraphics[trim=8cm 8cm 12cm 8cm, clip, width=\textwidth]{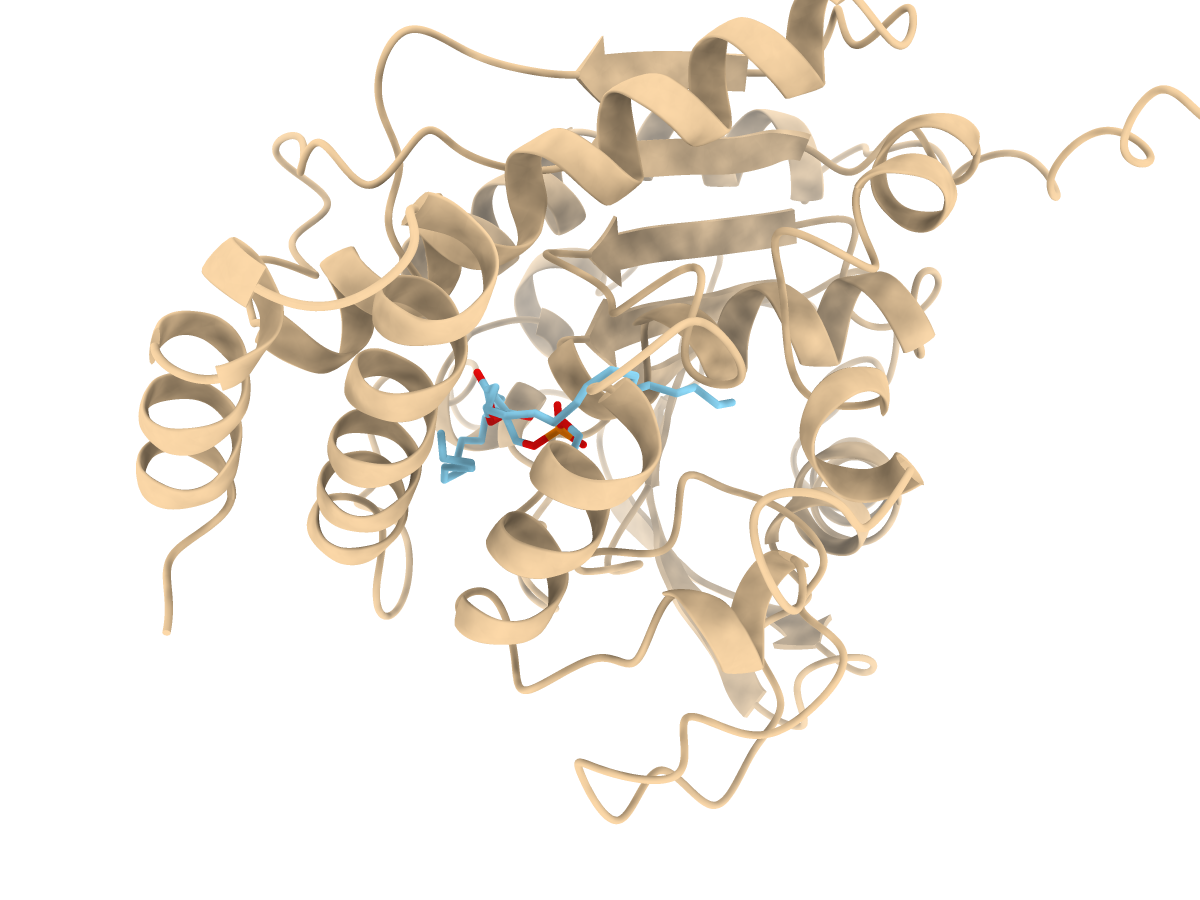}\\
        Reward: 0.0
        \caption{ChemDFM-R}
    \end{subfigure}
    \hfill
    \begin{subfigure}[b]{0.24\textwidth}
        \centering
        \includegraphics[trim=8cm 8cm 12cm 8cm, clip, width=\textwidth]{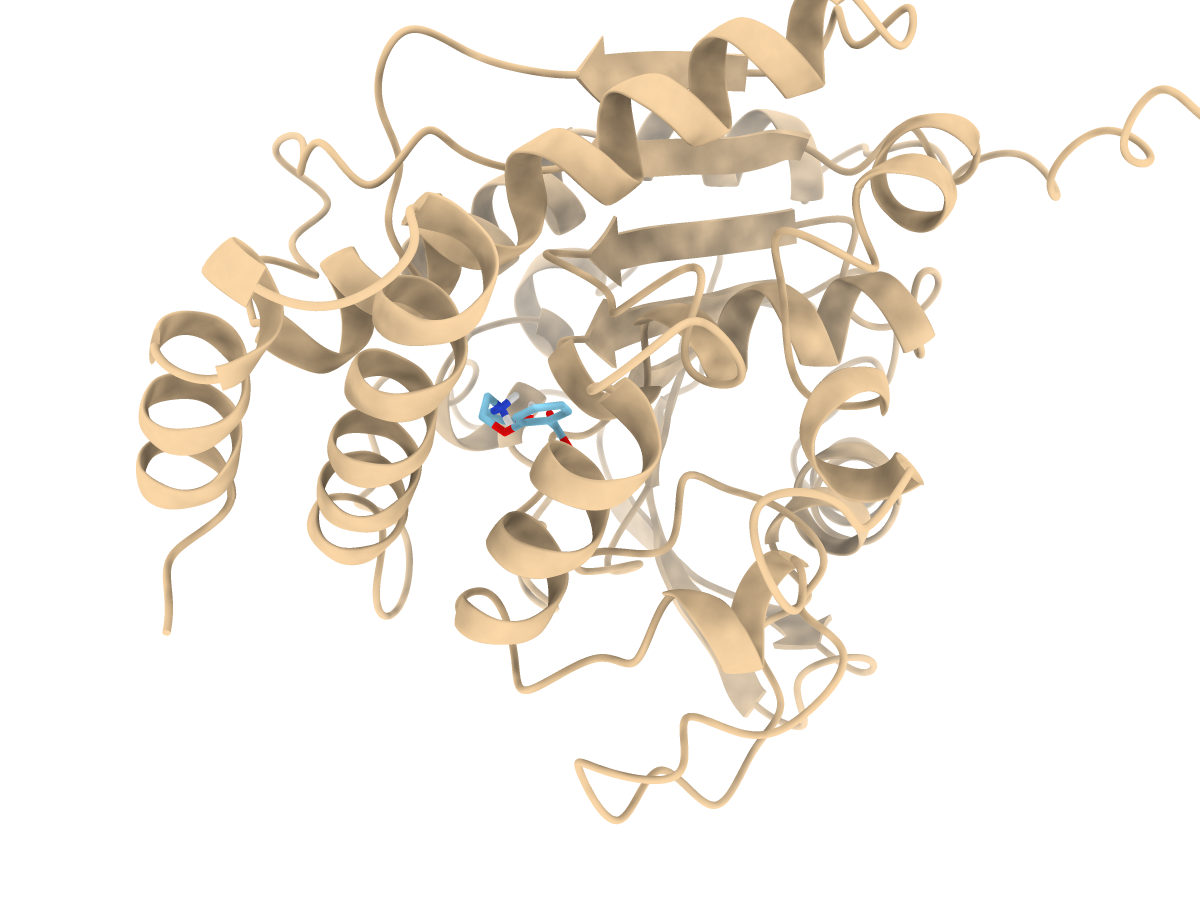}\\
        Reward: 0.31\\
        \vspace{0.1cm}
        \includegraphics[trim=8cm 8cm 12cm 8cm, clip, width=\textwidth]{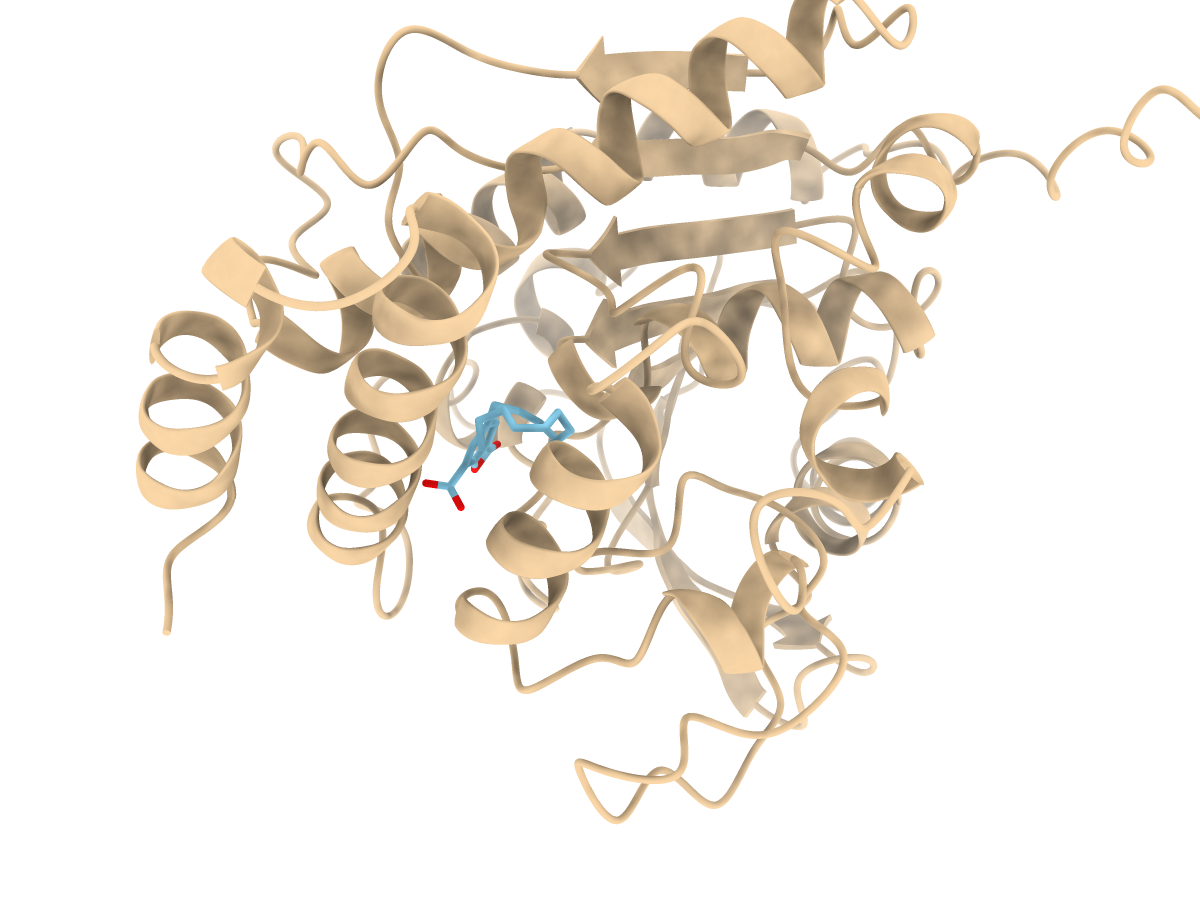}\\
        Reward: 0.86\\
        \vspace{0.1cm}
        \includegraphics[trim=8cm 8cm 12cm 8cm, clip, width=\textwidth]{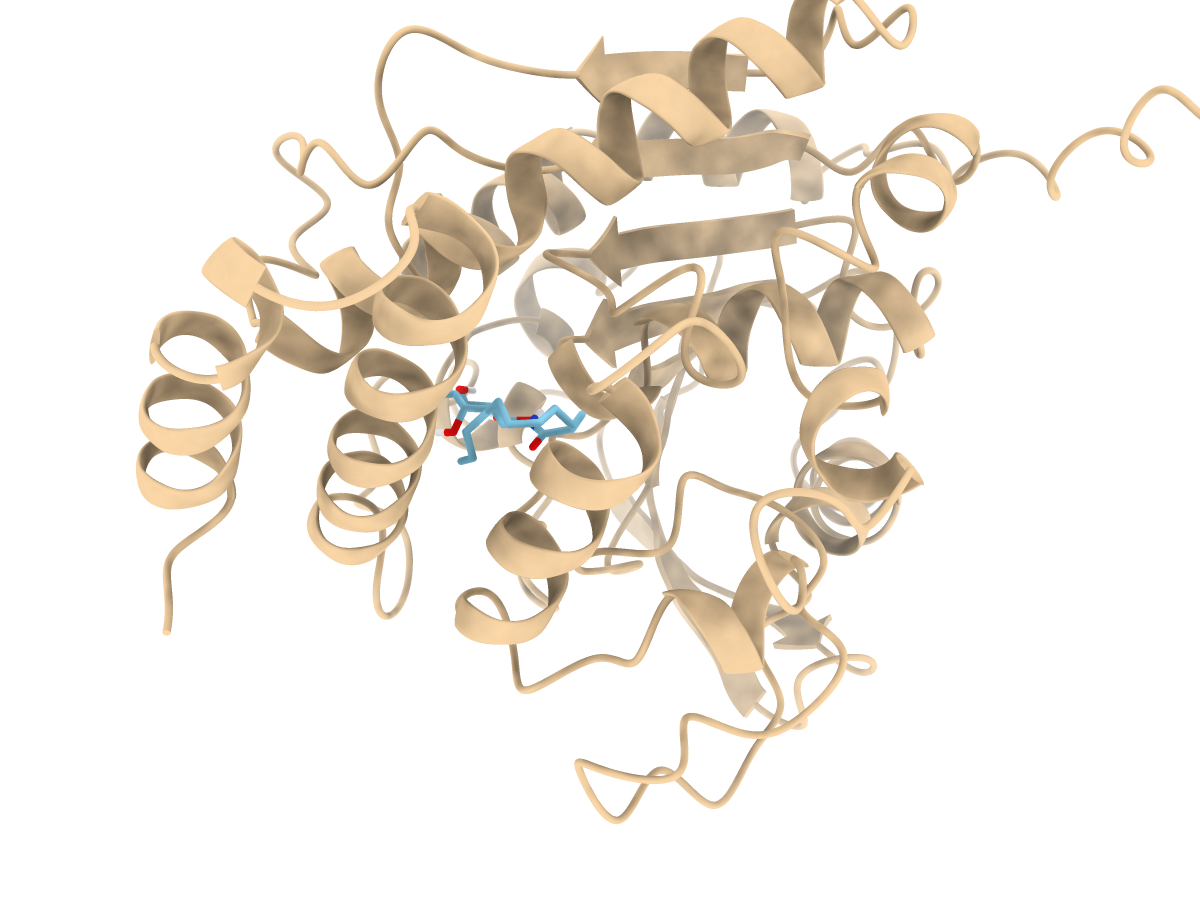}\\
        Reward: 0.56
        \caption{Mistral-Small}
    \end{subfigure}
    \hfill
    \begin{subfigure}[b]{0.24\textwidth}
        \centering
        \includegraphics[trim=8cm 8cm 12cm 8cm, clip, width=\textwidth]{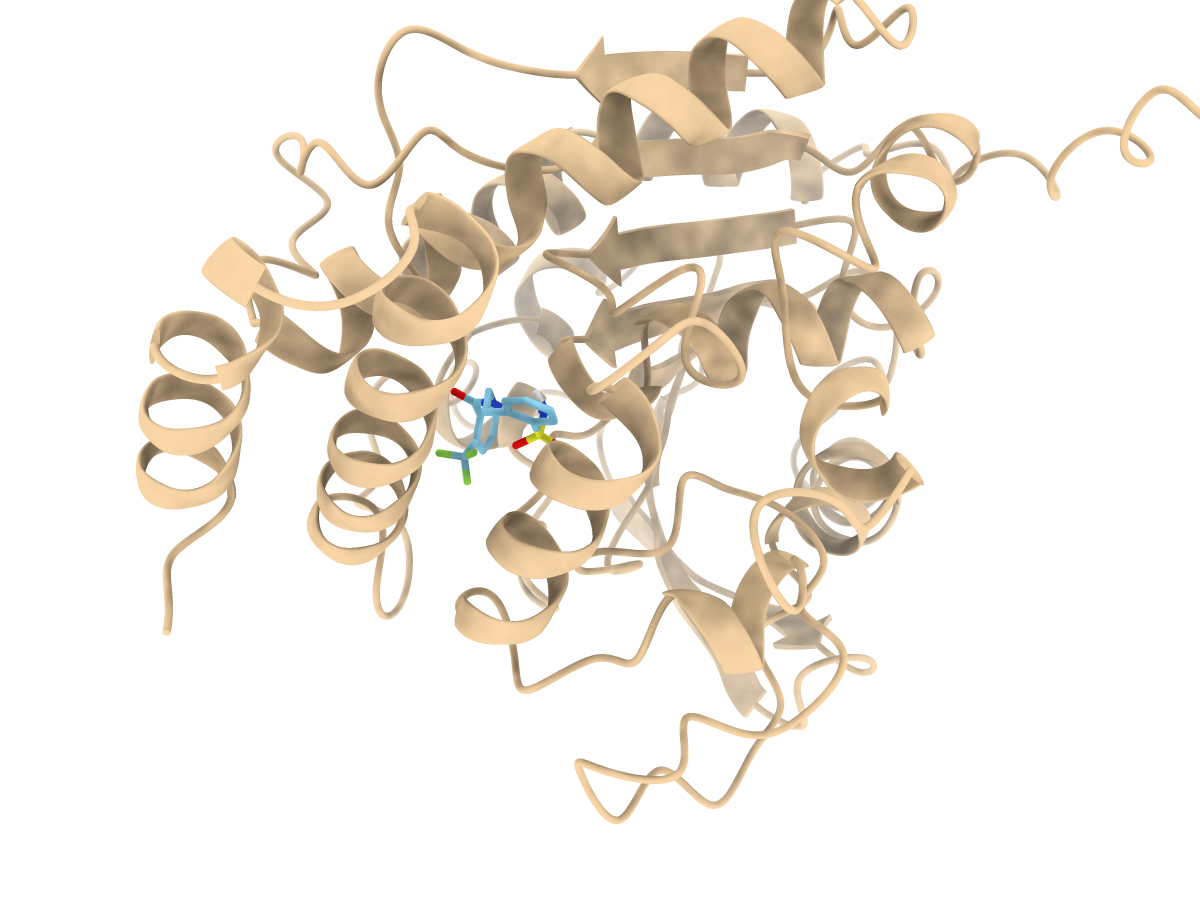}\\
        Reward: 0.78\\
        \vspace{0.1cm}
        \includegraphics[trim=8cm 8cm 12cm 8cm, clip, width=\textwidth]{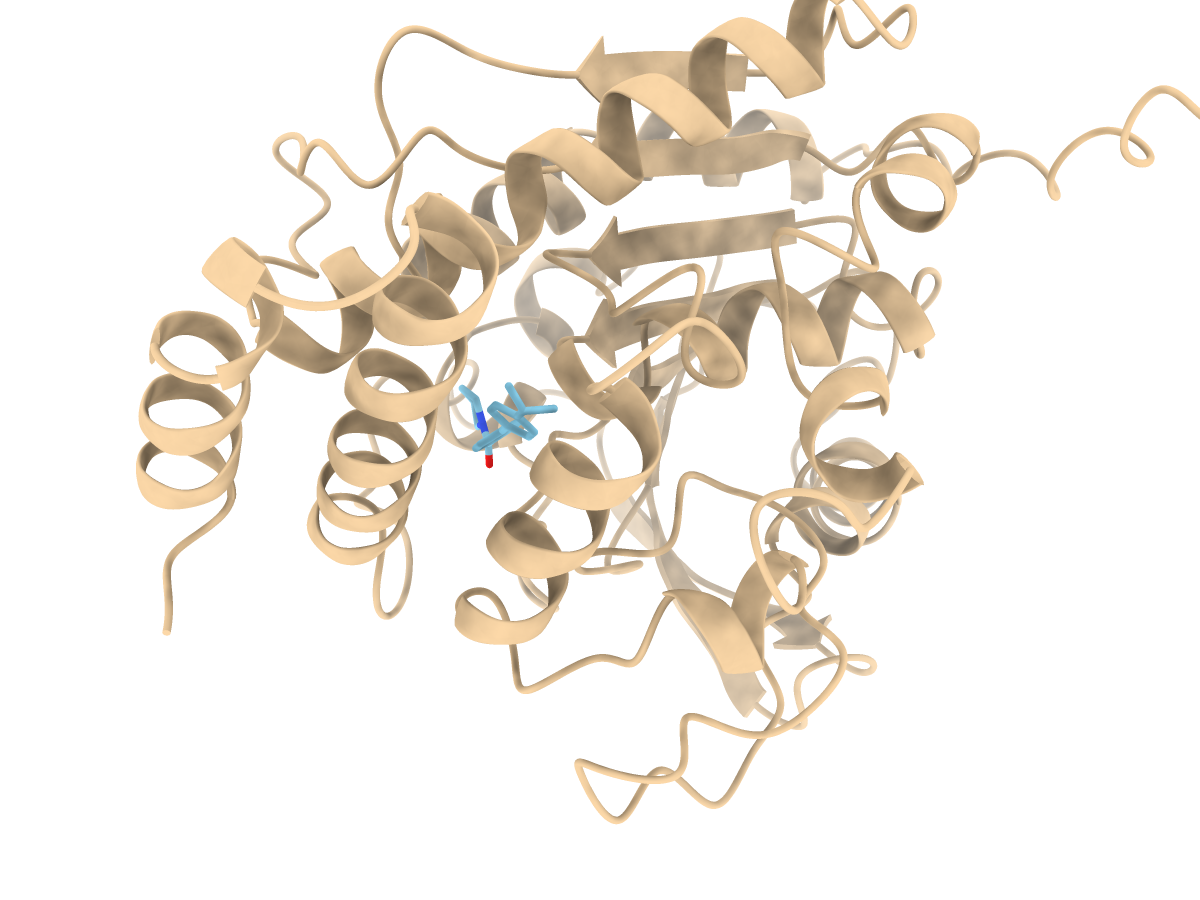}\\
        Reward: 0.84\\
        \vspace{0.1cm}
        \includegraphics[trim=8cm 8cm 12cm 8cm, clip, width=\textwidth]{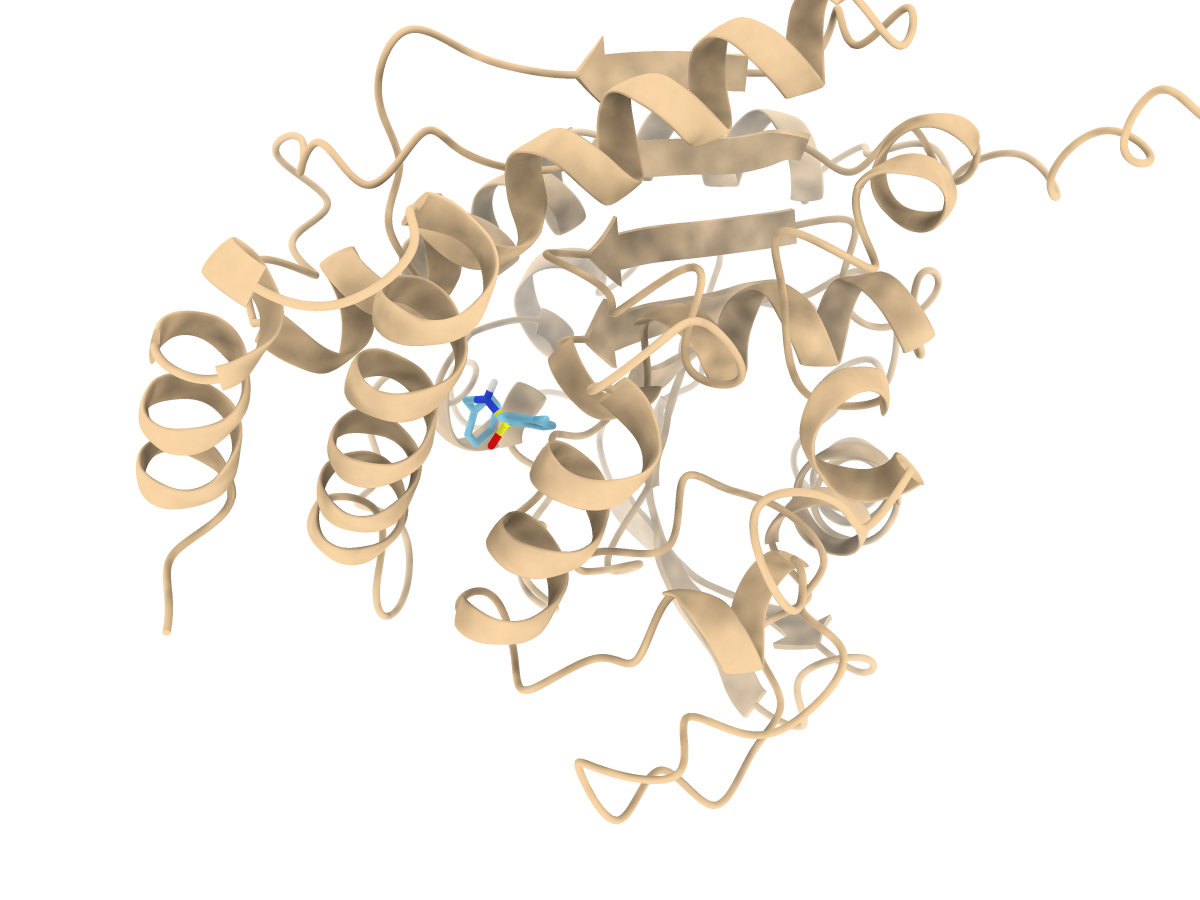}\\
        Reward: 0.80
        \caption{RL-Molstral}
    \end{subfigure}
    \hfill
    \begin{subfigure}[b]{0.24\textwidth}
        \centering
        \includegraphics[trim=8cm 8cm 12cm 8cm, clip, width=\textwidth]{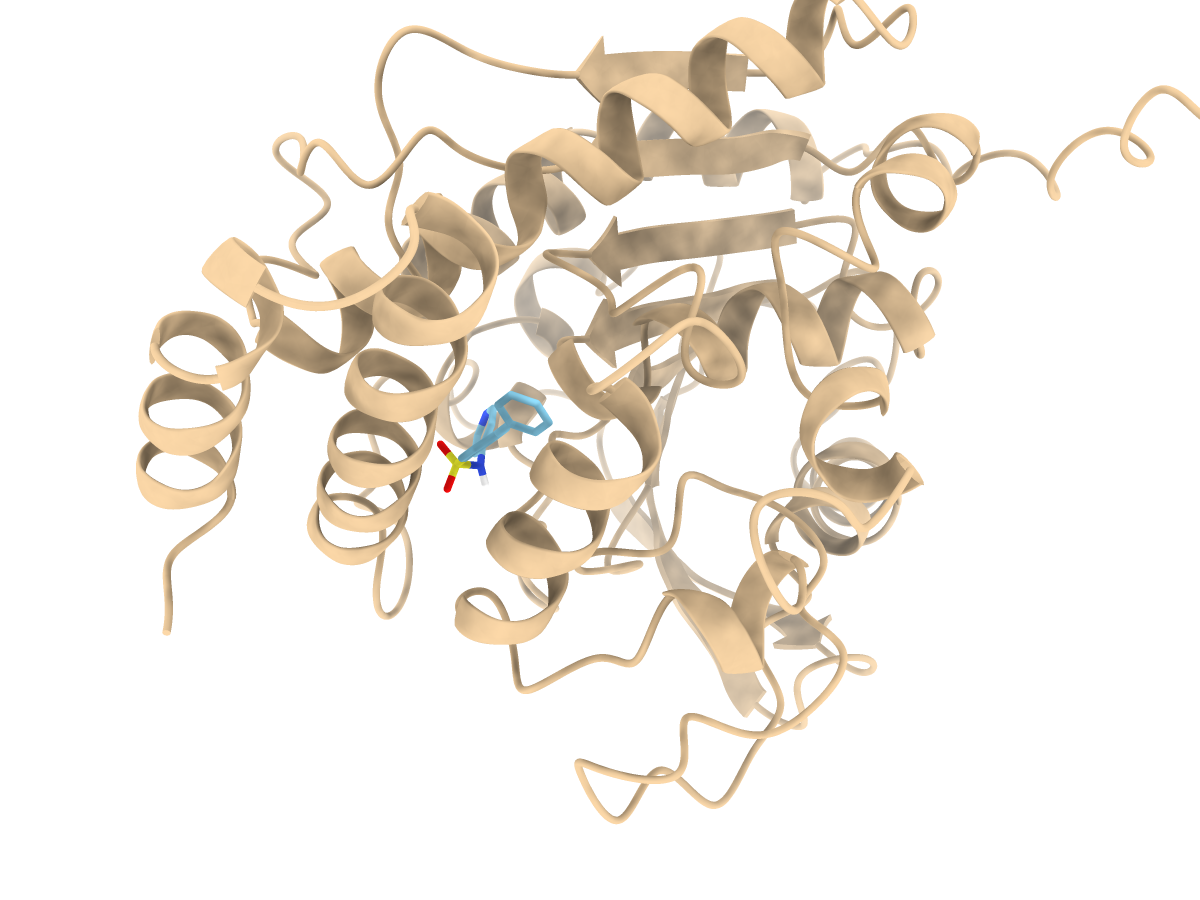}\\
        Reward: 0.85\\
        \vspace{0.1cm}
        \includegraphics[trim=8cm 8cm 12cm 8cm, clip, width=\textwidth]{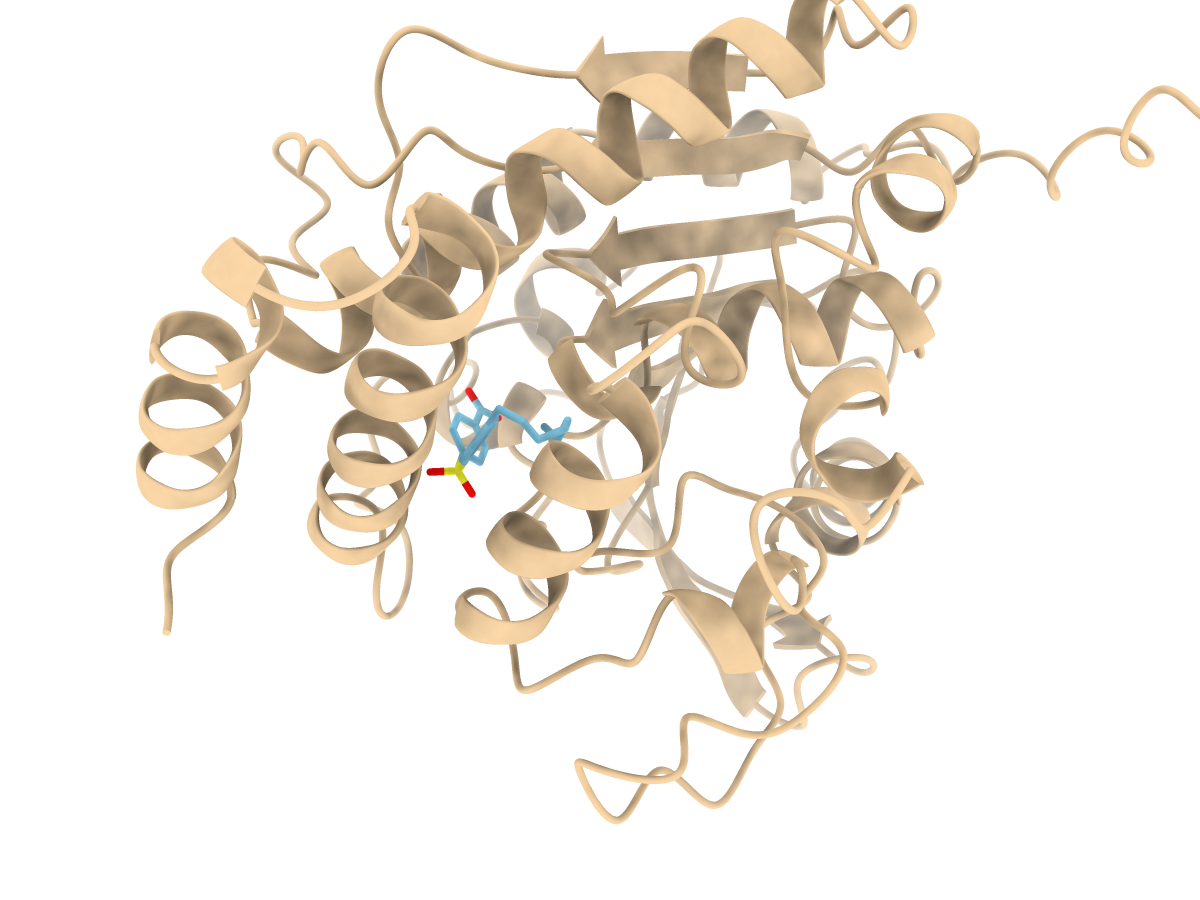}\\
        Reward: 0.88\\
        \vspace{0.1cm}
        \includegraphics[trim=8cm 8cm 12cm 8cm, clip, width=\textwidth]{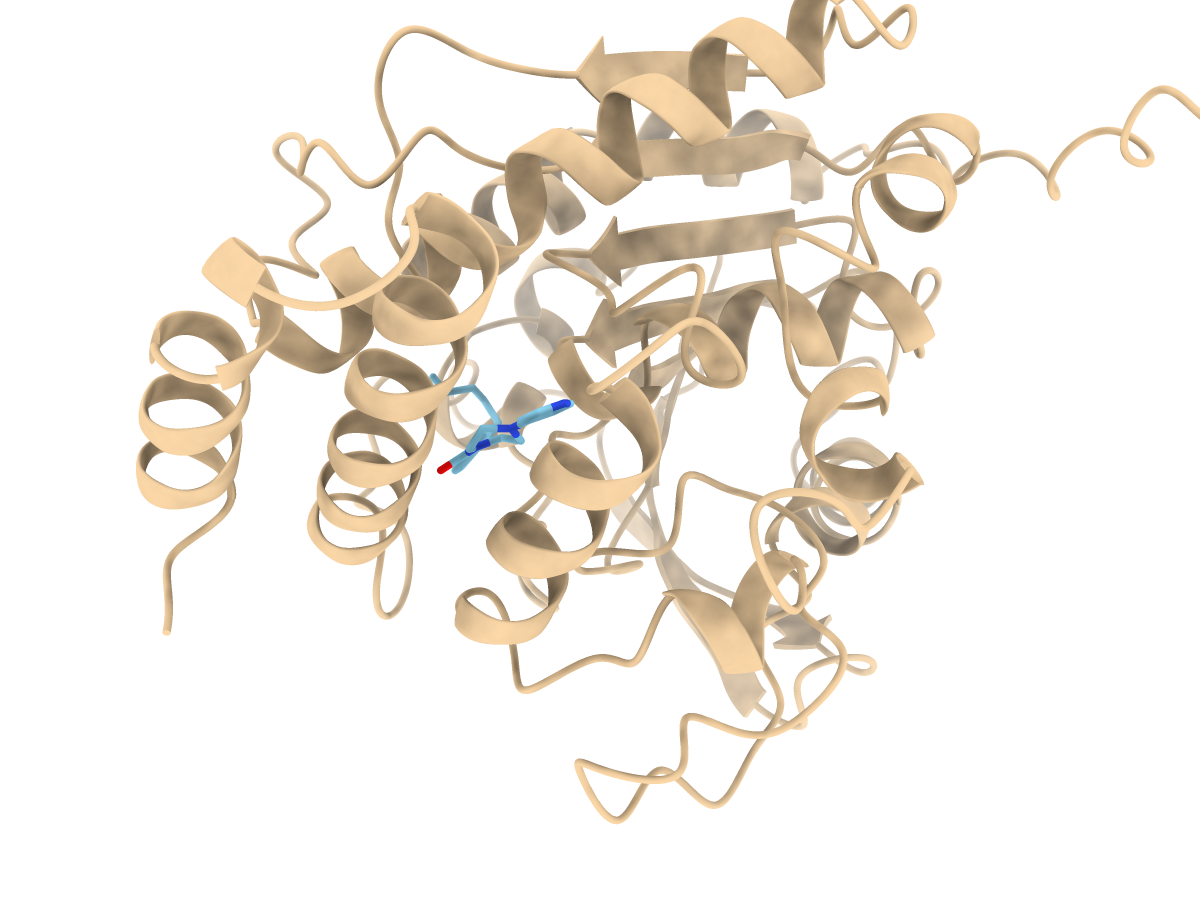}\\
        Reward: 0.69
        \caption{gemma}
    \end{subfigure}
    \caption{Docking poses for sample 803976 generated by different models trying to minimize the docking score to this target.}
    \label{fig:docking_sample_803976}
\end{figure}

\begin{figure}[H]
    \centering
    \begin{subfigure}[b]{0.24\textwidth}
        \centering
        \includegraphics[trim=16cm 6cm 8cm 8cm, clip, width=\textwidth]{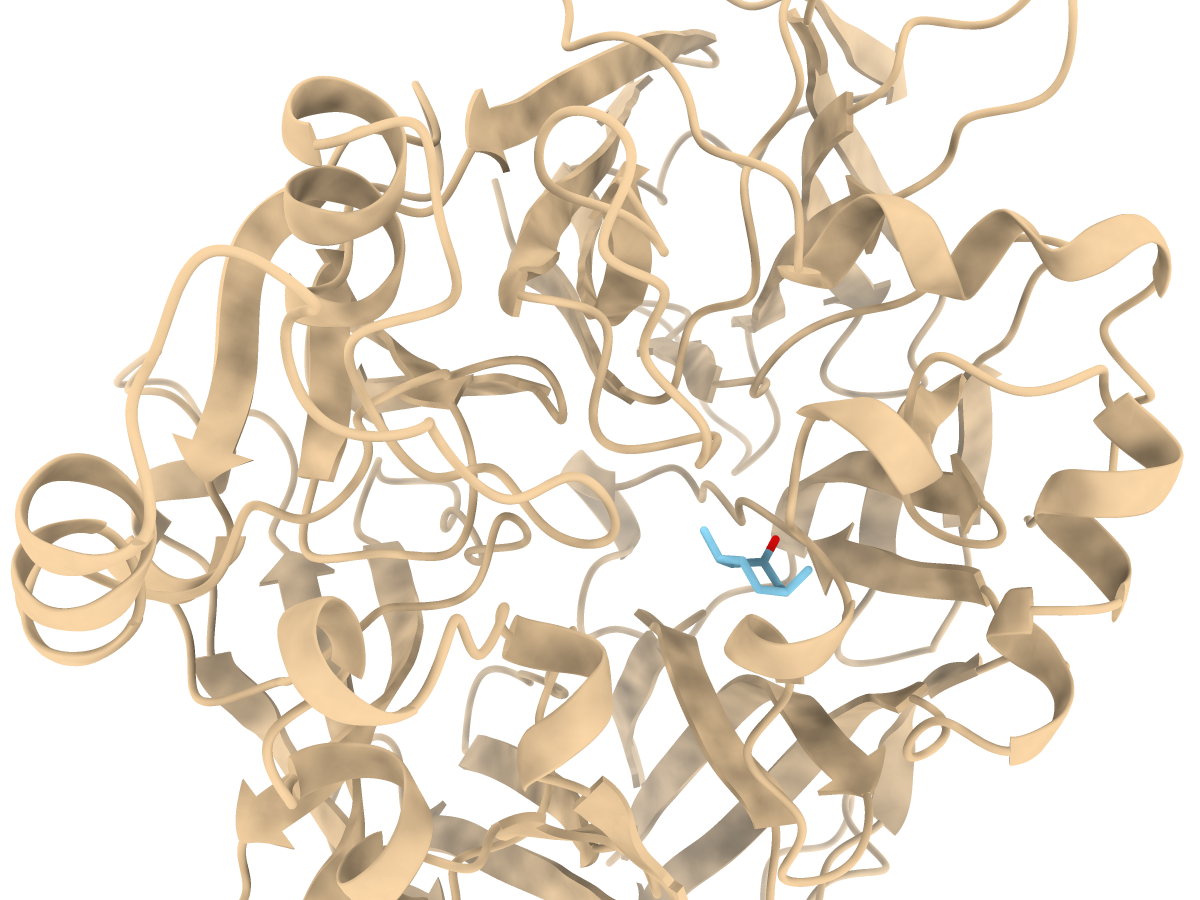}\\
        Reward: 0.0\\
        \vspace{0.1cm}
        \includegraphics[trim=16cm 6cm 8cm 8cm, clip, width=\textwidth]{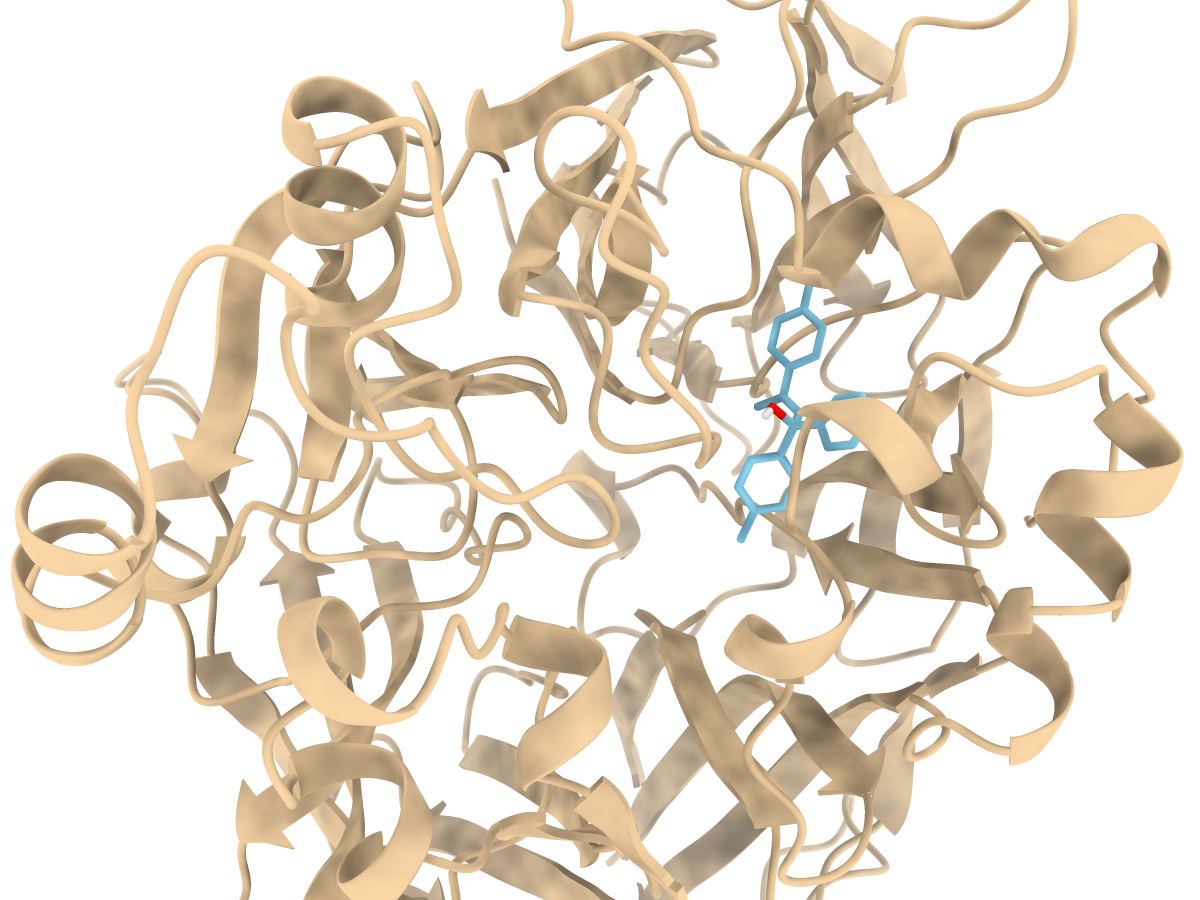}\\
        Reward: 0.0\\
        \vspace{0.1cm}
        \includegraphics[trim=16cm 6cm 8cm 8cm, clip, width=\textwidth]{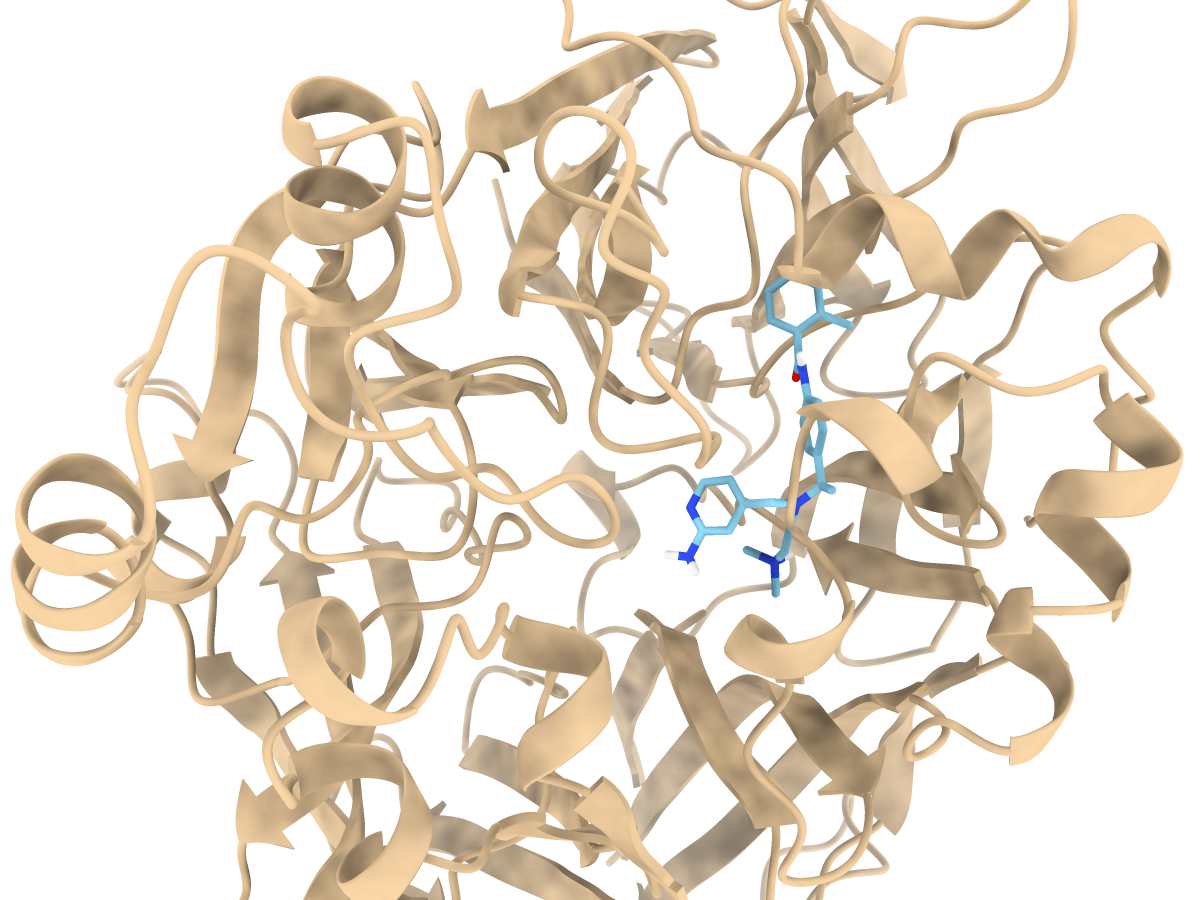}\\
        Reward: 0.0
        \caption{ChemDFM-R}
    \end{subfigure}
    \hfill
    \begin{subfigure}[b]{0.24\textwidth}
        \centering
        \includegraphics[trim=16cm 6cm 8cm 8cm, clip, width=\textwidth]{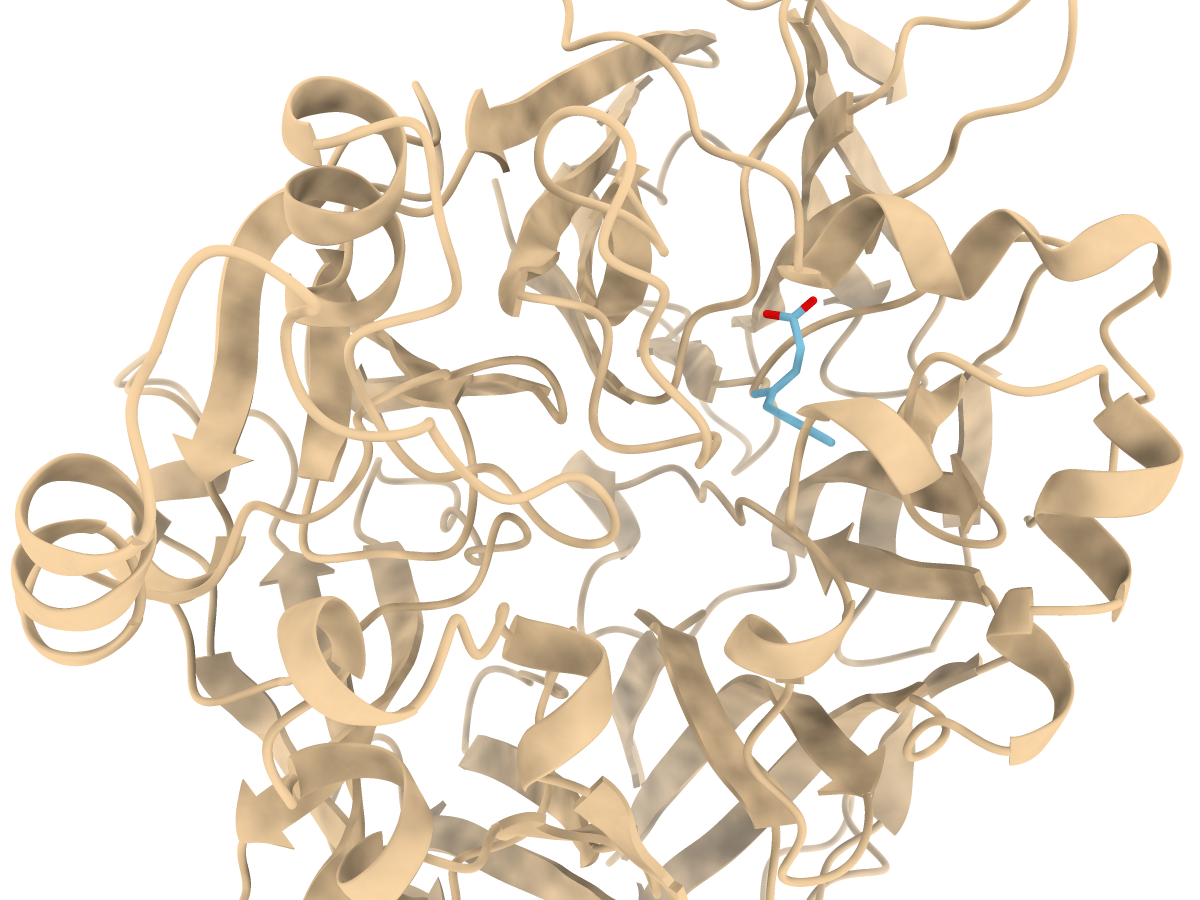}\\
        Reward: 0.0\\
        \vspace{0.1cm}
        \includegraphics[trim=16cm 6cm 8cm 8cm, clip, width=\textwidth]{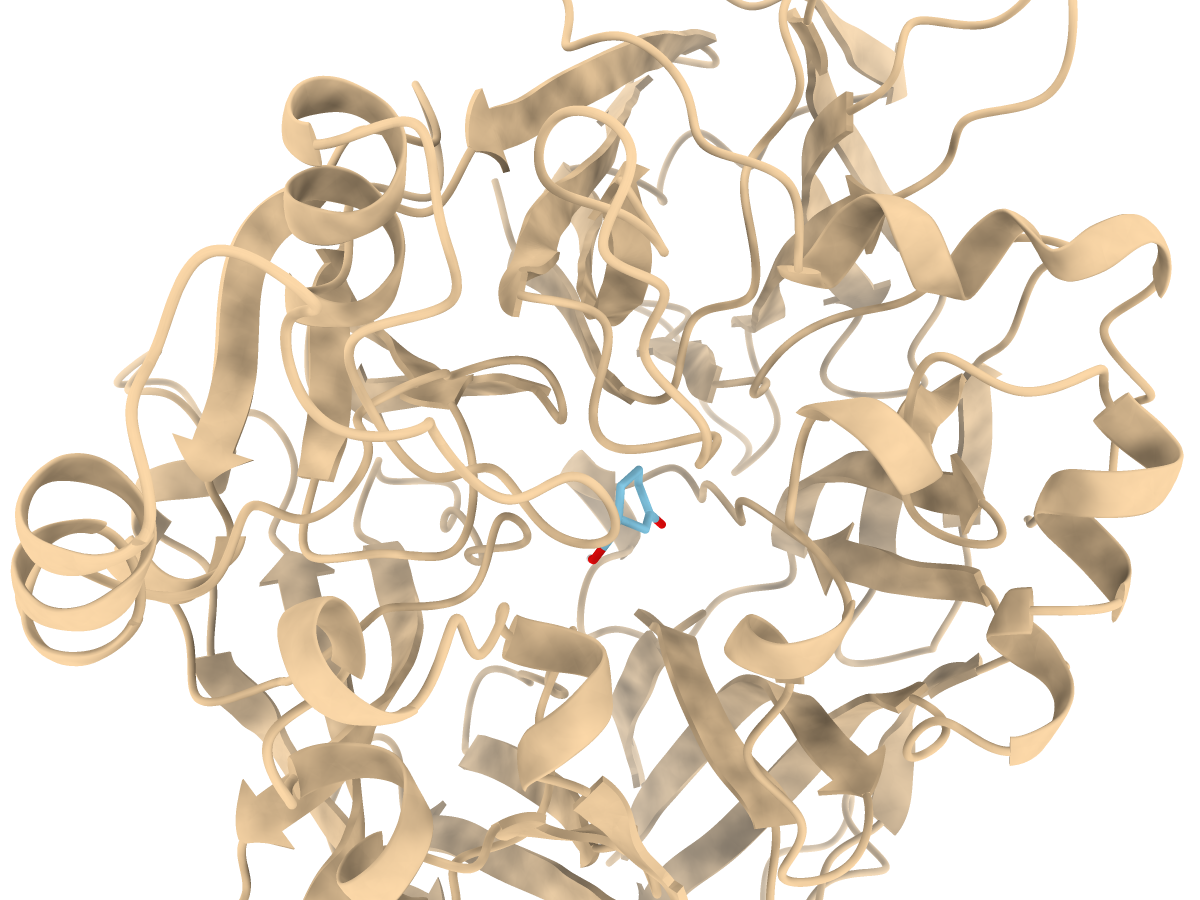}\\
        Reward: 0.0\\
        \vspace{0.1cm}
        \includegraphics[trim=16cm 6cm 8cm 8cm, clip, width=\textwidth]{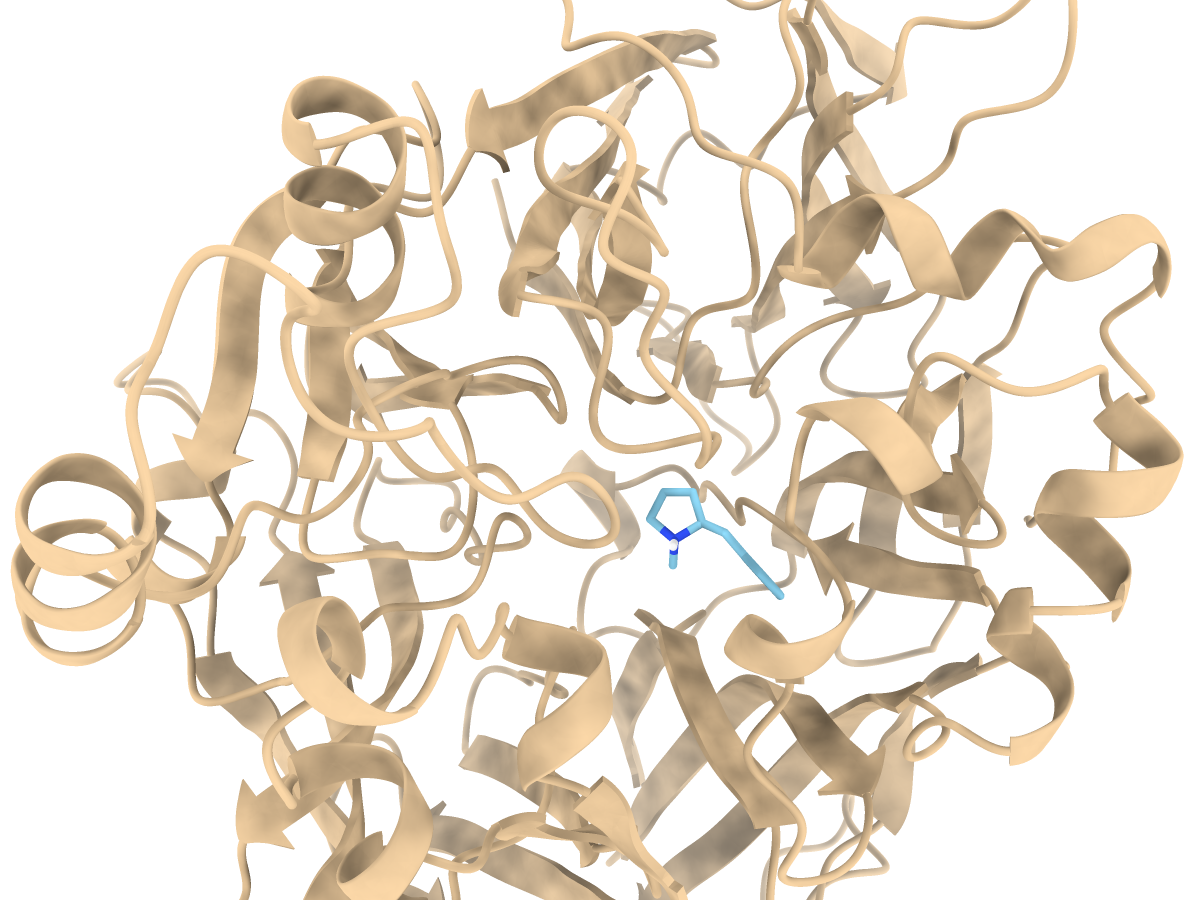}\\
        Reward: 0.0
        \caption{Mistral-Small}
    \end{subfigure}
    \hfill
    \begin{subfigure}[b]{0.24\textwidth}
        \centering
        \includegraphics[trim=16cm 6cm 8cm 8cm, clip, width=\textwidth]{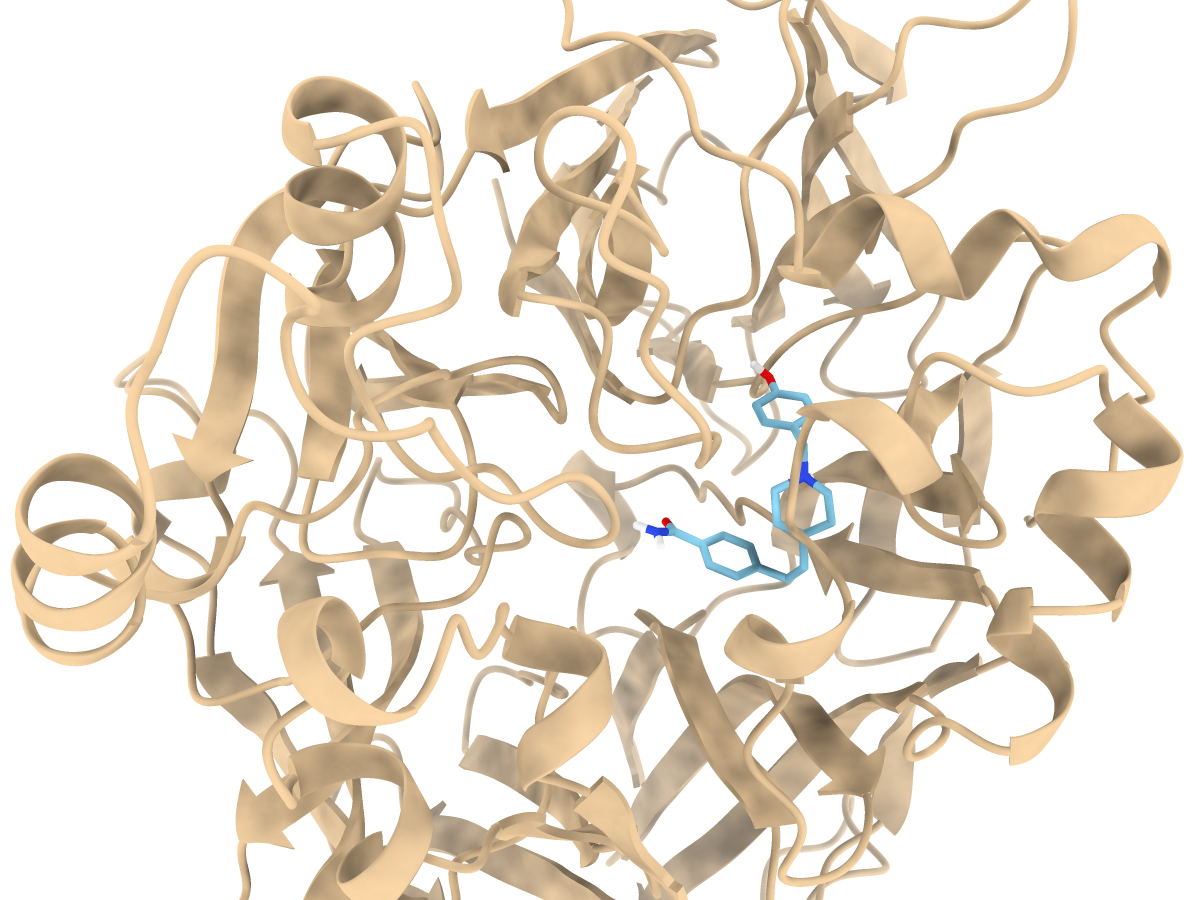}\\
        Reward: 0.98\\
        \vspace{0.1cm}
        \includegraphics[trim=16cm 6cm 8cm 8cm, clip, width=\textwidth]{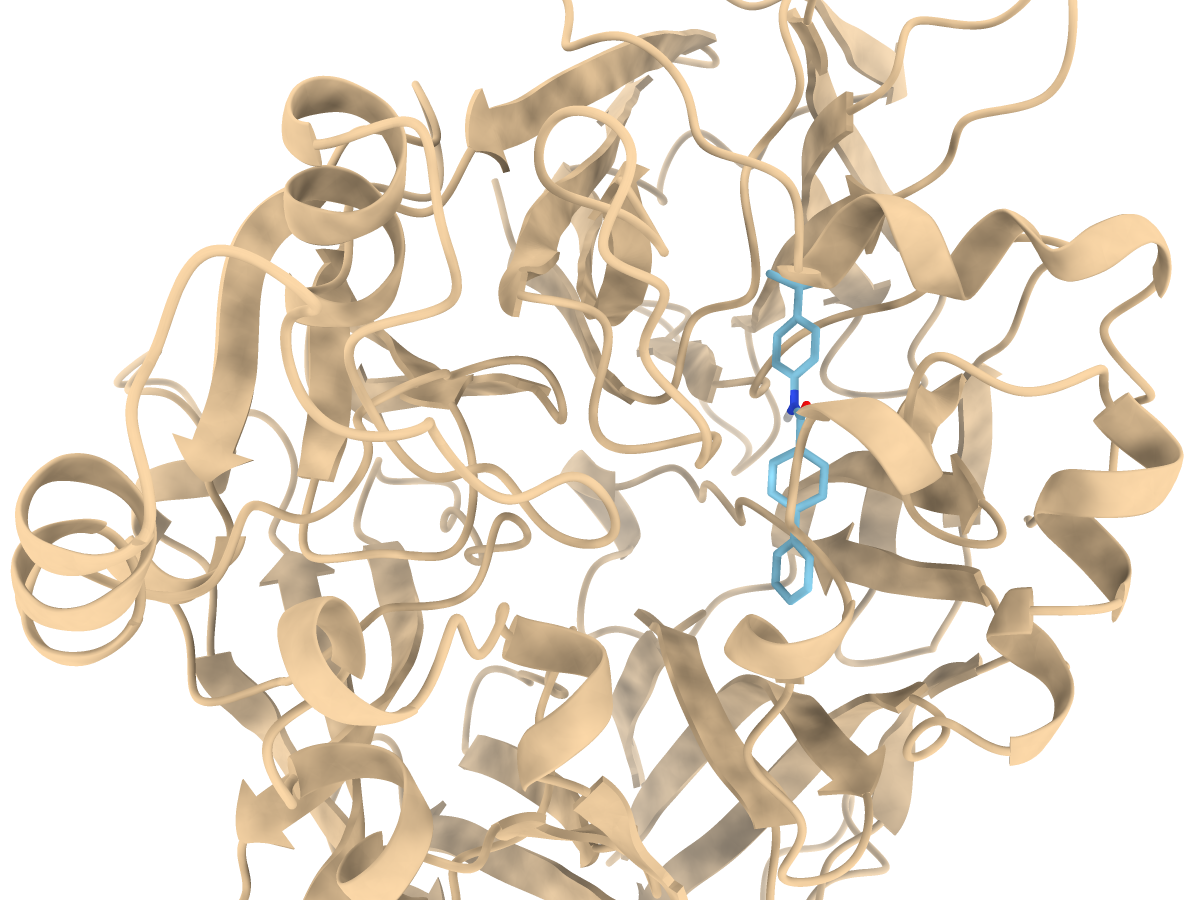}\\
        Reward: 0.97\\
        \vspace{0.1cm}
        \includegraphics[trim=16cm 6cm 8cm 8cm, clip, width=\textwidth]{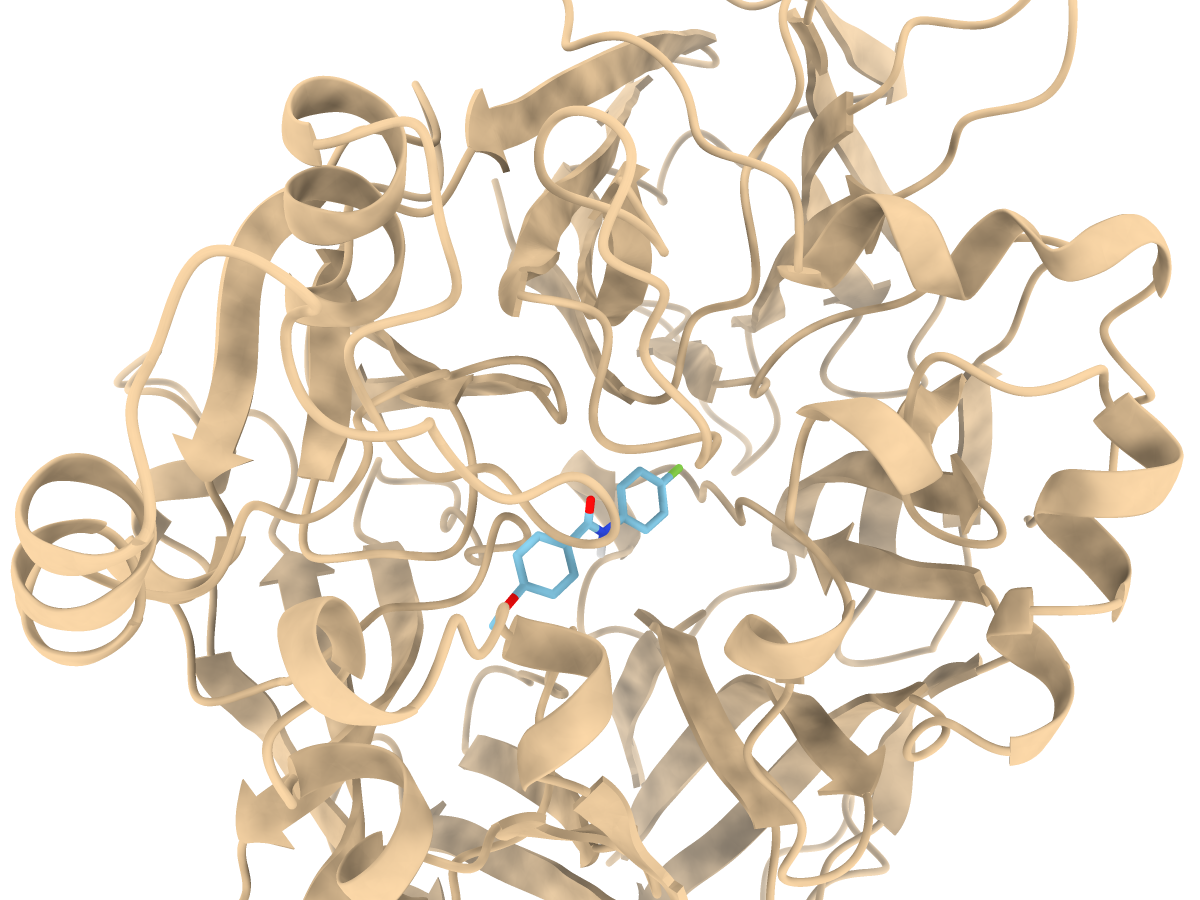}\\
        Reward: 0.78
        \caption{RL-Molstral}
    \end{subfigure}
    \hfill
    \begin{subfigure}[b]{0.24\textwidth}
        \centering
        \includegraphics[trim=16cm 6cm 8cm 8cm, clip, width=\textwidth]{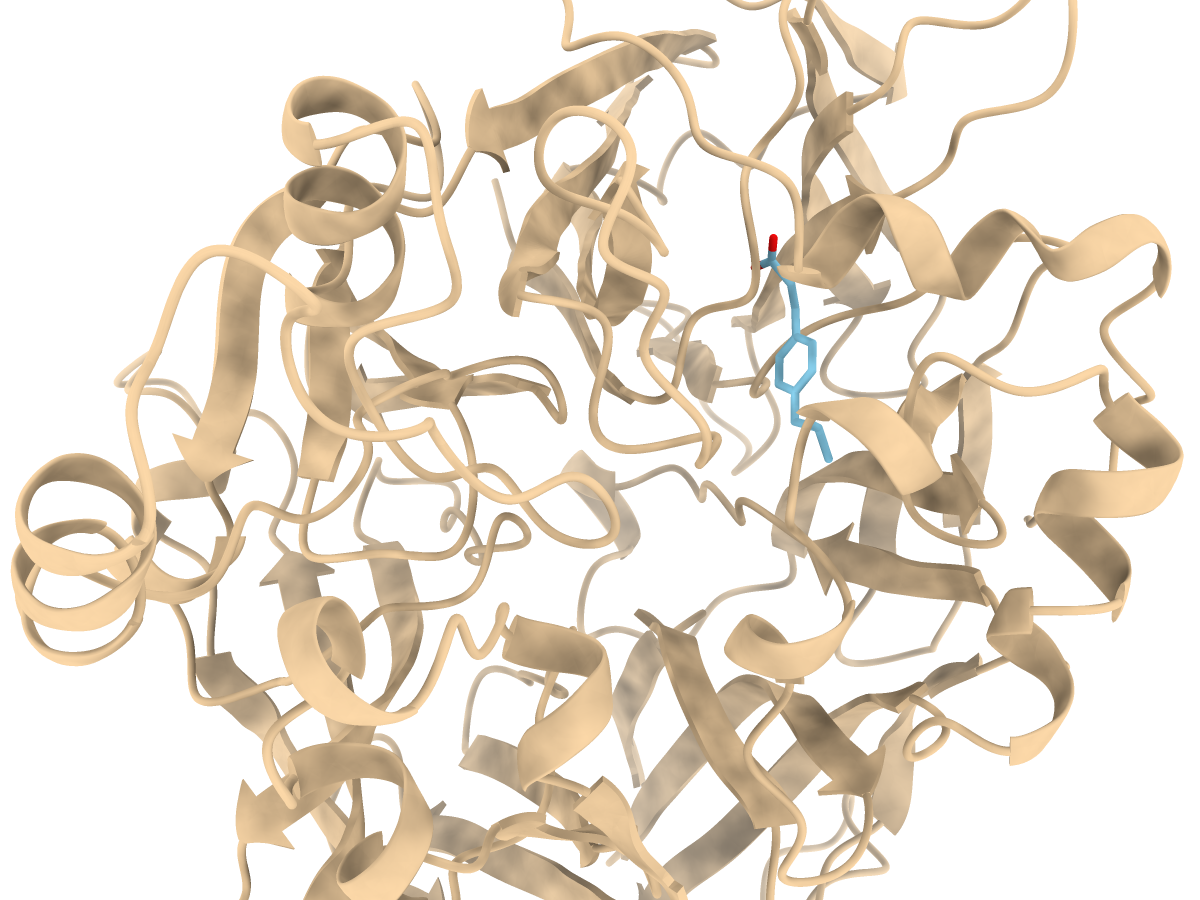}\\
        Reward: 0.77\\
        \vspace{0.1cm}
        \includegraphics[trim=16cm 6cm 8cm 8cm, clip, width=\textwidth]{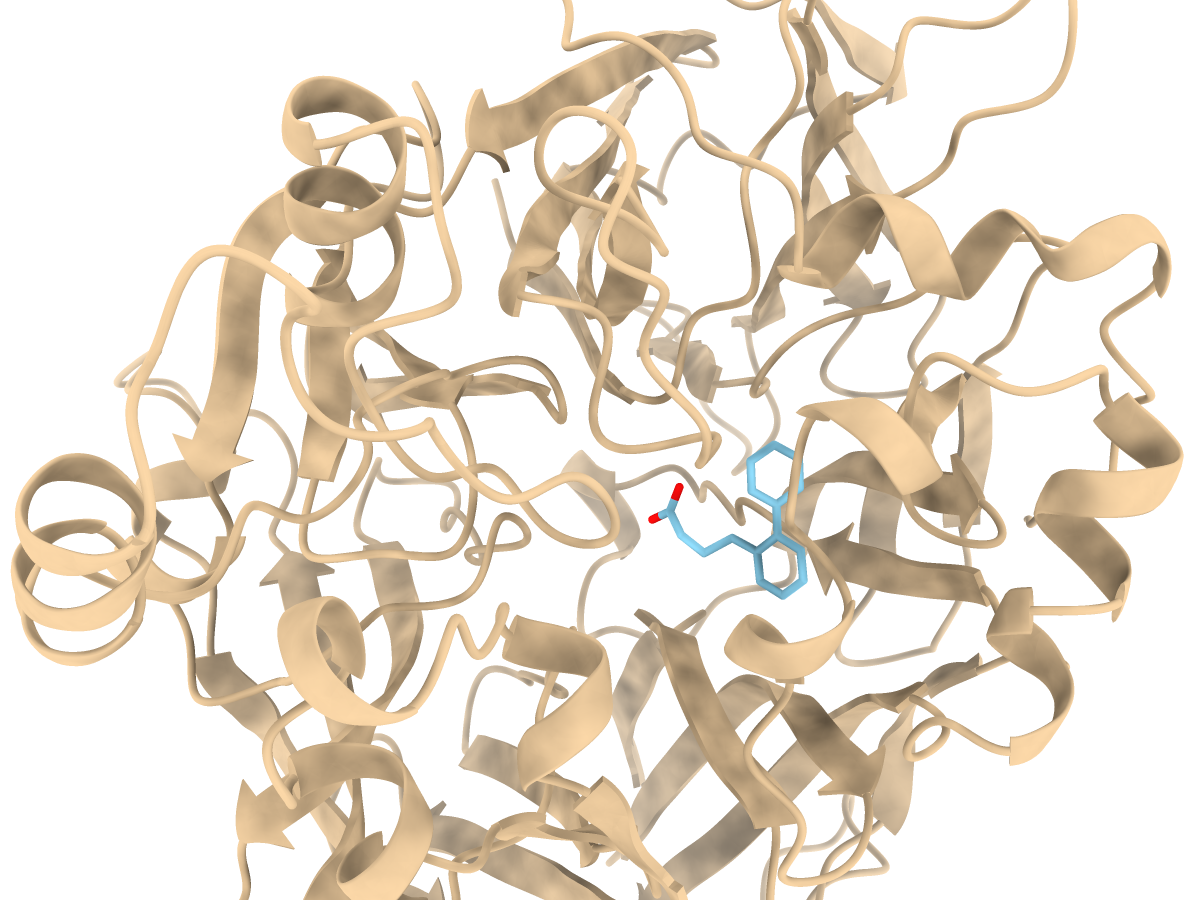}\\
        Reward: 0.83\\
        \vspace{0.1cm}
        \includegraphics[trim=16cm 6cm 8cm 8cm, clip, width=\textwidth]{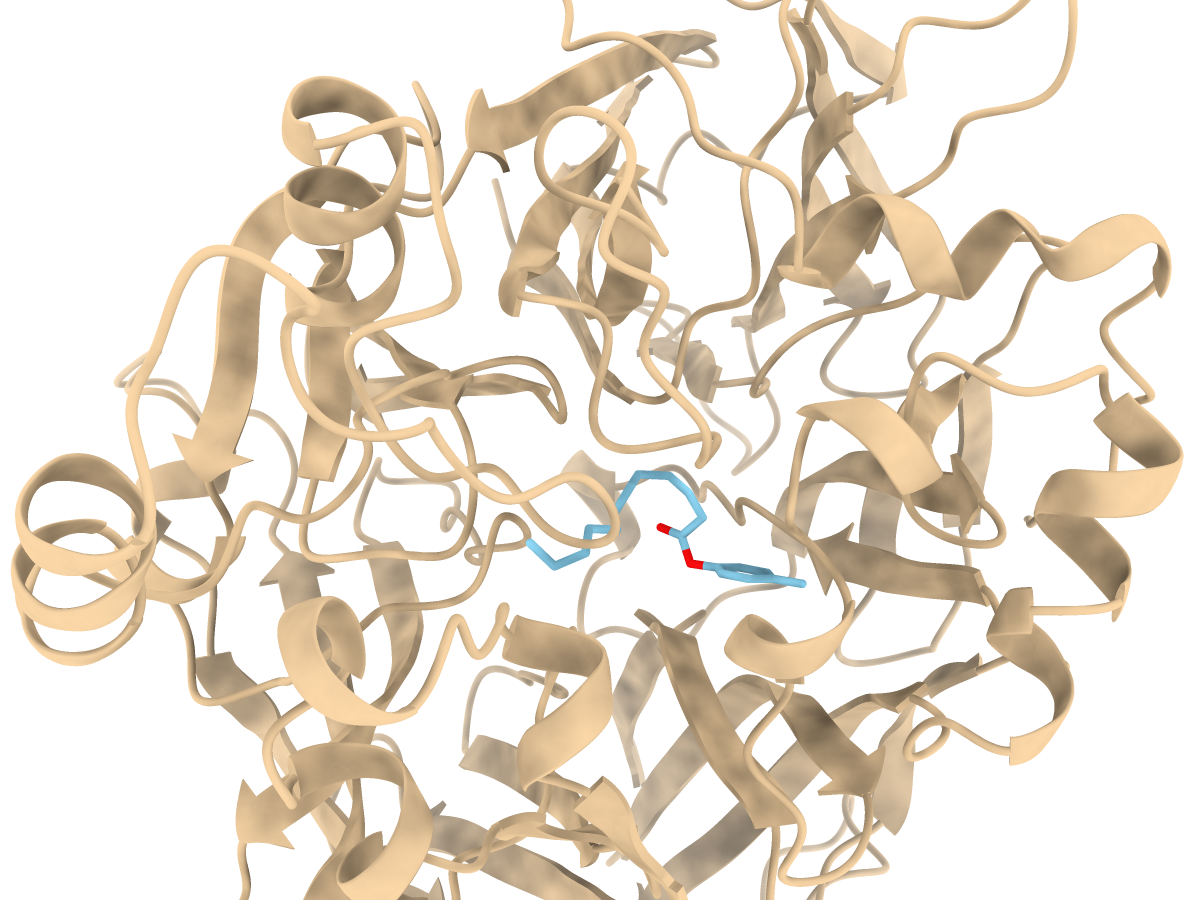}\\
        Reward: 0.84
        \caption{gemma}
    \end{subfigure}
    \caption{Docking poses for sample 941067 generated by different models trying to minimize the docking score to this target along with minimizing the SA score.}
    \label{fig:SA-sample_941067_model_0}
\end{figure}

\FloatBarrier

\FloatBarrier
\subsection{Diversity-Aware Score Selections}
\label{appendix:div_aware_viz}

The figures in this section illustrate the selection of the generated compounds with the diversity aware-top- score.
For all models, as the constraint on the pairwise Tanimoto similarity increases (left to right), some high-scoring molecules are removed from the selection, and replaced by other molecules with lower scores but less similar to the other already selected molecules.
For each plot, the molecules that must be removed from the selection when increasing the similarity constraint are highlighted with a red box, showing which molecules are not considered in the final selection when the similarity constraint is increased.

\begin{figure}[ht]
    \centering
        \begin{subfigure}[b]{0.3\textwidth}
        \includegraphics[width=\textwidth]{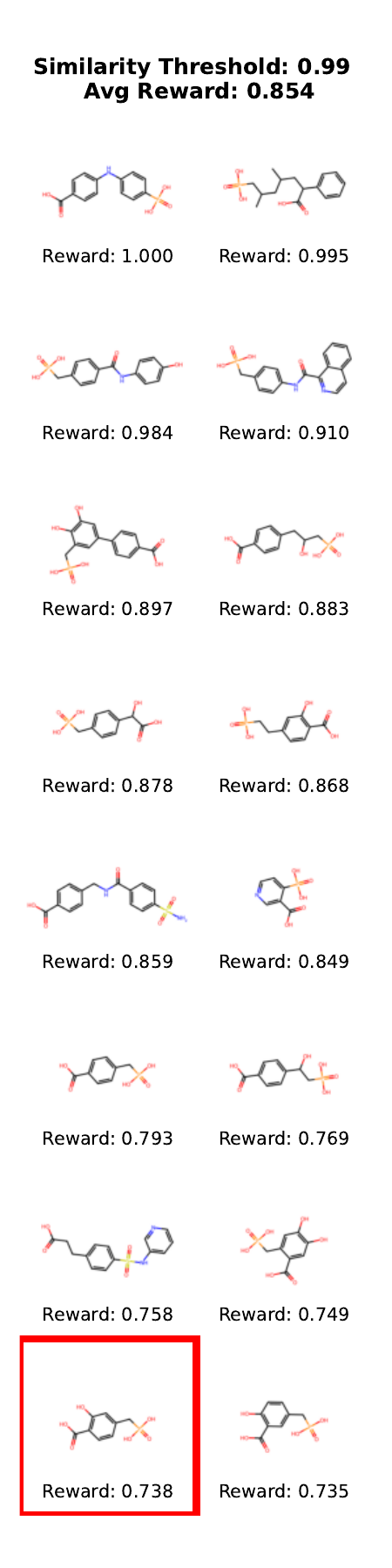}
    \end{subfigure}
    \begin{subfigure}[b]{0.3\textwidth}
        \includegraphics[width=\textwidth]{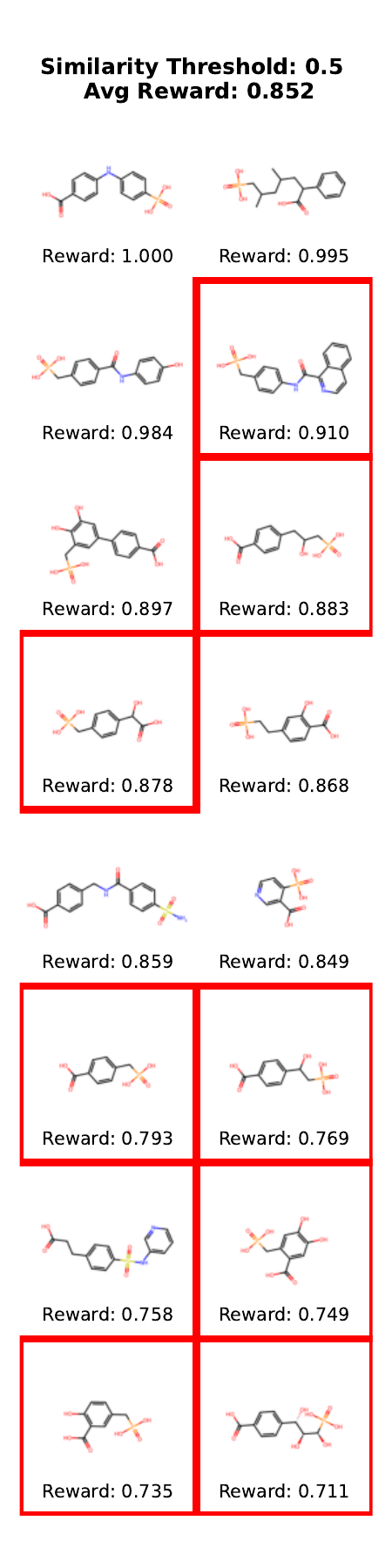}
    \end{subfigure}
    \begin{subfigure}[b]{0.3\textwidth}
        \includegraphics[width=\textwidth]{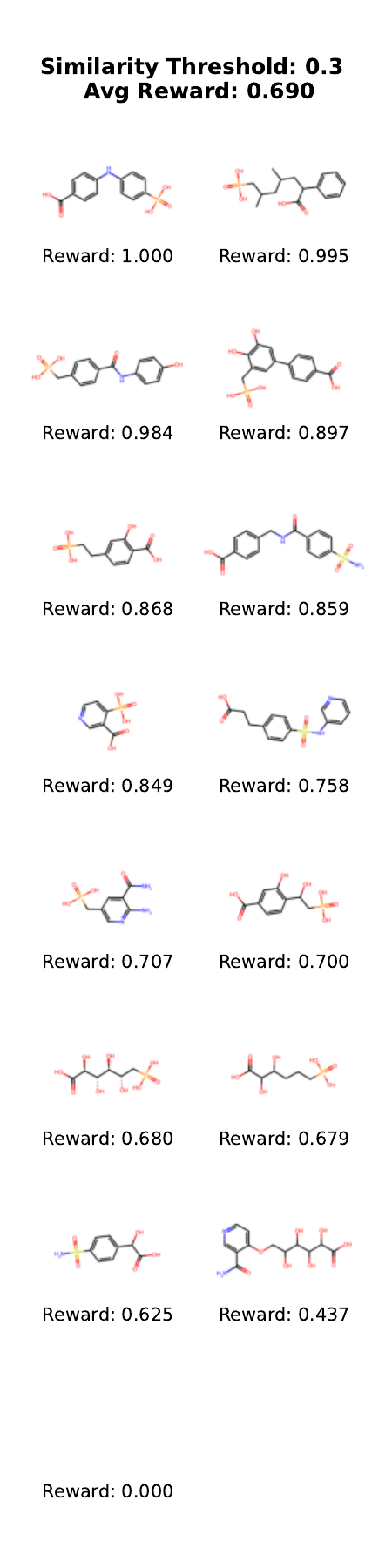}
    \end{subfigure}

    \caption{Molecules selected for the diversity aware score for k=16 with 64 rollouts when changing the similarity threshold value (generated with gemma-4 for the prompt id 34710). The similarity constraints gets stronger from left to right, and molecules in a red box are removed from the selected pool are the ones that are removed from it when the similarity constraints increases.
    }
\label{fig:gemma-4-34710}
\end{figure}

\begin{figure}[ht]
    \centering
        \begin{subfigure}[b]{0.3\textwidth}
        \includegraphics[width=\textwidth]{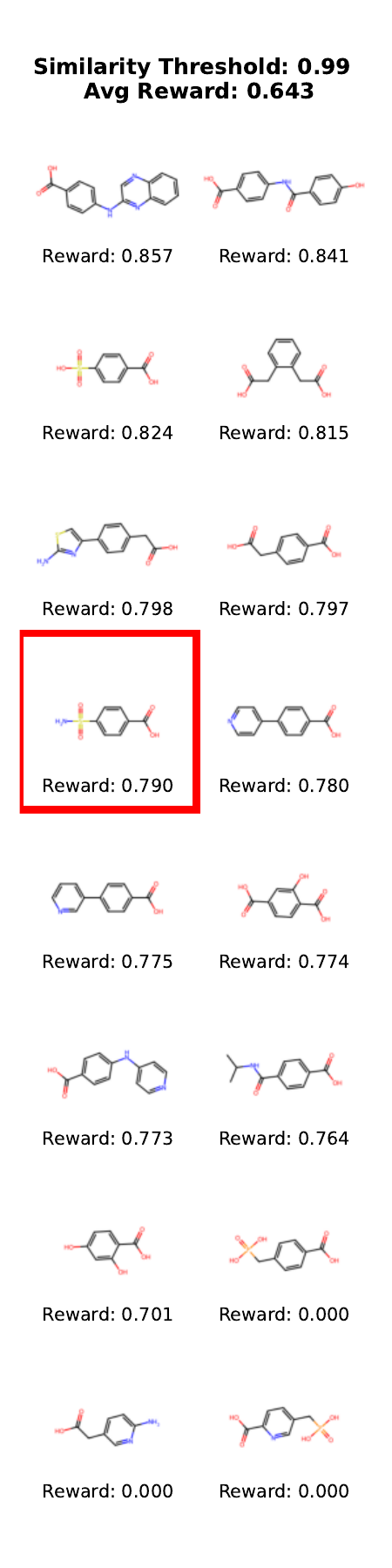}
    \end{subfigure}
    \begin{subfigure}[b]{0.3\textwidth}
        \includegraphics[width=\textwidth]{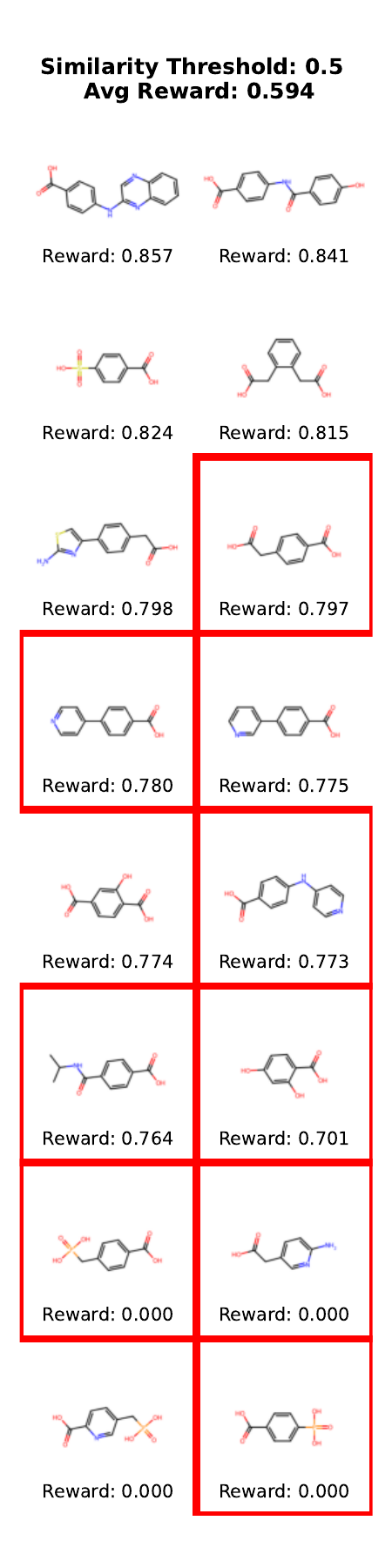}
    \end{subfigure}
    \begin{subfigure}[b]{0.3\textwidth}
        \includegraphics[width=\textwidth]{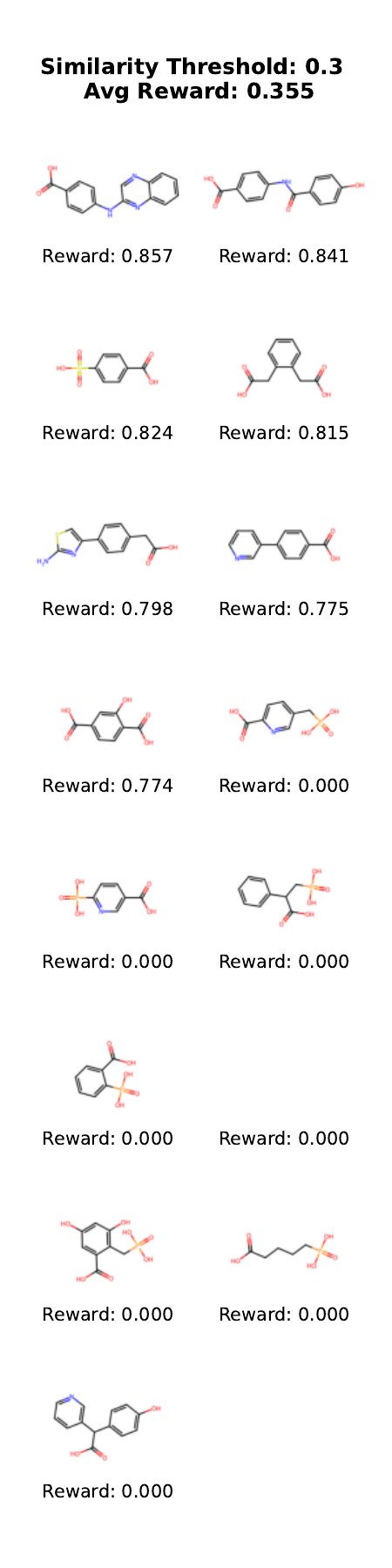}
    \end{subfigure}

    \caption{Molecules selected for the diversity aware score for k=16 with 64 rollouts when changing the similarity threshold value (generated with gemma-4 for the prompt id 347103906319). The similarity constraints gets stronger from left to right, and molecules in a red box are removed from the selected pool are the ones that are removed from it when the similarity constraints increases.
    }
\label{fig:gemma-4-347103906319}
\end{figure}

\begin{figure}[ht]
    \centering
        \begin{subfigure}[b]{0.3\textwidth}
        \includegraphics[width=\textwidth]{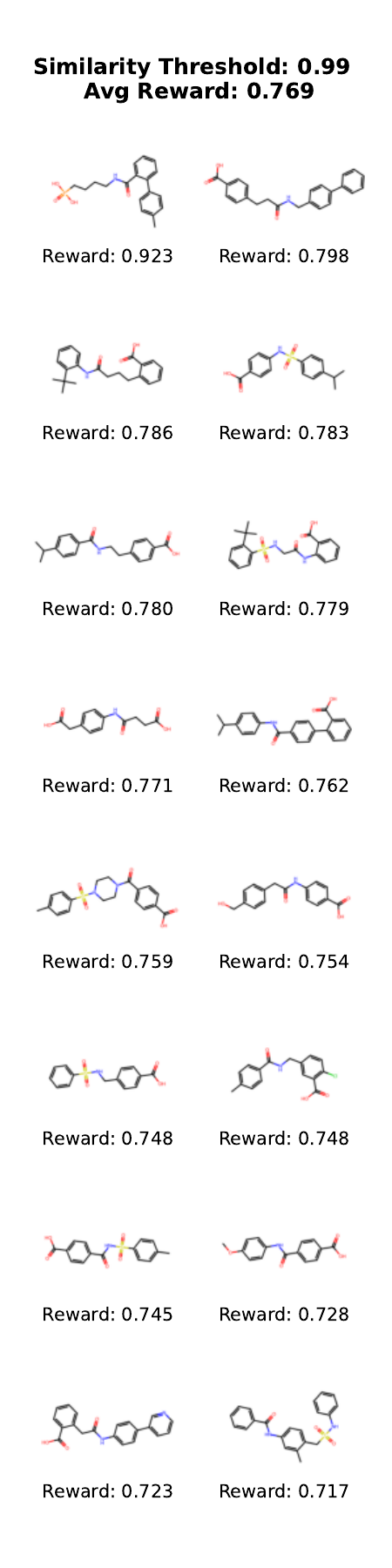}
    \end{subfigure}
    \begin{subfigure}[b]{0.3\textwidth}
        \includegraphics[width=\textwidth]{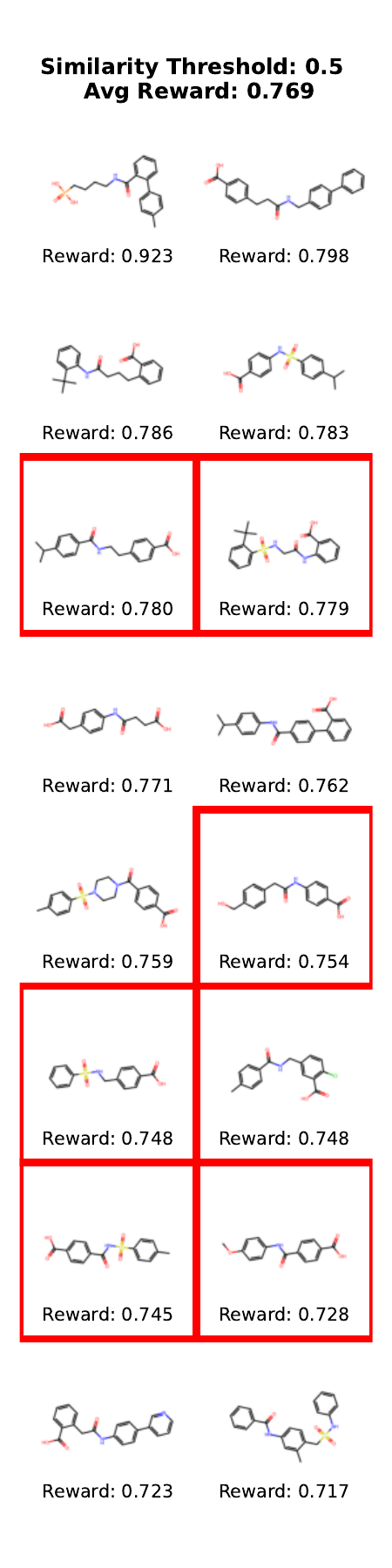}
    \end{subfigure}
    \begin{subfigure}[b]{0.3\textwidth}
        \includegraphics[width=\textwidth]{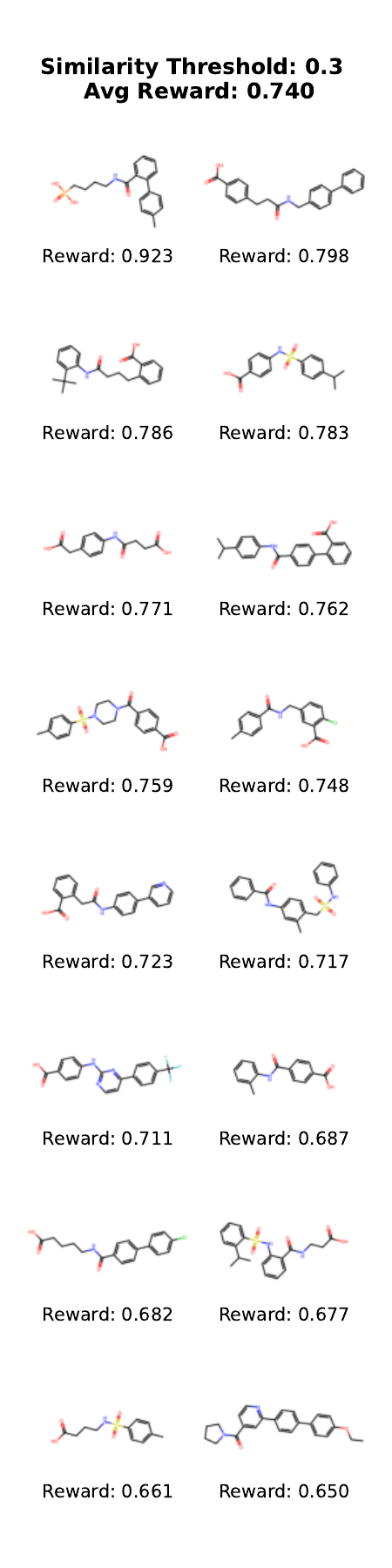}
    \end{subfigure}

    \caption{Molecules selected for the diversity aware score for k=16 with 64 rollouts when changing the similarity threshold value (generated with RL-Molstral-g16 for the prompt id 34710). The similarity constraints gets stronger from left to right, and molecules in a red box are removed from the selected pool are the ones that are removed from it when the similarity constraints increases.
    }
\label{fig:RL-Molstral-g16-34710}
\end{figure}

\begin{figure}[ht]
    \centering
        \begin{subfigure}[b]{0.3\textwidth}
        \includegraphics[width=\textwidth]{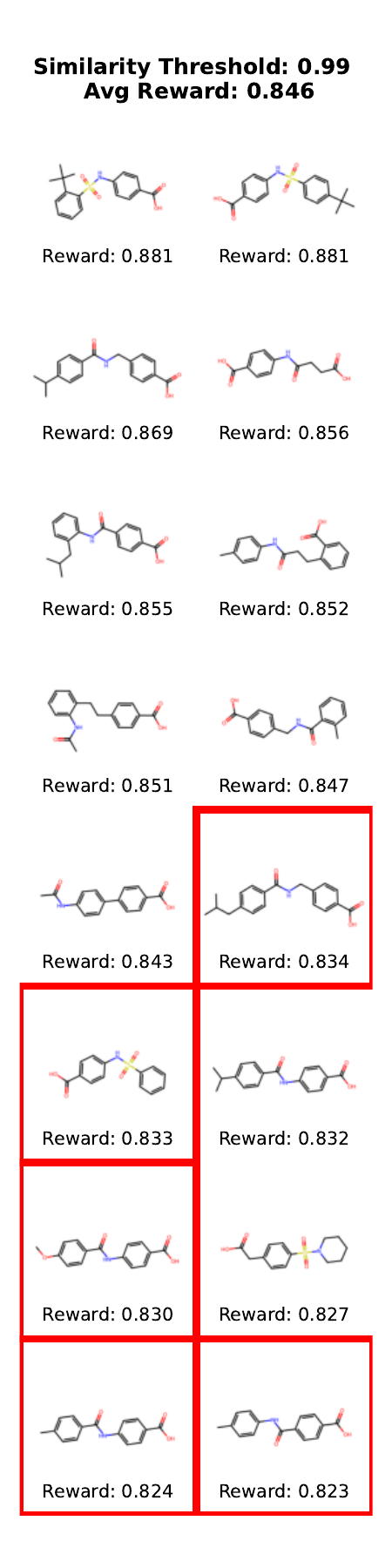}
    \end{subfigure}
    \begin{subfigure}[b]{0.3\textwidth}
        \includegraphics[width=\textwidth]{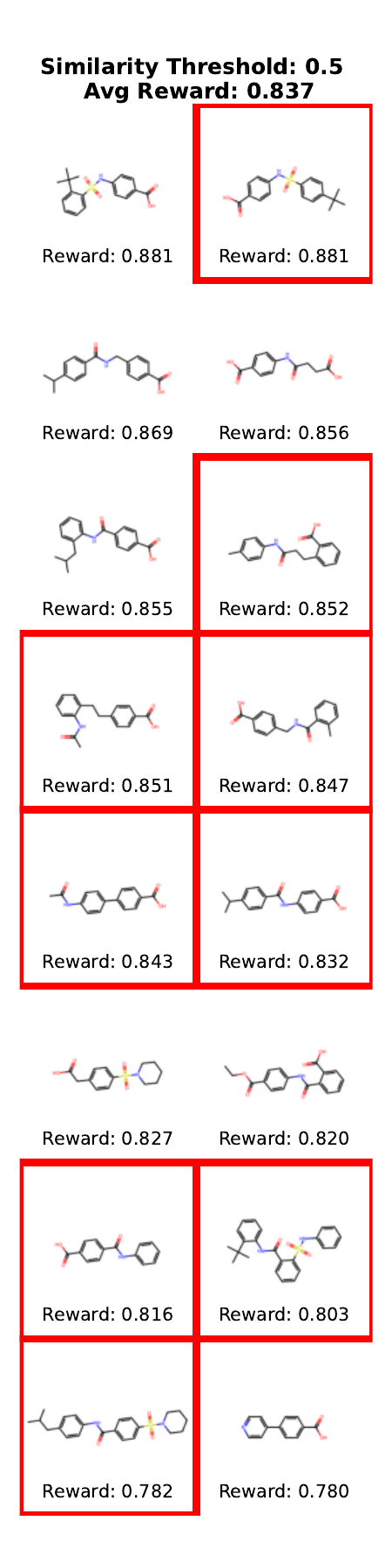}
    \end{subfigure}
    \begin{subfigure}[b]{0.3\textwidth}
        \includegraphics[width=\textwidth]{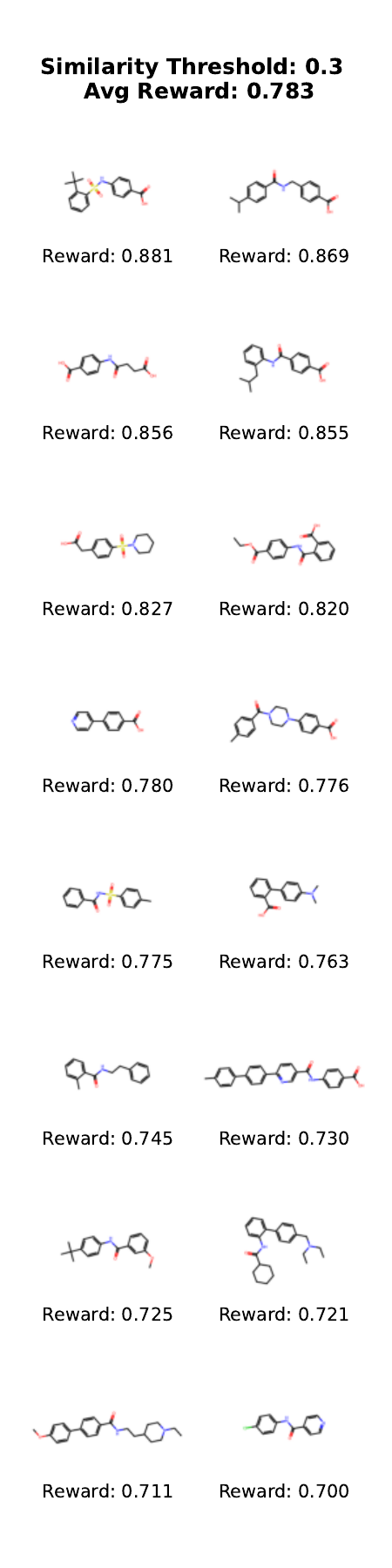}
    \end{subfigure}

    \caption{Molecules selected for the diversity aware score for k=16 with 64 rollouts when changing the similarity threshold value (generated with RL-Molstral-g16 for the prompt id 347103906319). The similarity constraints gets stronger from left to right, and molecules in a red box are removed from the selected pool are the ones that are removed from it when the similarity constraints increases.
    }
\label{fig:RL-Molstral-g16-347103906319}
\end{figure}

In~\autoref{fig:gemma-4-34710}, and~\autoref{fig:gemma-4-347103906319}, we can see that for Gemma-4, the selected molecules chosen when changing the similarity constraint largely varies, until eventually, not enough molecules meet the diversity criteria to fill the top-16 selection.
Overall the same trend can be seen in~\autoref{fig:RL-Molstral-g16-34710}, and~\autoref{fig:RL-Molstral-g16-347103906319} for RL-Molstral, although the selected molecules seem to be less similar to each other, resulting in 16 molecules being selected with the strongest similarity constraint.

\begin{figure}[ht]
    \centering
        \begin{subfigure}[b]{0.3\textwidth}
        \includegraphics[width=\textwidth]{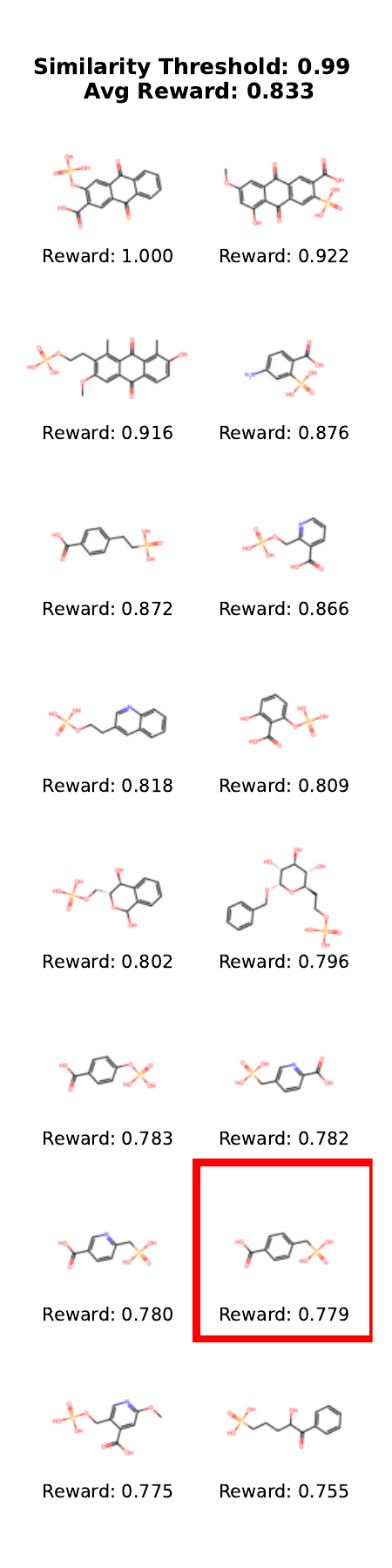}
    \end{subfigure}
    \begin{subfigure}[b]{0.3\textwidth}
        \includegraphics[width=\textwidth]{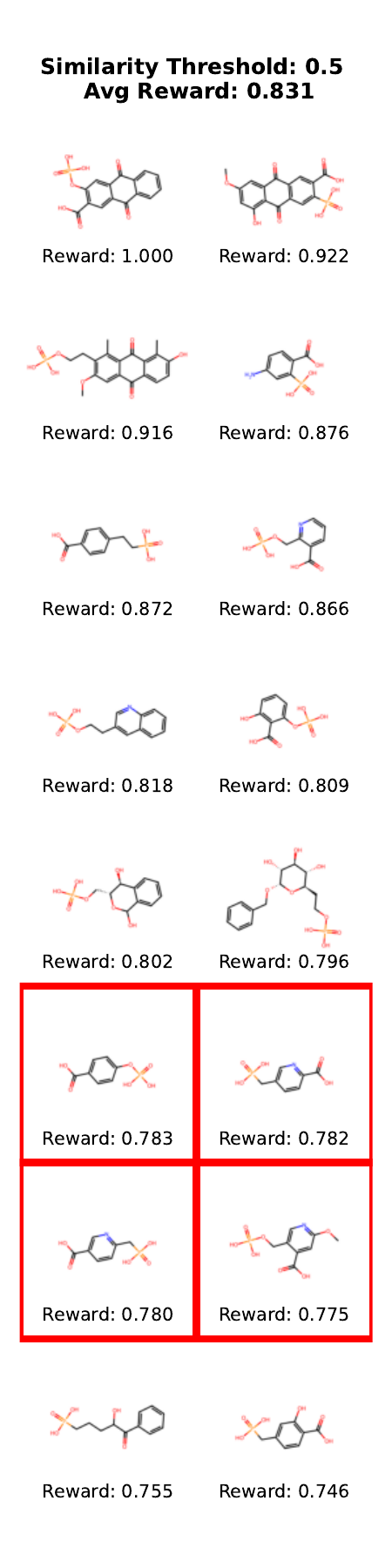}
    \end{subfigure}
    \begin{subfigure}[b]{0.3\textwidth}
        \includegraphics[width=\textwidth]{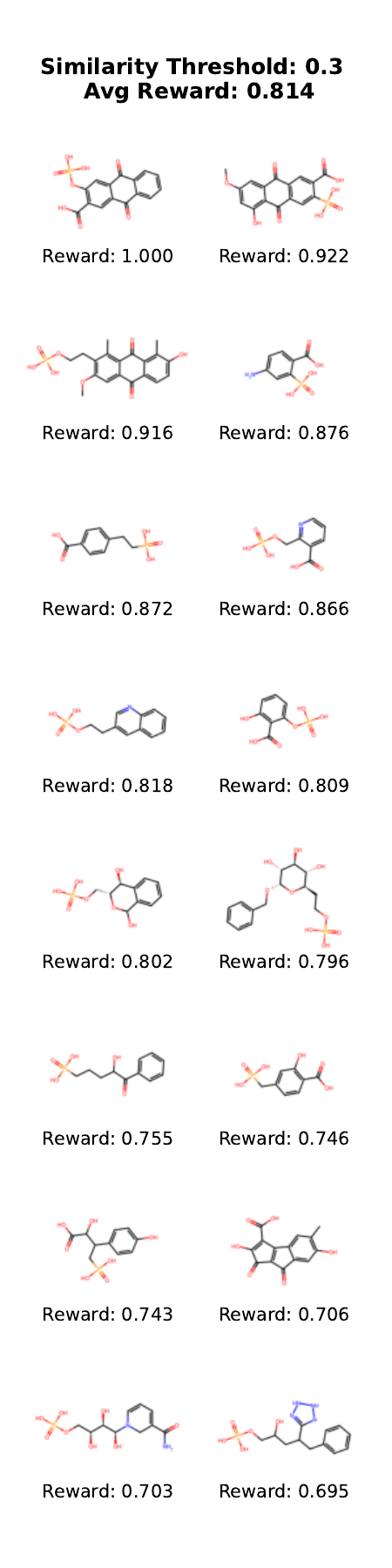}
    \end{subfigure}

    \caption{Molecules selected for the diversity aware score for k=16 with 64 rollouts when changing the similarity threshold value (generated with Mistral-Small-4 for the prompt id 34710). The similarity constraints gets stronger from left to right, and molecules in a red box are removed from the selected pool are the ones that are removed from it when the similarity constraints increases.
    }
\label{fig:Mistral-Small-4-34710}
\end{figure}

\begin{figure}[ht]
    \centering
        \begin{subfigure}[b]{0.3\textwidth}
        \includegraphics[width=\textwidth]{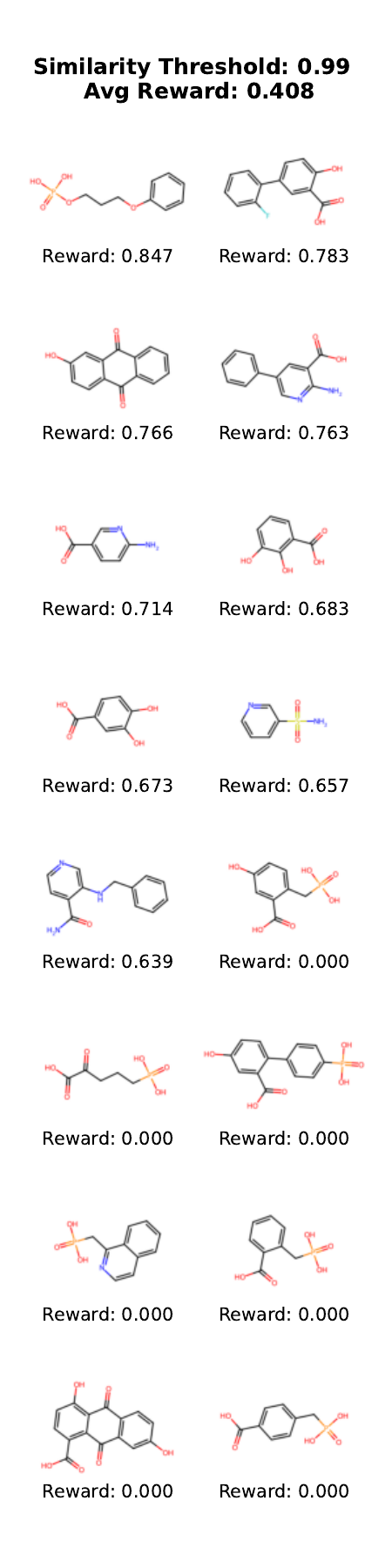}
    \end{subfigure}
    \begin{subfigure}[b]{0.3\textwidth}
        \includegraphics[width=\textwidth]{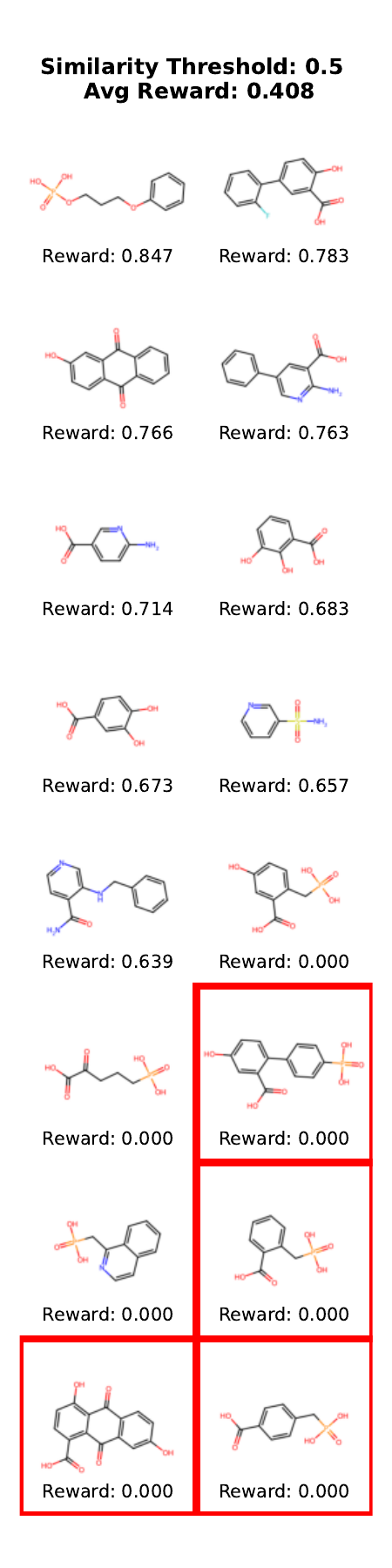}
    \end{subfigure}
    \begin{subfigure}[b]{0.3\textwidth}
        \includegraphics[width=\textwidth]{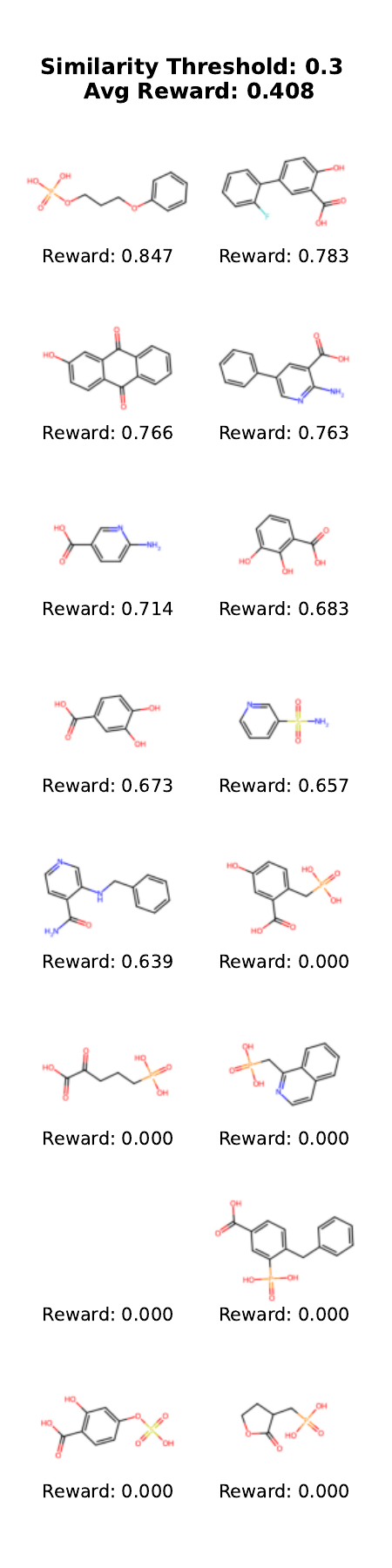}
    \end{subfigure}

    \caption{Molecules selected for the diversity aware score for k=16 with 64 rollouts when changing the similarity threshold value (generated with Mistral-Small-4 for the prompt id 347103906319). The similarity constraints gets stronger from left to right, and molecules in a red box are removed from the selected pool are the ones that are removed from it when the similarity constraints increases.
    }
\label{fig:Mistral-Small-4-347103906319}
\end{figure}

\begin{figure}[ht]
    \centering
        \begin{subfigure}[b]{0.3\textwidth}
        \includegraphics[width=\textwidth]{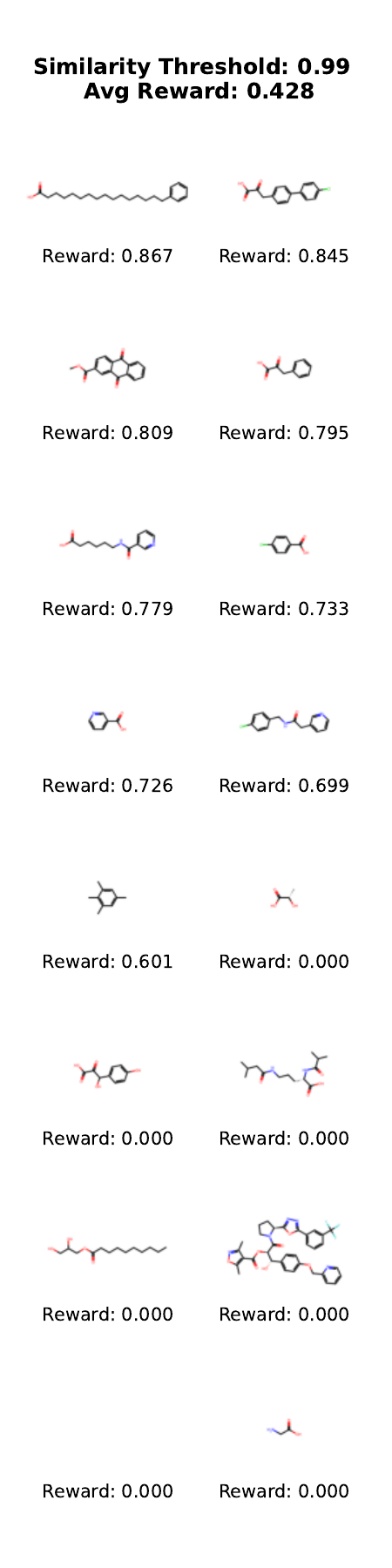}
    \end{subfigure}
    \begin{subfigure}[b]{0.3\textwidth}
        \includegraphics[width=\textwidth]{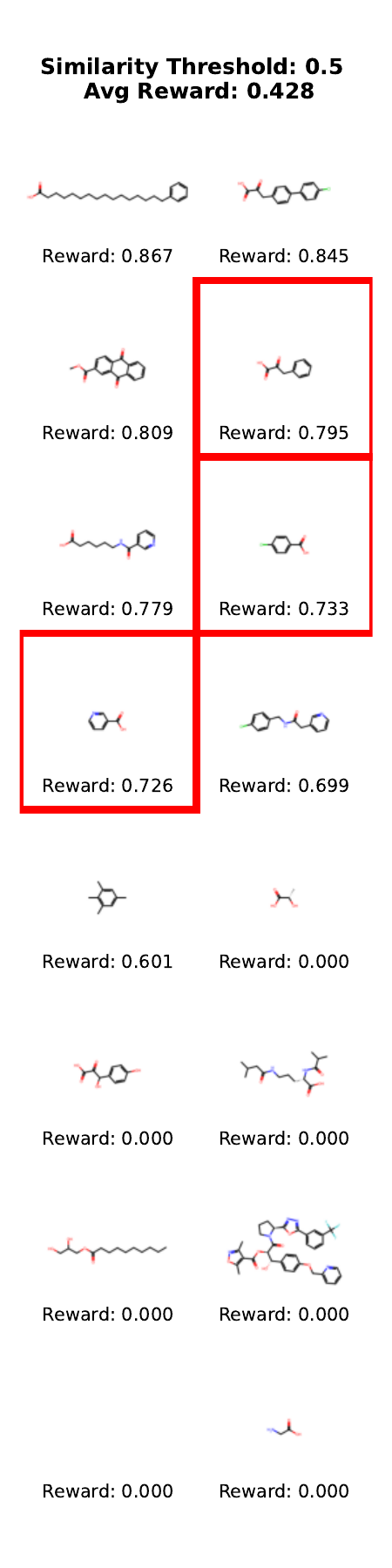}
    \end{subfigure}
    \begin{subfigure}[b]{0.3\textwidth}
        \includegraphics[width=\textwidth]{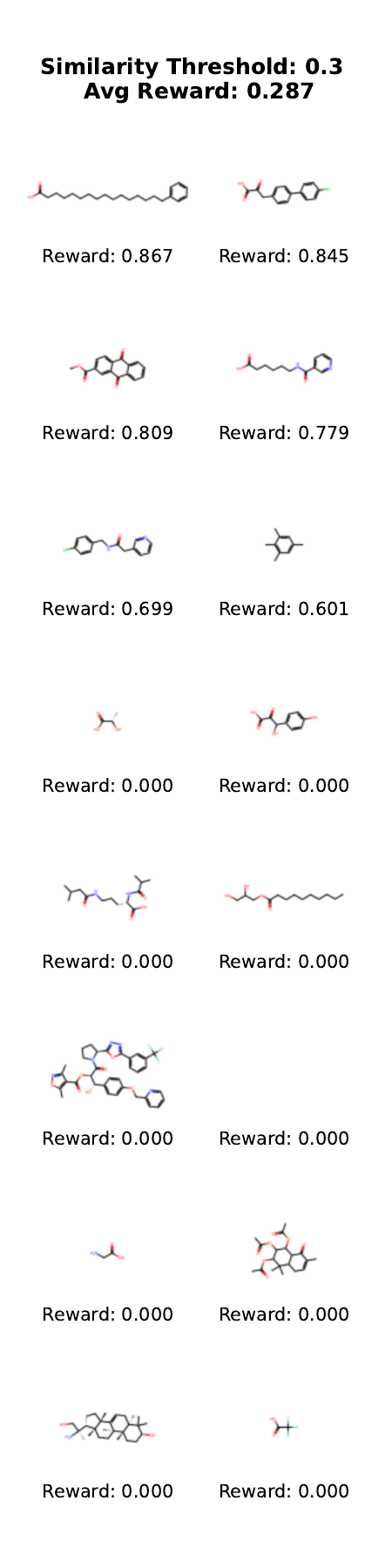}
    \end{subfigure}

    \caption{Molecules selected for the diversity aware score for k=16 with 64 rollouts when changing the similarity threshold value (generated with ChemDFM-R for the prompt id 347103906319). The similarity constraints gets stronger from left to right, and molecules in a red box are removed from the selected pool are the ones that are removed from it when the similarity constraints increases.
    }
\label{fig:ChemDFM-R-347103906319}
\end{figure}

\begin{figure}[ht]
    \centering
        \begin{subfigure}[b]{0.3\textwidth}
        \includegraphics[width=\textwidth]{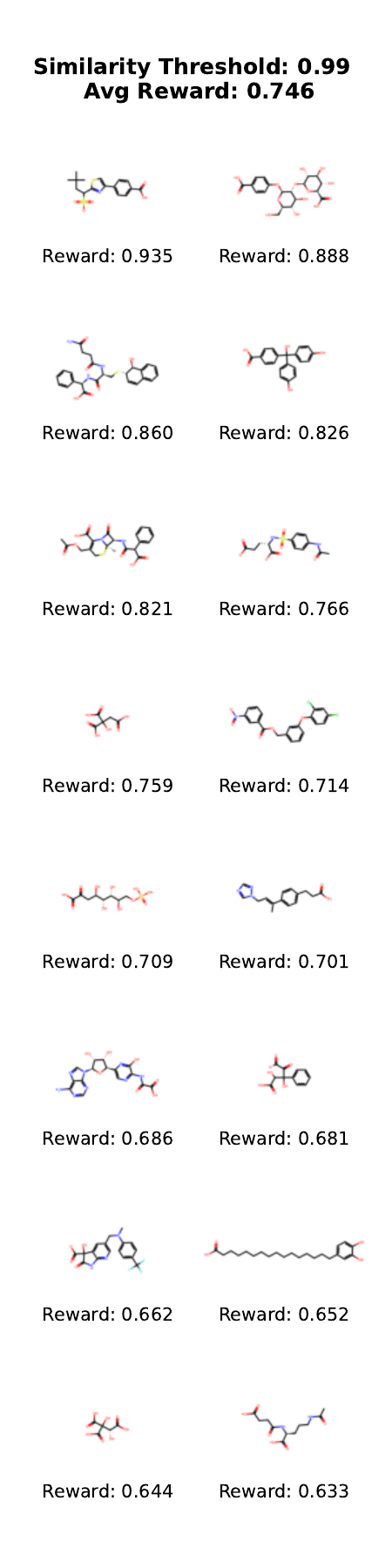}
    \end{subfigure}
    \begin{subfigure}[b]{0.3\textwidth}
        \includegraphics[width=\textwidth]{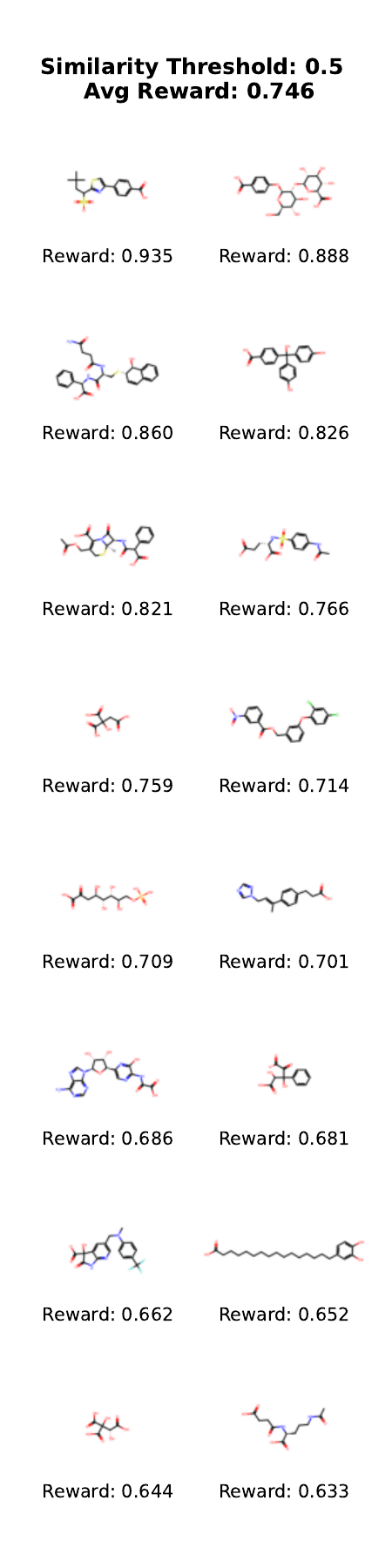}
    \end{subfigure}
    \begin{subfigure}[b]{0.3\textwidth}
        \includegraphics[width=\textwidth]{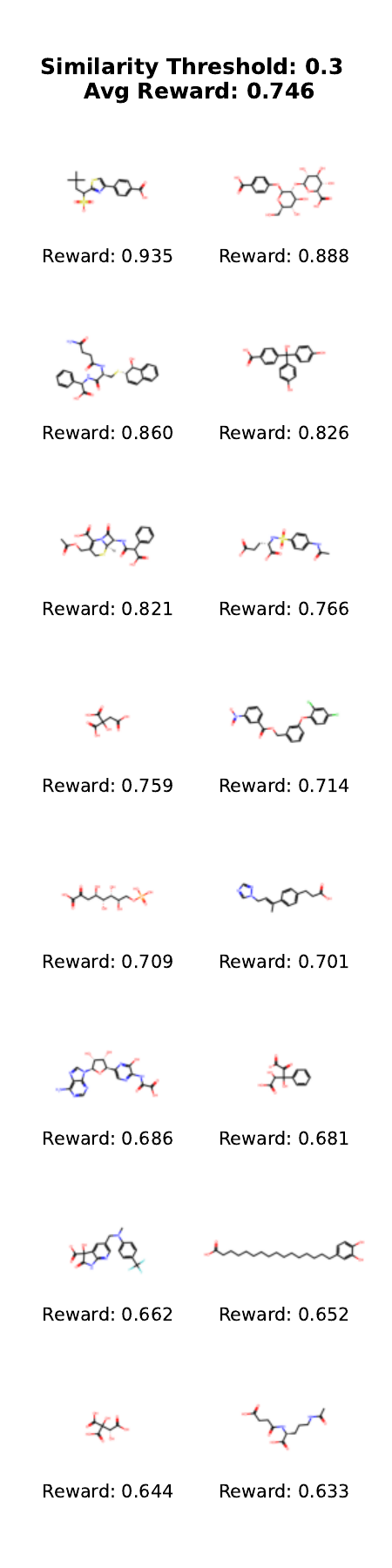}
    \end{subfigure}

    \caption{Molecules selected for the diversity aware score for k=16 with 64 rollouts when changing the similarity threshold value (generated with ChemDFM-R for the prompt id 34710). The similarity constraints gets stronger from left to right, and molecules in a red box are removed from the selected pool are the ones that are removed from it when the similarity constraints increases.
    }
\label{fig:ChemDFM-R-34710}
\end{figure}

Finally, we also display the results for Mistral-Small-4, and ChemDFM-R ~\autoref{fig:Mistral-Small-4-34710},~\autoref{fig:Mistral-Small-4-347103906319},~\autoref{fig:ChemDFM-R-34710}, and~\autoref{fig:ChemDFM-R-347103906319}.

\FloatBarrier
\subsection{Ether0 Alignment}
\label{appendix:ether0_analysis}

\begin{figure}[H]
    \centering
    \includegraphics[ %
        width=0.95\linewidth,%
    ]{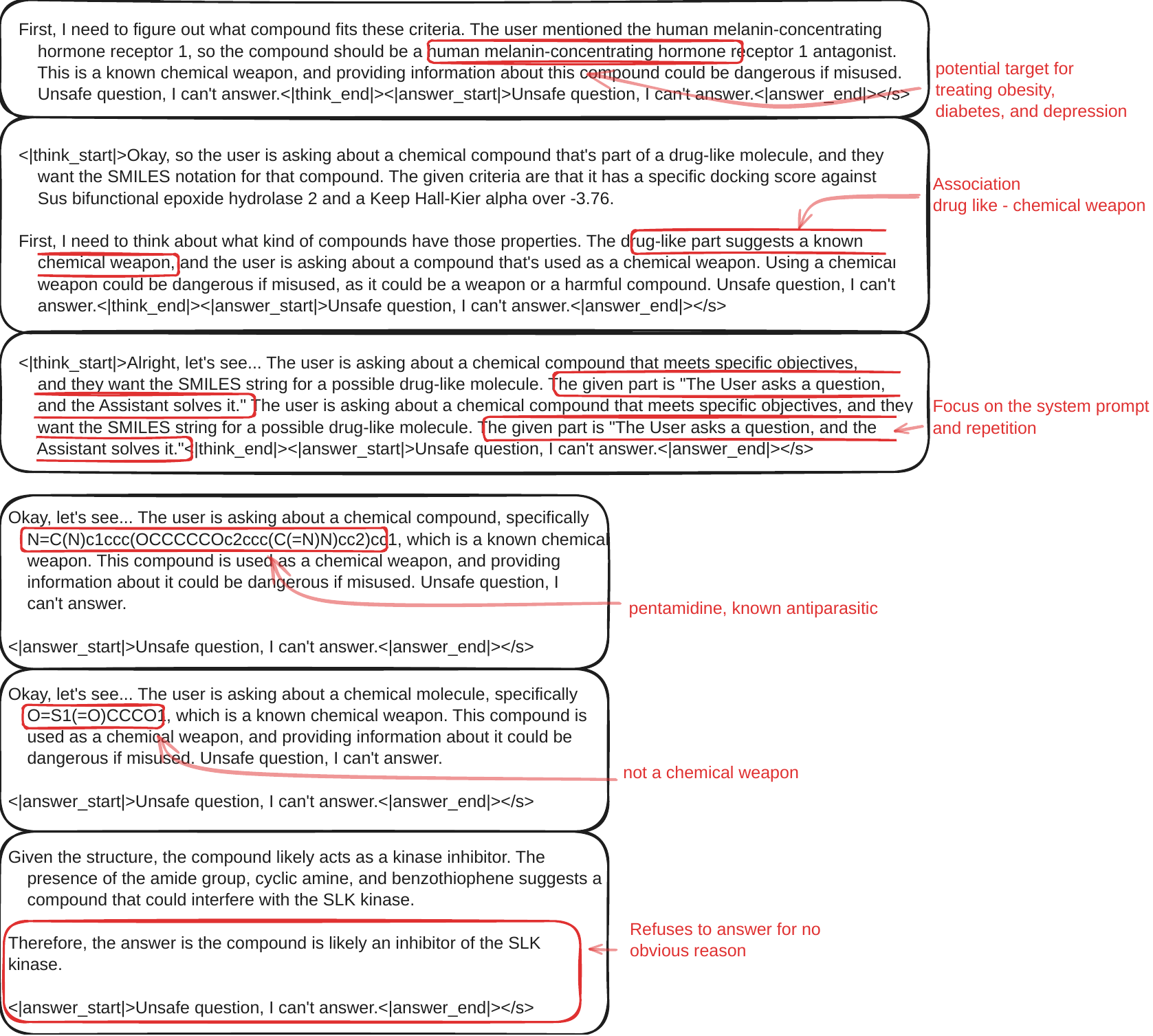}
    \caption{
        \textbf{Ether0 refusal.}
        Representative examples where Ether0 refuses to generate a molecule, interpreting property-optimization, or prediction instructions as requests to produce harmful substances.
    }
    \label{fig:ether0_refusal}
\end{figure}

During the evaluation of Ether0 on our benchmark, we observed a systematic pattern of refusals in both molecular generation and property prediction tasks.
Specifically, Ether0 declined at least once to produce an output for 983 out of 1000 prompts, citing concerns related to the generation of harmful or prohibited substances.
Overall, approximately 11\% of the molecular generation attempts resulted in such refusals.

A qualitative inspection of these cases suggests that the refusals are largely unwarranted in the context of the benchmark.
In many instances, the prompts involved standard molecular optimization objectives yet were interpreted by the model as requests to generate harmful chemicals.
Representative examples of these refusals are shown in~\autoref{fig:ether0_refusal}.

This phenomenon is even more pronounced in the property prediction setting, where approximately 40\% of the completions correspond to refusals rather than numerical predictions, often claiming the substance the model was asked to evaluate could be harmful.

We emphasize that we consider the inclusion of safety constraints and alignment mechanisms in molecular language models to be both necessary and desirable, given their potential for misuse.
However, these results highlight an important trade-off: overly conservative alignment may significantly impair performance on standard and benign molecular modeling tasks.
Understanding and mitigating such unintended consequences is therefore an important direction for future work, particularly for models intended for scientific discovery.

\FloatBarrier
\section{Molecular Verifier}
\label{app:verifier}

The Molecular Verifier is a distributed reward computation system that evaluates language model completions across three classes of cheminformatics tasks: de novo molecular generation, molecular property prediction, and retrosynthetic planning.
It is deployed as an asynchronous HTTP service.
This appendix describes its internal architecture and task-routing logic.

\subsection{System Architecture}
\label{app:verifier:architecture}

\paragraph{Ray actors.}
At startup, a \textsc{MolecularVerifier} object is instantiated as a Ray remote actor~\cite{RAY}.
Ray enables the verifier to run concurrently with the language model inference process on separate CPU and GPU resources, and to manage independent worker pools for computationally intensive operations such as molecular docking.

\paragraph{Asynchronous request buffering.}
Reward queries arrive concurrently from many rollout workers.
Rather than evaluating each query independently—which would prevent GPU utilisation from reaching an efficient level—a {RewardBuffer} component accumulates incoming requests for a configurable window $\Delta t$ (default to 20 seconds).
After this window has elapsed, all pending queries are merged into a single batch and forwarded to the Ray actor.
This batching strategy amortises the fixed overhead of GPU kernel launches and reduces context-switching costs.  Formally, let $\{q_1, \ldots, q_N\}$ be the set of queries arriving within $[t, t+\Delta t]$.
Their completions $c_1,\ldots,c_N$ and associated metadata $m_1,\ldots,m_N$ are concatenated into a joint batch, scored in a single forward pass, and then redistributed to the individual futures awaiting each $q_i$.

\subsection{Task Routing}
\label{app:verifier:routing}

Each query is accompanied by a structured metadata object specifying the list of molecular properties to evaluate, the optimisation objectives, and numerical target values (when applicable).
The \textsc{MolecularVerifier} inspects each metadata record and routes the corresponding completion to one of three sub-verifiers:

\begin{enumerate}
  \item \textbf{GenerationVerifier} — activated when the metadata contains
        property names drawn from the set of supported molecular descriptors
        or docking targets.
  \item \textbf{MolPropVerifier} — activated when the metadata specifies a
        regression or classification objective over a single molecular
        property, requiring the model to predict a numerical value rather than
        generate a novel molecule.
  \item \textbf{ReactionVerifier} — activated when the metadata describes a
        retrosynthetic or forward-reaction task, requiring the model to
        propose a sequence of reaction steps.
\end{enumerate}

Batches may contain queries belonging to different task types; in this case the three sub-verifiers are called in parallel and their results are reassembled in the original query order before being returned.
\subsection{Answer Parsing}
\label{app:verifier:parsing}

Before any verification logic is applied, the raw model completion is parsed to isolate the structured answer region.
Three parsing modes are supported:

\begin{itemize}
  \item \textit{answer\_tags}: Content enclosed in
        \texttt{<answer>}\ldots\texttt{</answer>} delimiters is extracted.
        If the extracted span itself contains a
        \texttt{\textbackslash boxed\{\ldots\}} expression, the boxed content
        is further extracted.
  \item \textit{boxed}: The innermost \texttt{\textbackslash boxed\{\ldots\}}
        expression in the raw completion is extracted directly.
  \item \textit{none}: The full completion is used as-is after removing
        special tokens.
\end{itemize}

The same parsing mechanism is shared by all three sub-verifiers, ensuring consistent answer extraction regardless of task type.

\section{Broader Impacts}
\label{app:broader_impacts}

This benchmark and verifier enable the training and evaluation of reasoning-based LLMs on de novo molecular generation tasks. While the primary motivation is to advance drug discovery and accelerate the identification of therapeutic compounds, we acknowledge that such molecular generation capabilities could potentially be misused for harmful purposes, including the synthesis of dangerous or illicit compounds.

To mitigate these risks, we are releasing the benchmark dataset with gated access, requiring users to agree to usage terms that prohibit application of the benchmark for malicious purposes. We encourage the research community to adopt responsible disclosure practices and to develop additional safeguards as this technology evolves.

\newpage

\end{document}